\pdfoutput=1

\documentclass[11pt]{article}

\usepackage{acl}

\usepackage{times}
\usepackage{latexsym}
\usepackage[most]{tcolorbox}
\tcbuselibrary{listingsutf8}

\usepackage[T1]{fontenc}
\usepackage{booktabs} 
\usepackage{multirow} 
\usepackage{array}    
\usepackage{graphicx} 
\usepackage{multirow}
\usepackage{makecell}
\usepackage[utf8]{inputenc}

\usepackage{microtype}

\usepackage{inconsolata}
\usepackage{algorithm}
\usepackage{algorithmic}
\usepackage{amsmath}
\usepackage{xparse}
\usepackage{xcolor}

\usepackage{graphicx}
\usepackage{subcaption}  
\usepackage{caption} 
\newif\ifshowCadenComments
\showCadenCommentstrue 

\title{Can LLMs Reason Abstractly Over Math Word Problems Without CoT? Disentangling Abstract Formulation From Arithmetic Computation}

\author{
  Ziling Cheng$^{1,2}$\textsuperscript{\dag} \quad
  Meng Cao$^{1,2}$ \\
  \textbf{Leila Pishdad}$^{3}$ \quad
  \textbf{Yanshuai Cao}$^{3}$ \quad
  \textbf{Jackie Chi Kit Cheung}$^{1,2,4}$ \\
  $^{1}$Mila – Quebec AI Institute \quad
  $^{2}$McGill University \quad
  $^{3}$Borealis AI \quad
  $^{4}$Canada CIFAR AI Chair \\
  \texttt{\{ziling.cheng, meng.cao\}@mail.mcgill.ca} \\
  \texttt{\{leila.pishdad, yanshuai.cao\}@borealisai.com}, \texttt{cheungja@mila.quebec}
}

\begin{document}

\maketitle

\begingroup
\renewcommand{\thefootnote}{\dag}
\footnotetext{Work done during a Mitacs internship at Borealis AI.}
\endgroup

\begin{abstract}
Final-answer-based metrics are commonly used for evaluating large language models (LLMs) on math word problems, often taken as proxies for reasoning ability. However, such metrics conflate two distinct sub-skills: \textbf{abstract formulation} (capturing mathematical relationships using expressions) and \textbf{arithmetic computation} (executing the calculations). 
Through a disentangled evaluation on GSM8K and SVAMP, we find that the final-answer accuracy of Llama-3 and Qwen2.5 (1B-32B) without CoT is overwhelmingly bottlenecked by the arithmetic computation step and not by the abstract formulation step. Contrary to the common belief, we show that CoT primarily aids in computation, with limited impact on abstract formulation. Mechanistically, we show that these two skills are composed conjunctively even in a single forward pass without any reasoning steps via an \textbf{abstract-then-compute} mechanism: models first capture problem abstractions, then handle computation. Causal patching confirms these abstractions are present, transferable, composable, and precede computation. These behavioural and mechanistic findings highlight the need for disentangled evaluation to accurately assess LLM reasoning and to guide future improvements.\footnote{Code and data will be made publicly available upon acceptance.}

\end{abstract}

\section{Introduction}
\begin{figure*}[ht]
    \centering
    \includegraphics[width=0.8\linewidth]{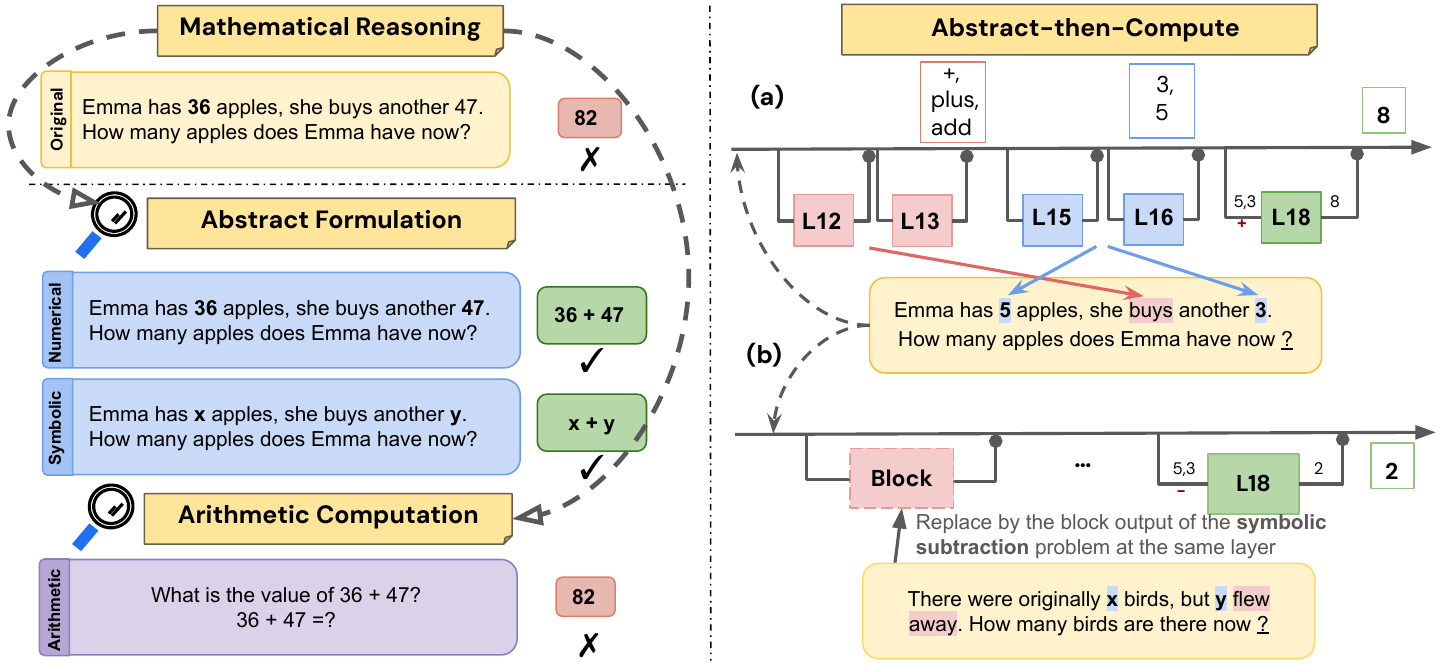}
    \caption{\textbf{Left (Disentangled evaluation framework):} Final-answer accuracy obscures reasoning ability due to conflating abstract formulation and arithmetic computation. \textbf{Right (Abstract-then-Compute Mechanism in Llama-3 8B):} (a) Residual stream at the last token position shows that models first capture problem abstraction (L13-14), followed by computation (L18). (b) Same as (a), but one critical layer output is patched with a different symbolic abstraction (e.g., $x-y$), causally changing the computation from $5 + 3 = 8$ to $5 - 3 =2$.}
    \label{fig:motivation}
\end{figure*}

Large language models (LLMs) have demonstrated impressive progress on various math problem datasets \cite{cobbe2021training, hendrycksmath2021, patel-etal-2021-nlp}, often leveraging Chain-of-Thought (CoT) prompting \cite{wei2022chain}. Despite the availability of step-by-step reasoning chains, standard evaluation predominantly relies on final-answer accuracy (comparing the model's final numerical output against a gold answer), which reduces model performance to a single metric \cite{liu-etal-2024-ecbd, opedal2024language}. This reduction limits the possible insights when diagnosing LLMs' reasoning abilities, especially in zero-shot scenarios without CoT. When an LLM fails to produce the correct answer, is it due to ``reasoning deficits'', or could it be a calculation error?

To investigate this, we propose a disentangled evaluation framework that separately measures two core skills of mathematical problem-solving (See Figure~\ref{fig:motivation}): (1) \textbf{abstract formulation} (hereafter, abstraction) — the ability to identify relevant quantities and translate the natural language problem into its underlying mathematical relationships (e.g., $36+47$ or $x+y$ in Figure~\ref{fig:motivation}); and (2) \textbf{arithmetic computation} (hereafter, computation) — the capacity to calculate the final answer from that expression (e.g., evaluate $36+47$ to $83)$.

Using this disentangled evaluation on GSM8K \cite{cobbe2021training} and SVAMP \cite{patel-etal-2021-nlp} with Llama-3 and Qwen-2.5 models (1B-32B), we find that even without CoT: (i) models surprisingly perform better at abstraction than computation, despite the former's perceived conceptual complexity. (ii) if deriving the final answer in math word problems depends on these two skills conjunctively, final-answer accuracy alone may give a misleading picture of models' reasoning abilities in math word problems. Moreover, we show that CoT primarily improves computation, with limited gains in abstraction, further demonstrating the value of disentangled evaluation.

While these behavioural findings suggest that models can formulate abstractions without explicit CoT when separately prompted, it remains unclear whether abstraction and computation are composed conjunctively when deriving the final answer during single-pass inference. To explore this, we move beyond outcome-based evaluation, and conduct mechanistic interpretability analyses. Using logit attribution and activation patching, we identify a consistent and sequential \textbf{abstract-then-compute} mechanism (see Figure~\ref{fig:motivation}a). Moreover, cross-prompt patching provides evidence that models do form abstractions internally independent of the surface form (numerical or symbolic,  see Figure~\ref{fig:motivation}b): when these symbolic abstractions (e.g. $x-y$) are transferred into a different problem, they are utilized and composed with the subsequent computation stages, altering the final answer.

\noindent \textbf{Contributions}: (i) Through disentangled evaluation, we show that without CoT, models exhibit stronger reasoning ability than final-answer accuracy suggests, and that CoT primarily aids calculation. (ii) Using mechanistic interpretability, we uncover an abstract-then-compute mechanism in a single-pass generation, where abstractions are transferrable across problem variants. Collectively, our findings suggest an alternative narrative: poor final-answer accuracy without CoT \cite{wei2022chain, sprague2025to}, or performance declines on problem variants \cite{zhang2024a, shi2023large, mirzadeh2025gsmsymbolic}, can stem from arithmetic errors rather than reasoning deficits.
    
\section{Related Work}

\paragraph{Mathematical Reasoning Evaluation}  

Existing math problem-solving benchmarks spans elementary word problems \cite{cobbe2021training, patel-etal-2021-nlp, amini-etal-2019-mathqa, miao-etal-2020-diverse, ling-etal-2017-program, koncel-kedziorski-etal-2016-mawps, shi2015automatically} to higher levels \cite{hendrycksmath2021,hendrycks2021measuring, zhong-etal-2024-agieval, zhang2023evaluating, he-etal-2024-olympiadbench}. Early datasets paired expressions with answers, but evaluation largely focused on final-answer-based metrics \cite{patel-etal-2021-nlp, shi2015automatically}. With the rise of LLMs and CoT prompting \cite{wei2022chain}, rationale-based formats became common \cite{hendrycksmath2021, cobbe2021training}, yet standard evaluations still predominantly use final-answer metrics, and occasionally code execution from rationales \cite{mishra-etal-2022-lila, gao2023pal}. In contrast, we move beyond this final-answer-centric paradigm, by decomposing problem-solving into abstract formulation and arithmetic computation, inspired by the cognitive theories \cite{opedal2024language}.

\paragraph{Memorization vs. Generalization} 
Variants of math word problems with perturbations were introduced to test generalization beyond memorization \cite{zhang2024a, ye2025physics, gao2023pal, shi2023large, li2024gsm, mirzadeh2025gsmsymbolic}. While performance drops are often interpreted as reasoning failures, our results suggest they may instead stem mainly from arithmetic errors, pointing to a different improvement strategy.

\paragraph{Mechanistic Interpretability}
Mechanistic interpretability methods, such as logit attribution \cite{nostalgebraist2020, belrose2023eliciting} and causal patching \cite{goldowsky2023localizing, wang2023interpretability, meng2022locating, zhang2023towards,  merullo-etal-2024-language, cheng2025stochasticchameleonsirrelevantcontext}, have been used to trace model computations. Prior work on math reasoning largely focuses on the mechanisms behind arithmetic computations \cite{nikankin2025arithmetic, zhang2024interpreting}, while recent work on word problems use probing classifiers to track explicit variable reasoning \cite{ye2025physics}. In contrast, we examine implicit reasoning within a single forward pass to uncover abstraction beyond computation.

\begin{figure}[ht]
    \centering

    \begin{subfigure}[t]{0.55\linewidth}  
        \includegraphics[width=\linewidth]{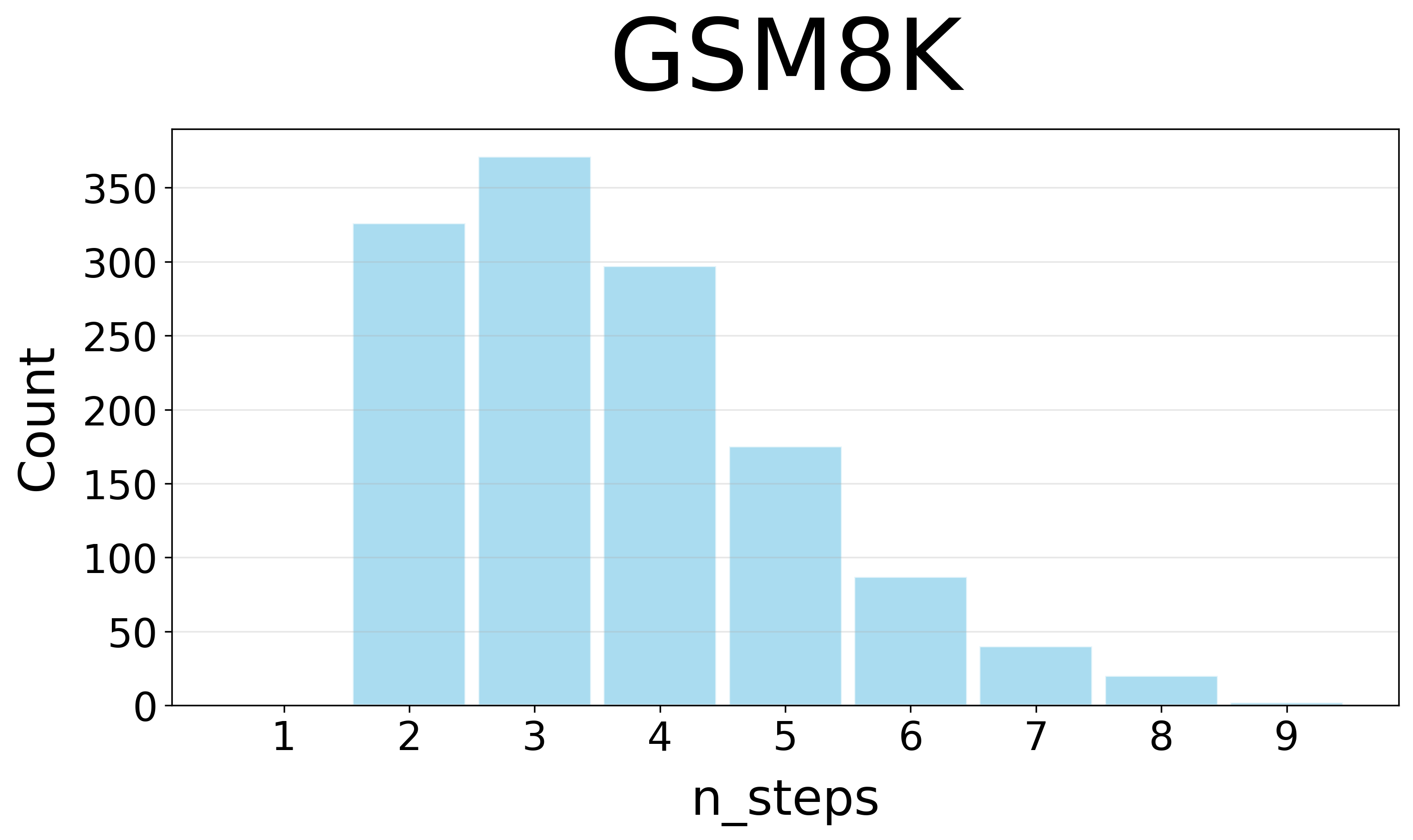}
        \caption{N-steps}
        \label{fig:construction}
    \end{subfigure}
     \begin{subfigure}[t]{0.3\linewidth}
        \includegraphics[width=\linewidth]{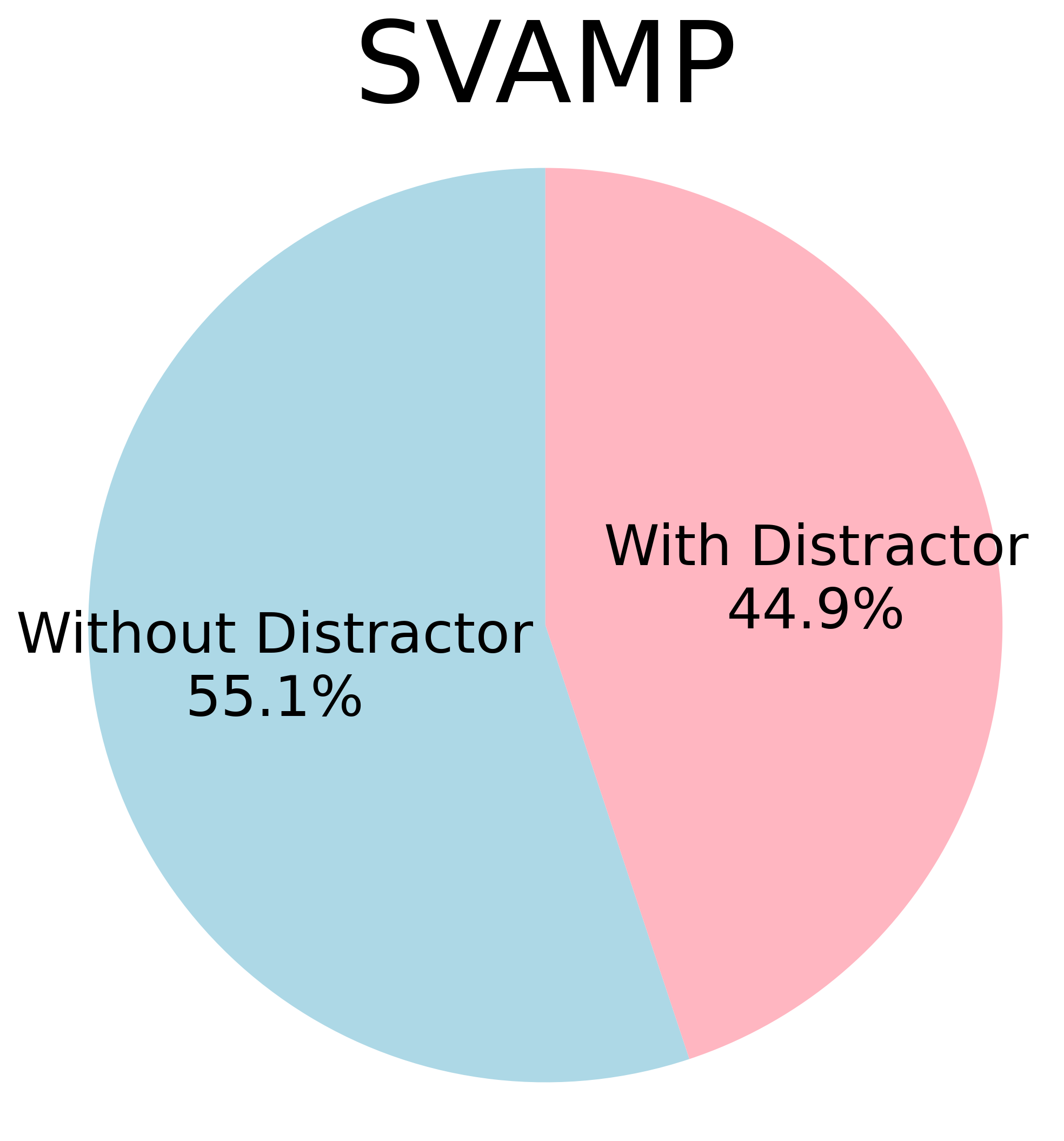}
        \caption{Distractor}
        \label{fig:example}
    \end{subfigure}
    \caption{Distribution of problem characteristics by number of reasoning steps (GSM8K) and presence of distractors (SVAMP).}
    \label{fig:dataset_stats}
\end{figure}

\begin{table*}
\centering
\small
\begin{tabular}{@{}p{1.8cm} p{1.8cm} p{7cm} p{3.6cm}}
\toprule
\textbf{Setting} & \textbf{Skills Tested} & \textbf{Question Form and Example} & \textbf{Answer Form and Example} \\
\midrule
\textbf{Original} & Abstraction +Computation & \underline{Numerical}: Weng earns \$12 for every hour she works. If she worked for 50 minutes, how much did she earn? & \underline{Number}: 10 \\
\midrule
\textbf{Arithmetic \newline Computation} & Computation & \underline{Numerical}: What is the value of $12 \times \left(\frac{50}{60}\right)$? \newline or $12 \times \left(\frac{50}{60}\right) =$? & \underline{Number}: 10 \\

\midrule
\textbf{Numerical \newline Abstraction} & Abstraction & \underline{Numerical}: Weng earns \$12 for every hour she works. If she worked for 50 minutes, how much did she earn? & \underline{Expression}: $12 \times \left(\frac{50}{60}\right)$ \\

\midrule
\textbf{Symbolic \newline Abstraction} & Abstraction  & \underline{Symbolic}: Weng earns \$x for every hour she works. If she worked for y minutes, how much did she earn? & \underline{Expression}: $x \times \left(\frac{y}{60}\right)$  \\

\bottomrule
\end{tabular}
\caption{Disentangled evaluation in math word problems with tested skills, varying by question and answer forms. Instructions in Appendix Table~\ref{tab:experiments}.}
\label{tab:eval_settings}
\end{table*}

\section{Dataset and Experimental Design}

\paragraph{Task and Dataset} We study math word problems using GSM-8K \cite{cobbe2021training} and SVAMP \cite{patel-etal-2021-nlp}. GSM-8K spans 2–8 steps without distractors (See Figure~\ref{fig:dataset_stats} for statistics), while SVAMP involves single-step reasoning with distractor variants. To evaluate abstract formulation, we create symbolic variants: SVAMP expressions are templated into variable-based forms; GSM-8K symbolic versions from the test set are generated using \texttt{gpt-4o-mini} \cite{openai2024} via a two-stage \textit{generate-then-validate} to ensure the correctness. See Appendix~\ref{app: gsm8k-svamp} and Table~\ref{tab:symbolic_examples} for details and examples. For interpretability, we generate 3,600 simple 1–2 step\footnote{We focus on 1–2 step problems, as models often fail simple word problems involving multi-step computations in a single forward pass.} word problems involving basic operations ($+, -, \times, \div$) from 1,200 diverse LLM-generated templates, covering varied scenarios, verb choices, entities, names and sentence structures. See Appendix~\ref{sec:data_interpretability} and Table~\ref{tab:interpretability_data_examples} for details and examples.

\paragraph{Models} We evaluate instruction-tuned Llama-3 (1B, 3B, 8B) \cite{grattafiori2024Llama} and Qwen 2.5 \cite{Yang2024Qwen25TR} (3B, 7B, 14B, 32B) models. Mechanistic interpretability analyses focus on Llama-3 8B, Qwen 2.5 7B, and Qwen 2.5 14B.

\paragraph{Evaluation} 
All experiments use greedy decoding and FP16 precision on RTX 8000/A100L GPUs. Numeric answers are evaluated via normalized Exact Match. Symbolic expressions are evaluated using \texttt{gpt-4o-mini} (94\% agreement with humans on 120 samples, prompt and details in Appendix~\ref{app:eval_details}) and with \texttt{sympy} for numeric expressions. We report standard accuracy. CoT generations are capped at 512 tokens. See Appendix~\ref{app:eval_details} for details.

\begin{figure*}[h]
    \centering
    \includegraphics[width=\linewidth]{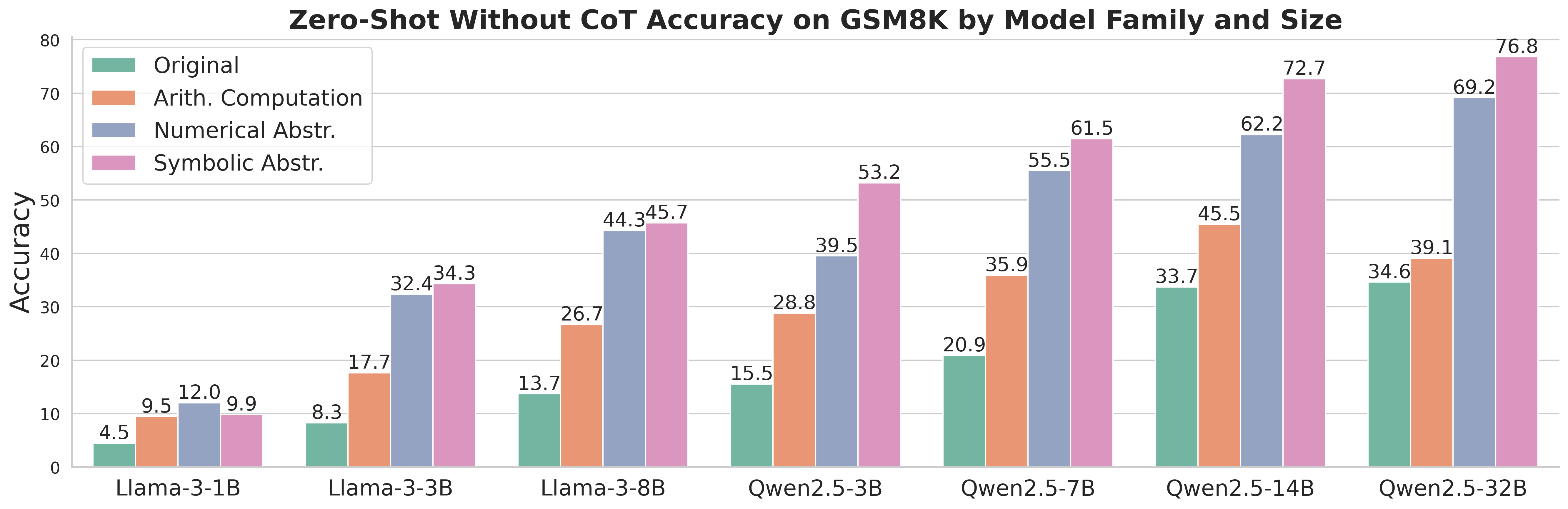}
    \caption{Model zero-shot \textbf{without CoT} performance on GSM8K. (i) Models exhibit much better abstraction performance (\textit{Symbolic} and \textit{Numerical}) than in actually computing the expressions (\textit{Arithmetic Computation}). (ii) Final-answer accuracy in the \textit{Original} setting may provide a misleading picture of models' reasoning ability, possibly due to arithmetic limitations.}
    \label{fig:computation_zs}
\end{figure*}

\section{Disentangled Evaluation}\label{sec:eval_results}
We first introduce the disentangled evaluation framework, then present results without CoT in Sec.~\ref{sec:no_cot}, followed by an analysis of CoT's impact in Sec.~\ref{sec:with_cot}.

\paragraph{Framework}

Suppose a task \( T \) can be decomposed into a set of sub-skills \( \{s_1, s_2, \dots, s_n\} \), such that solving \( T \) requires executing these skills conjunctively (i.e., \( T = s_1 \cap s_2 \cap \dots \cap s_n \)). Disentangled evaluation aims to assess each sub-skill \( s_i \) independently via a corresponding subtask \( t_i \), designed to isolate and test that specific skill. Let \( \mathrm{Eval}(T) \) denote the evaluation metric on the full task, and \( \mathrm{Eval}(t_i) \) the metric for subtask \( t_i \). Measuring \( \mathrm{Eval}(t_1), \dots, \mathrm{Eval}(t_n) \) enables finer-grained attribution of performance, identifying failure of specific skills. In math word problems, let $(Q, E, A)$ be the question, expression and answer triplets, we decompose mathematical problem-solving into \textbf{abstract formulation} (translating $Q$ to mathematical relationships $E$) and \textbf{arithmetic computation} (executing the calculation from $E$ to produce $A$). Besides the standard \textit{Original} setting (requiring both abstraction and computation), we design three targeted subtasks: \textit{Symbolic Abstraction}, which assess abstraction using symbolic variables; \textit{Numerical Abstraction}, evaluating abstraction with concrete numbers but without computation; and \textit{Arithmetic Computation}, which directly tests execution of fully specified expressions from $Q$. See Table~\ref{tab:eval_settings} and Appendix~\ref{app:eval_details} for details.

\subsection{Understanding Model Failures: Reasoning or Arithmetic Error?}\label{sec:no_cot}

We first apply disentangled evaluation \textbf{zero-shot without CoT} across multiple model sizes of Llama-3 and Qwen2.5 families. As shown in Figure~\ref{fig:computation_zs}, the error rates are consistently lower for abstract formulation (both \textit{Numerical} and \textit{Symbolic Abstraction}) compared to arithmetic computation. This suggests that if final-answer accuracy in the \textit{Original} setting depends on both competencies conjunctively, poor performance observed in the \textit{Original} setting could stem from arithmetic computation failures, rather than reasoning deficits. Consequently, this indicates that final-answer accuracy alone from the \textit{Original} setting may substantially mislead a model's underlying reasoning ability. See additional results in Appendix~\ref{app:additional_eval_results}. To assess the reliability and external validity of the symbolic abstraction evaluation, we perform ablations over symbol order and symbol choice in Appendix~\ref{app:symbolic_ablation}.

\begin{table}[h]
\centering
\small
\begin{tabular}{lccccr}
\toprule
\textbf{$\Delta$ Accuracy} & \textbf{8B} & \textbf{7B} & \textbf{14B} & \textbf{32B} & \textbf{Avg.} \\
\midrule
Original & 64.8 & 68.5 & 58.4 & 59.7 & 62.8 \\
Arith. Comp. & 64.8 & 60.5 & 51.2 & 58.2 & 58.7 \\
Numerical Abstr. & 15.8 & 21.6 & 21.6 & 11.6 & 17.6 \\
Symbolic Abstr. & 11.0 & 13.2 & 1.1 & 1.3 & 6.7 \\
\bottomrule
\end{tabular}
\caption{Accuracy difference (\%) with and without CoT. Results are shown for Llama 3 (8B) and Qwen2.5 models (7B, 14B, 32B).}
\label{tab:cot_deltas}
\end{table}

\subsection{Disentangling CoT Gains}\label{sec:with_cot}
We now apply disentangled evaluation with CoT to disentangle CoT gains (Table~\ref{tab:cot_deltas}). We show that CoT yields the largest gains in computation (e.g., +62.8\%), confirming its effectiveness in multi-step arithmetic. In contrast, abstraction shows limited improvement (e.g., +6.7\% for \textit{Symbolic abstraction} and +17.6\% for \textit{Numerical abstraction}), even with extended generation budgets (512 tokens), suggesting CoT is less helpful for abstraction. Gains in the \textit{Original} setting (e.g., +62.8\%) likely reflect a mix of benefits from both components and possible data leakage. See additional results in Appendix~\ref{app:cot_ablation}.

\noindent \textbf{Summary:} These findings challenge the view that poor final-answer accuracy in math reasoning benchmarks always implies `poor reasoning'. Instead, our disentangled design reveals that many models do possess a level of abstract formulation capabilities, which are often obscured in standard evaluations due to their limited arithmetic competence. Crucially, while abstraction variants indicate far higher performance than the \textit{Original} setting, models are still not perfect — performance in \textit{Symbolic Abstraction} remains far from 100\% (45.7\% for Llama-8B, 76.8\% for Qwen-32B), but the gap is significantly narrower than previously assumed, calling for more precise definition and evaluation of reasoning.

\section{Inside the Model: Probing Abstraction and 
Computation}\label{sec:interpretability_results}
To investigate whether abstraction and computation are composed conjunctively when producing a final numerical answer in a single forward pass, we move beyond outcome-based evaluation and apply mechanistic interpretability. We hypothesize an \textbf{abstract-then-compute} process: first inferring the abstraction (e.g., `+' from ``buys''), then performing the computation (e.g., $5 + 3$). Section~\ref{sec:hypo_generation} identifies key layers for each stage; Section~\ref{sec:validation} validate these layers and tests abstraction transferability across forms (symbolic/concrete) and logic.

\subsection{Uncovering the Abstract-Then-Compute Mechanism in One Forward Pass} \label{sec:hypo_generation}
\begin{figure*}[ht]
    \centering
    \includegraphics[width=1\linewidth]{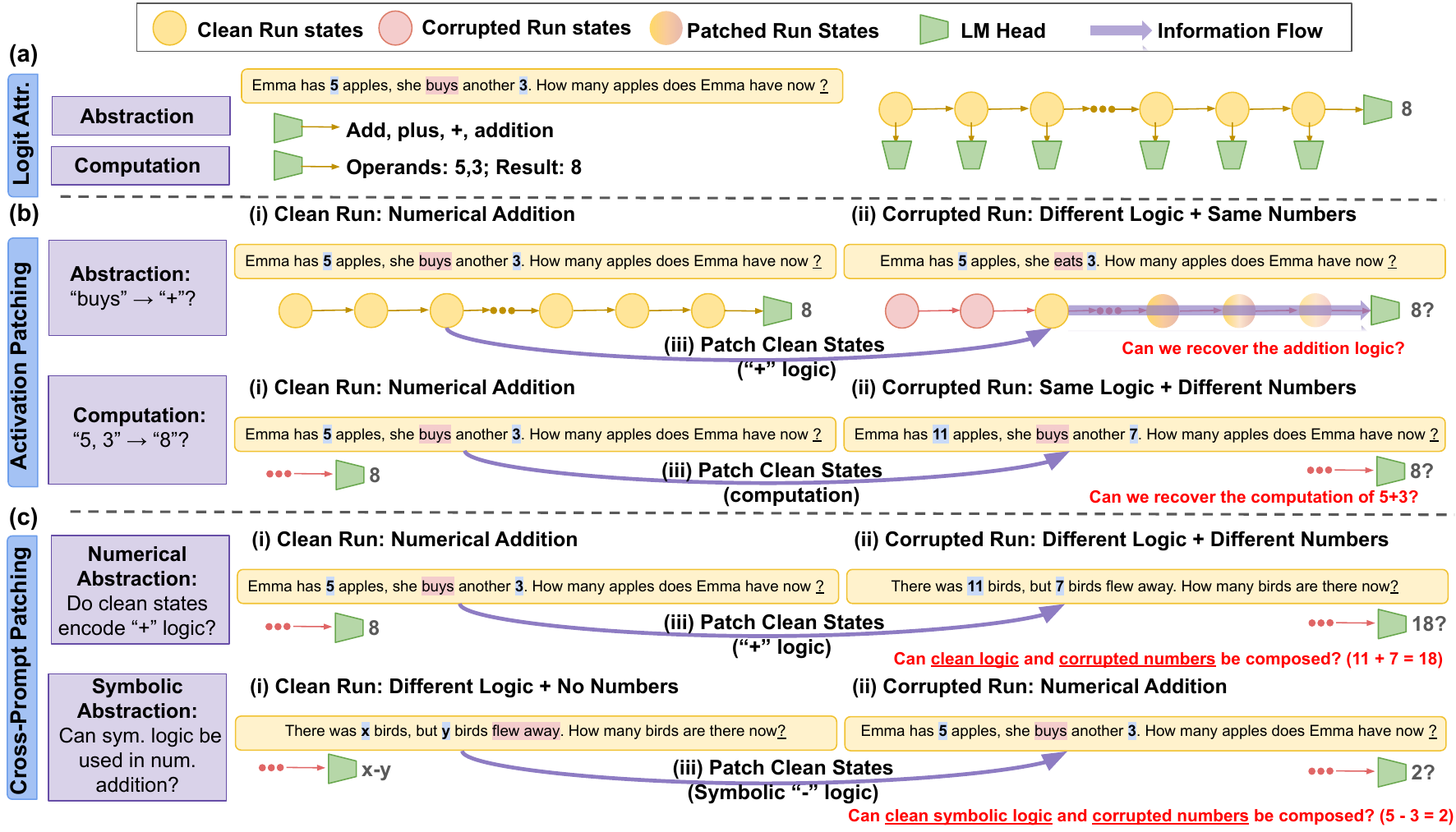}
    \caption{Overview of interpretability methods probing the abstract-then-compute mechanism in simple math problems, focusing on hidden states at the last token position across layers.}
    \label{fig:interp_method}
\end{figure*}

\begin{figure*}[ht]
    \captionsetup[subfigure]{labelformat=simple}
    \renewcommand{\thesubfigure}{(\roman{subfigure})}
    \centering
    \begin{minipage}{0.32\textwidth}
        \includegraphics[width=\linewidth]{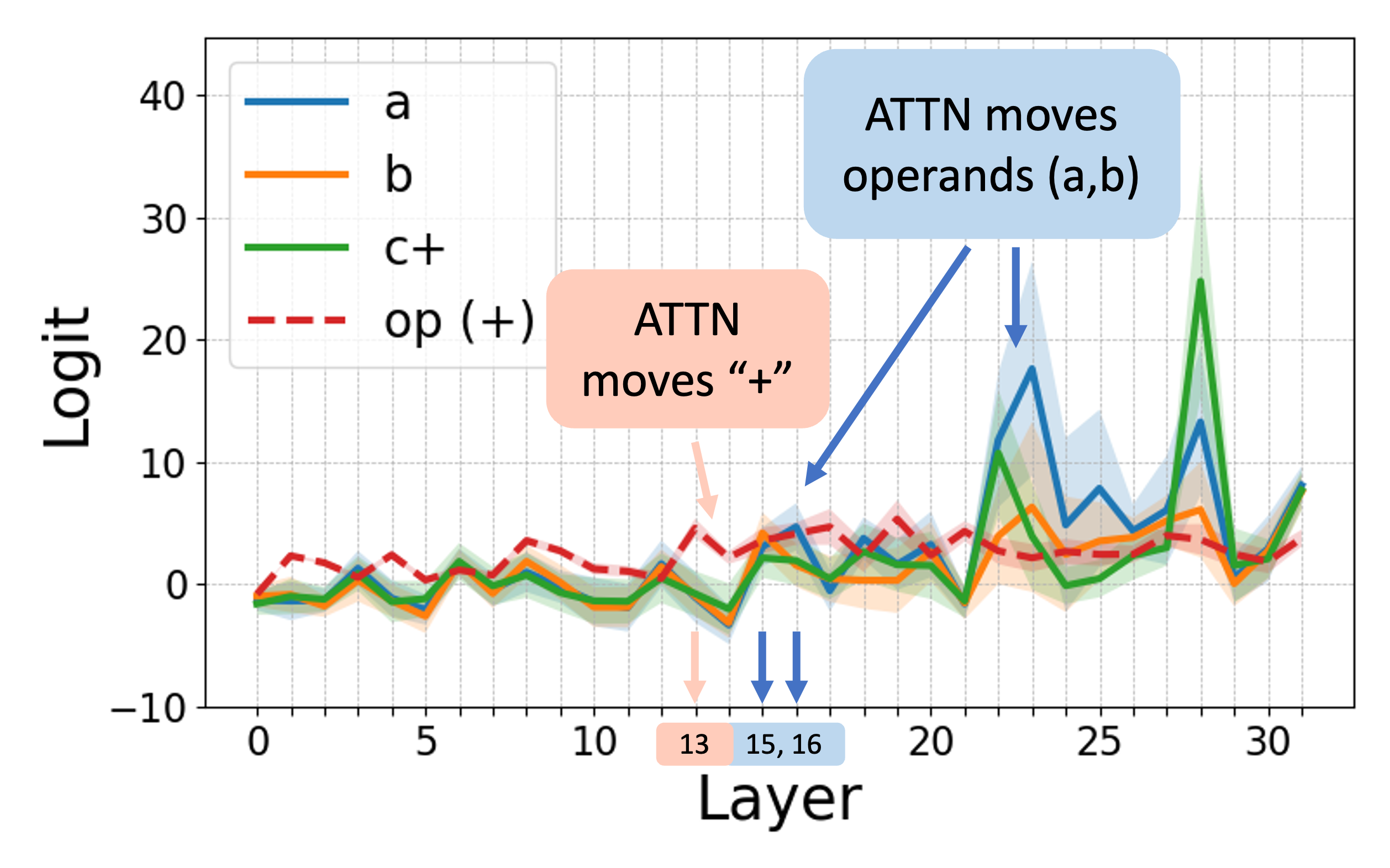}
        \subcaption{Attention}
    \end{minipage}
    \begin{minipage}{0.32\textwidth}
        \includegraphics[width=\linewidth]{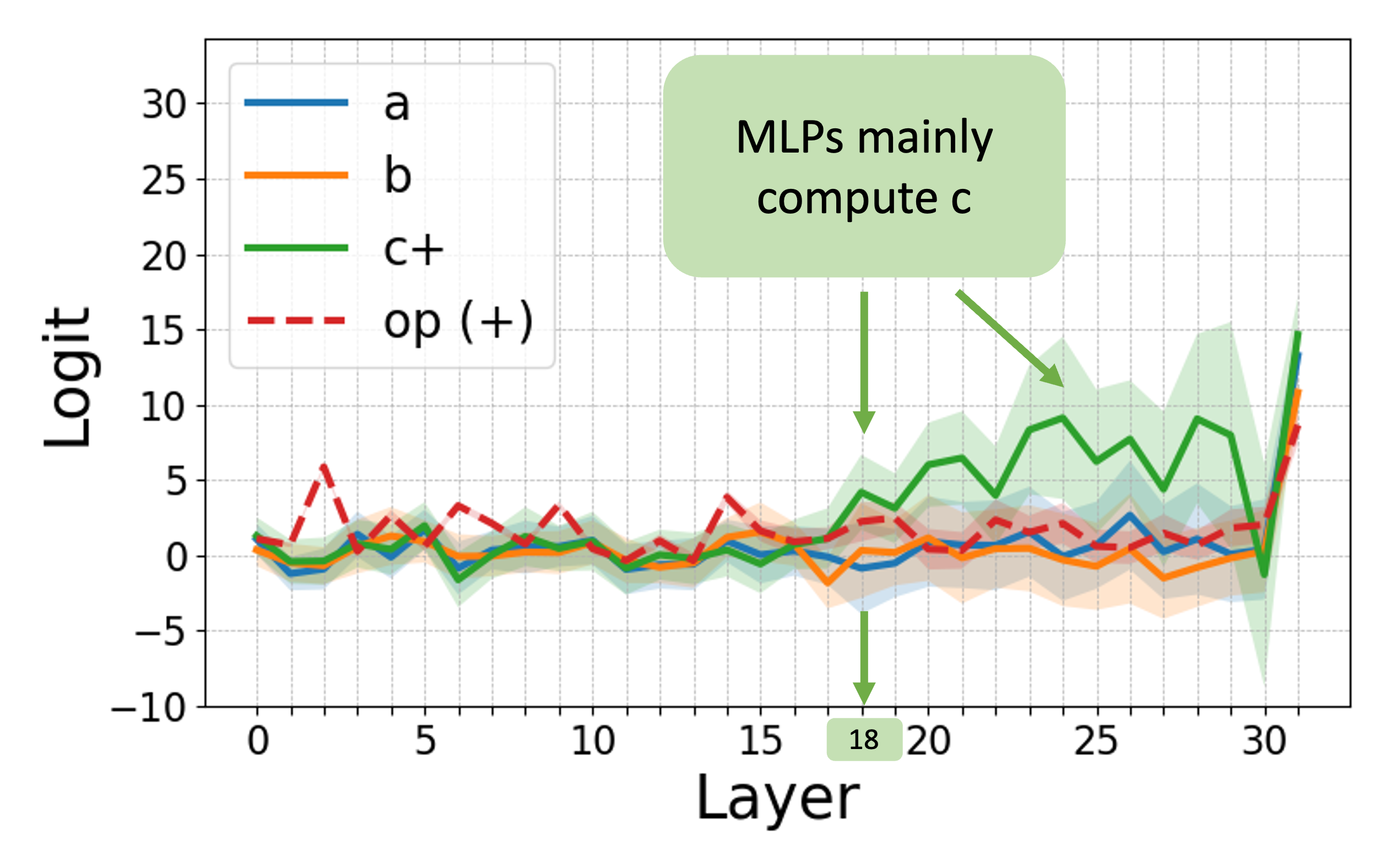}
        \subcaption{MLP}
    \end{minipage}
    \begin{minipage}{0.32\textwidth}
        \includegraphics[width=\linewidth]{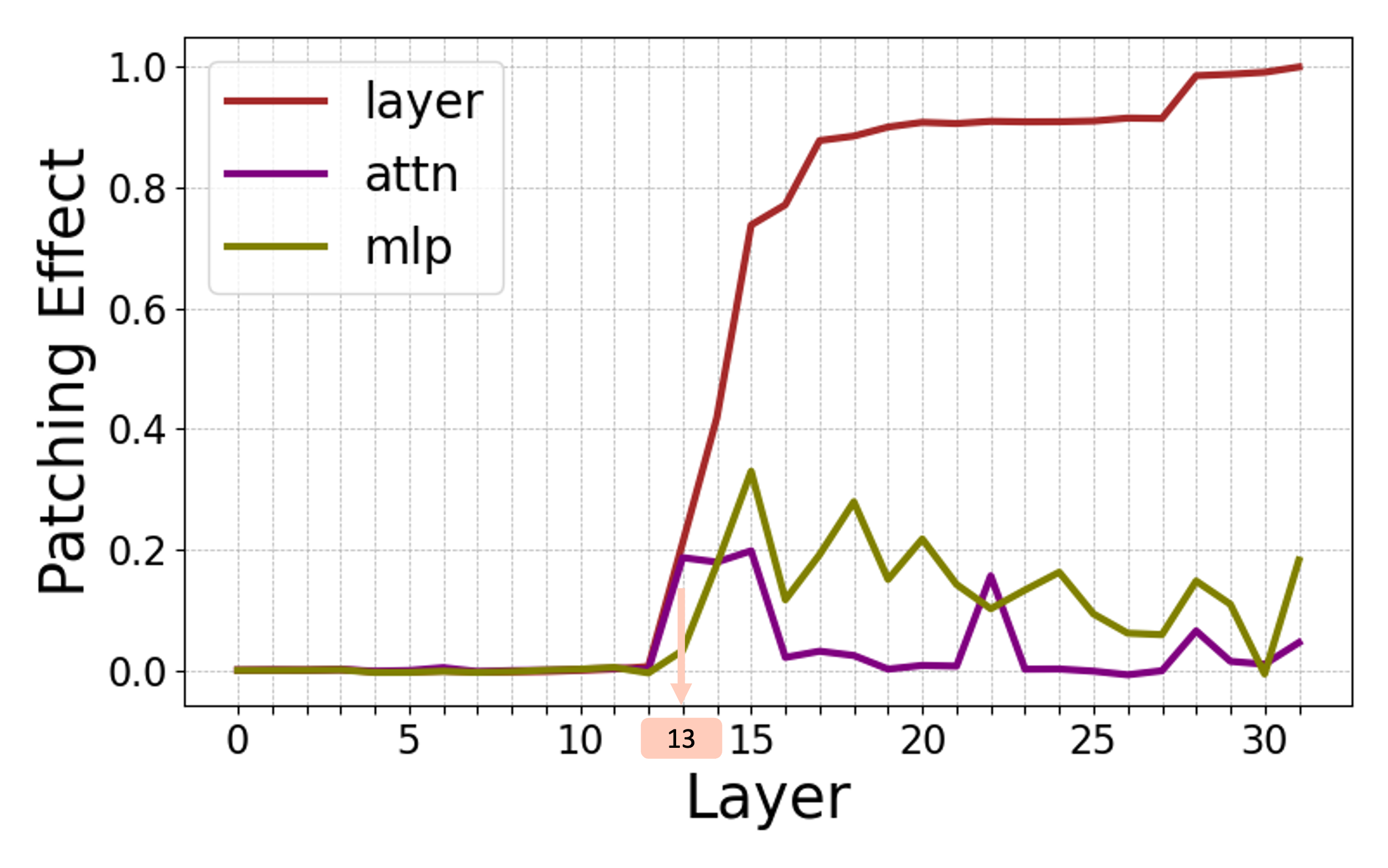}
        \subcaption{Abstraction}
    \end{minipage}


    \begin{minipage}{0.32\textwidth}
        \includegraphics[width=\linewidth]{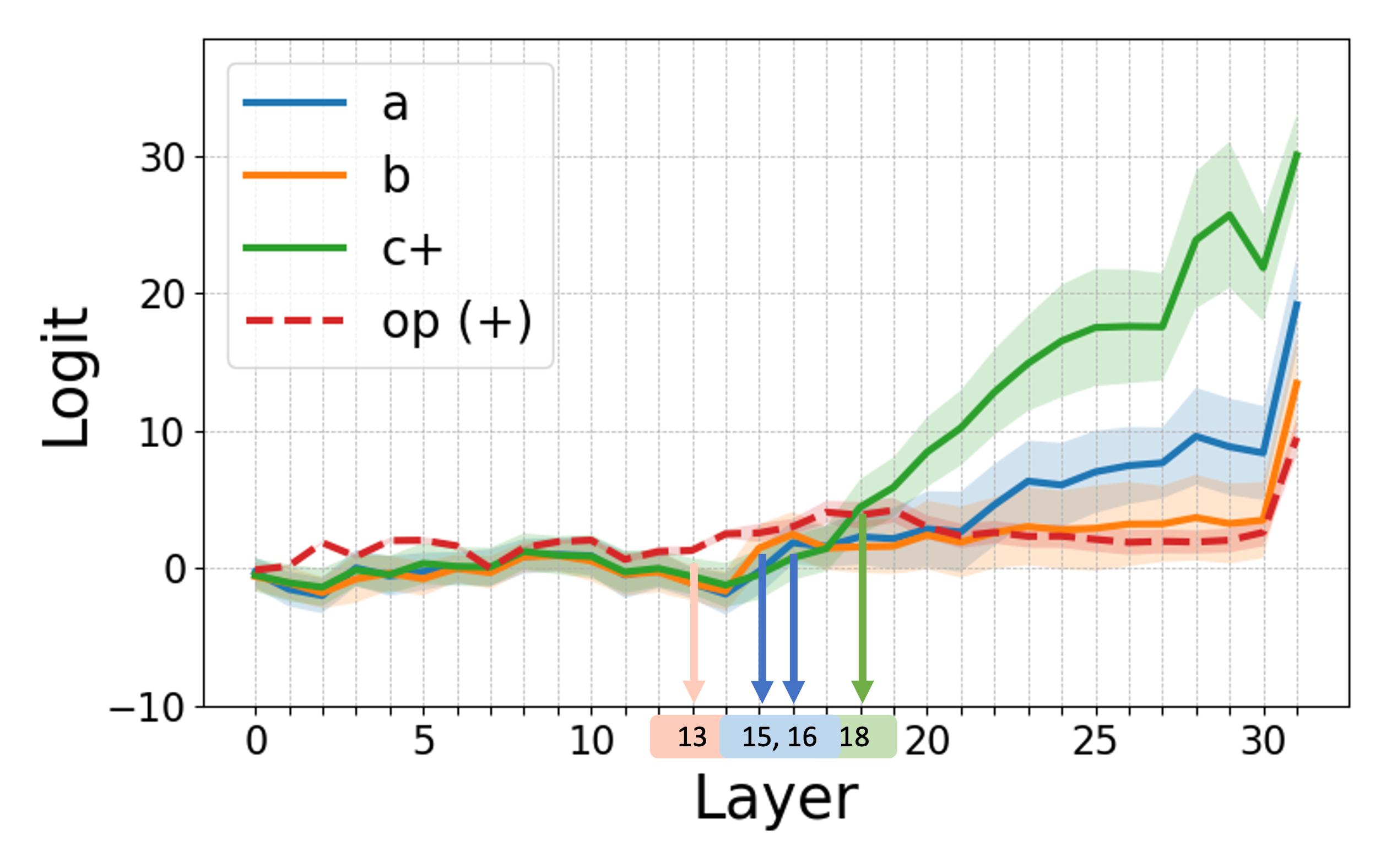}
        \subcaption{Resid Final}
    \end{minipage}
    \begin{minipage}{0.32\textwidth}
        \includegraphics[width=\linewidth]{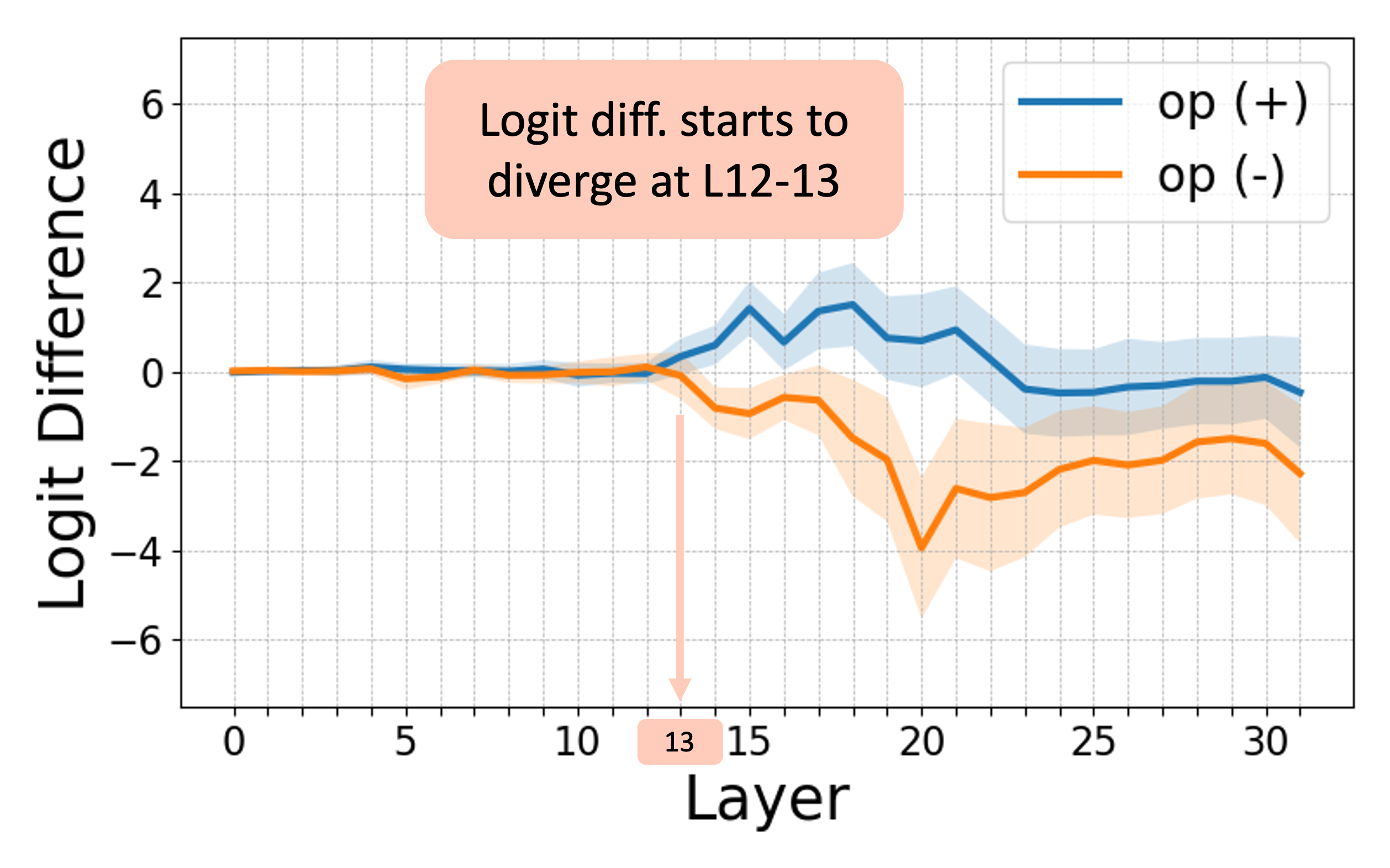}
        \subcaption{Resid Mid}
    \end{minipage}
    \begin{minipage}{0.32\textwidth}
        \includegraphics[width=\linewidth]{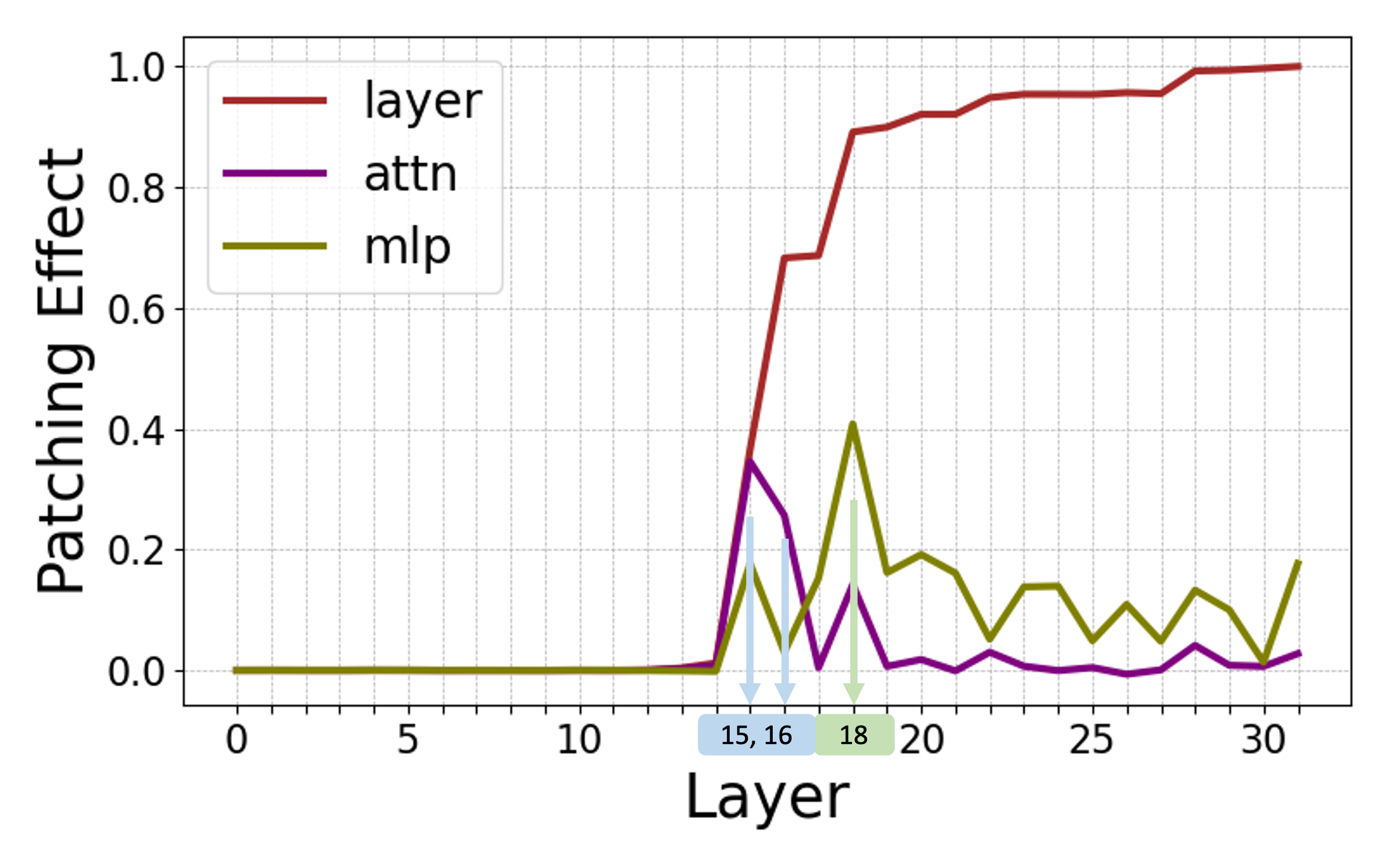}
        \subcaption{Computation}
    \end{minipage}

    \caption{Visualizations of internal computations at last token position in Llama-3 8B for addition math word problems: (i,ii, iv, v) for logit attribution results where $a,b$ are operands and $c$ is the result; (iii, vi) for activation patching results. We label the starting layer of abstraction, operand moving and computation in pink, blue and green, respectively.}
    \label{fig:hypo_gen_results}
\end{figure*}

\subsubsection{Methods}  
We use logit attribution \cite{nostalgebraist2020, belrose2023eliciting} and activation patching \cite{ghandeharioun2024patchscopes, zhang2023towards, meng2022locating} to probe whether abstraction and computation occur during single-step generation. As summarized in Figure~\ref{fig:interp_method}, we seek evidence of abstraction and computation.

\paragraph{Logit Attribution} 
We use logit attribution to examine specific information (e.g., operator or answer tokens) at each layer (See Figure~\ref{fig:interp_method}a for illustration). Specifically, we compute \textit{direct logit attribution}~\cite{nostalgebraist2020, belrose2023eliciting} of a target token \( t \) by projecting hidden states at various points in each layer onto the vocabulary space: $
\text{logit}(t) = \langle W_U[t], \text{LN}(h) \rangle,
$ where \( h \) is the hidden state, \( LN\) is LayerNorm, and \( W_U[t] \) is the unembedding vector. We probe four points within each layer at the last token position: the attention output, MLP output, and the residual stream immediately after merging the attention output (\textit{resid mid}) and after merging the MLP output (\textit{resid final}). As summarized in Figure~\ref{fig:interp_method}a, we track abstraction via the logits of operator tokens (e.g., ``+'', ``add'', ``addition'') and computation via the logits of operand and answer tokens across layers.

\begin{algorithm}
\caption{Activation Patching}
\label{alg:activation_patching}
\begin{algorithmic}[1]
\STATE \textbf{Input:} Set $\Omega$ of clean and corrupted sample pairs $(X_{cl}, X_{cor})$, model $\mathcal{M}$ with hidden states $\mathcal{S}$.
\STATE \textbf{Output:} Patching effects for $\mathcal{S}$: $E_{\mathcal{S}}$.
\FOR{$(X_{cl}^{(i)}, X_{cor}^{(i)}) \in \Omega$}
    
    \STATE $logit_o,  A_{cl} \leftarrow \mathcal{M}(X_{cl}^{(i)}, A_{cl})$
    \# Clean run:  get clean logits and store all layer activations $A_{cl}$
    \STATE $logit_c,  A_{cor} \leftarrow \mathcal{M}(X_{cor}^{(i)}, A_{cor})$ 
    \# Corrupted run: get corrupted logits and store all layer activations $A_{cor}$
    
    \FOR{$s \in \mathcal{S}$}  
        \STATE $A'_{cor}(s) \leftarrow A_{cl}(s)$ \hfill \hfill \# Patched run: replace hidden state s in $A_{cor}$ by $A_{cl}$
        \STATE $logit_p \leftarrow \mathcal{M}(X_{cor}^{(i)}, A'_{cor})$ \hfill \# get patched logits
        \STATE $e_s^{(i)} \leftarrow \frac{logit_p - logit_c}{logit_o - logit_c}$ \hfill \# patching effect
    \ENDFOR
\ENDFOR
\STATE \textbf{Return:} $E_s\gets \frac{1}{|\Omega|}  \sum_{i=1}^{|\Omega|} e_s^{(i)}$
\end{algorithmic}
\end{algorithm}

\paragraph{Activation Patching}
To identify components causally responsible for abstraction and computation, we apply activation patching (Algorithm~\ref{alg:activation_patching}, see Figure~\ref{fig:interp_method}b for visualization)) \cite{ghandeharioun2024patchscopes, zhang2023towards, meng2022locating}. To quantify the contribution of each component across layers, this method replaces a single intermediate hidden state in the corrupted forward pass with the corresponding hidden state from the clean run and measures how much this single hidden state injected in corrupted forward pass can restore the prediction of the clean answer. This patching effect per state per layer is a normalized score from 0 (no recovery) to 1 (full recovery to clean performance), with higher indicating more contribution. We patch attention, MLP and final layer outputs across layers at the last position. Formally, we quantify causal impact using the logit difference between clean $a_{cl}^{(i)}$ and corrupted answers $a_{cor}^{(i)}$ in Eq.~\ref{eq:patching_effect}.\begin{align}
\text{LD}_*(i) &= \text{logit}_*(a_{\text{cl}}^{(i)}) - \text{logit}_*(a_{\text{cor}}^{(i)}) \\
e_s^{(i)} &= \frac{\text{LD}_p(i) - \text{LD}_c(i)}{\text{LD}_o(i) - \text{LD}_c(i)}
\label{eq:patching_effect}
\end{align}
To probe \textbf{abstraction} (Figure~\ref{fig:interp_method}b), we construct minimally different clean/corrupted pairs that vary in their underlying logic (e.g., “buys” for addition vs. “eats” for subtraction) but have the same numbers (e.g., “5,3”). In Figure~\ref{fig:interp_method}b, the clean input implies $5+3=8$, while the corrupted input implies $5-3=2$. We patch individual clean states to the corrupted run to identify critical layers for restoring the addition logic and recovering the clean answer `8'. For \textbf{computation} (Figure~\ref{fig:interp_method}b), we use pairs with the same logic (e.g., addition), but different numbers (e.g., “5,3” vs. “11,7”). Here, we seek to identify layers whose states when patched individually from the clean run to the corrupted run, are critical to perform the clean-run-specific computation with clean operands $5,3$ and output ``8''.

\subsubsection{Abstract-then-Compute Hypothesis}
As shown in Figure~\ref{fig:hypo_gen_results}, we observe distinct stages for abstraction and computation, supporting the \textit{abstract-then-compute} hypothesis. Logit attribution reveals that around L13–14, attention begins moving the inferred operator (e.g., `+', plus', add') to the last position (Figure~\ref{fig:hypo_gen_results}i, iv). This coincides with a divergence in logit differences between target operators (+' vs. `–') in addition and subtraction problems (Figure~\ref{fig:hypo_gen_results}v), suggesting that while earlier layers encode generic operator features, problem-specific abstraction emerges here. Subsequently, around L15–16, Figure~\ref{fig:hypo_gen_results}i,iv shows operands transfer to the last position; Following abstraction, the computation phase appears to begin at L18, primarily through MLPs layers (Figure~\ref{fig:hypo_gen_results}ii, iv). Activation patching confirms the distinct stages: abstraction starts at around L13, with rising attention and layer patching effects (Figure~\ref{fig:hypo_gen_results}iii); The rise of attention and layer patching effects in L15,16 in Figure~\ref{fig:hypo_gen_results}vi aligns with our previous observation that operands are being moved to the last position. Finally, the peak patching effect of MLP at L18 highlight their crucial role in calculating the answer. These combined results support our hypothesis that the model follows an \textit{abstract-then-compute} mechanism within a single forward pass. Additional logit attribution and activation patching results for other models and two-operator problems are in Appendix~\ref{app: logit_lens_results}.

\begin{figure*}[ht]
    \centering
    \begin{subfigure}[b]{0.24\linewidth}
        \centering
        \includegraphics[width=\linewidth]{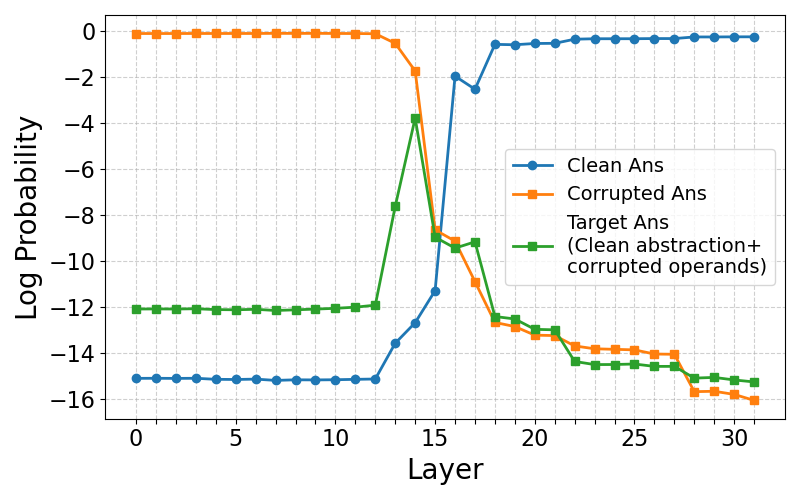}
        \caption{Cl: $a+b$; Cor: $a-b$}
        \label{fig:mult_plot1}
    \end{subfigure}
    \hfill
    \begin{subfigure}[b]{0.24\linewidth}
        \centering
        \includegraphics[width=\linewidth]{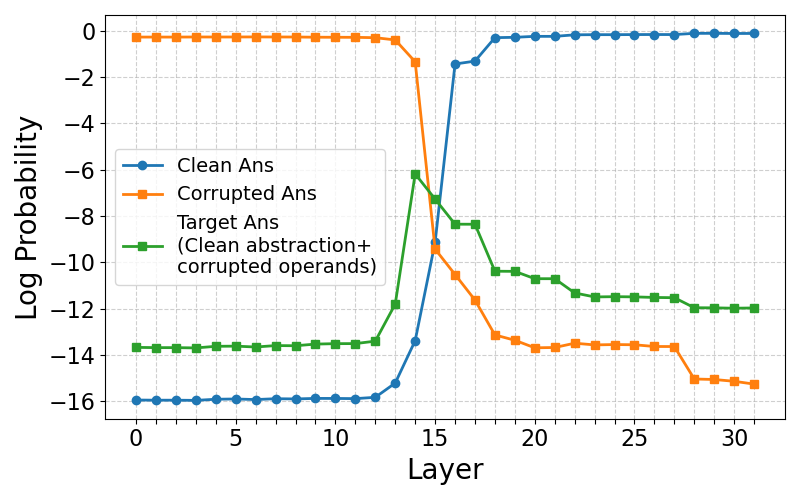}
        \caption{Cl: $a-b$; Cor: $a+b$}
        \label{fig:mult_plot2}
    \end{subfigure}
    \hfill
    \begin{subfigure}[b]{0.24\linewidth}
        \centering
        \includegraphics[width=\linewidth]{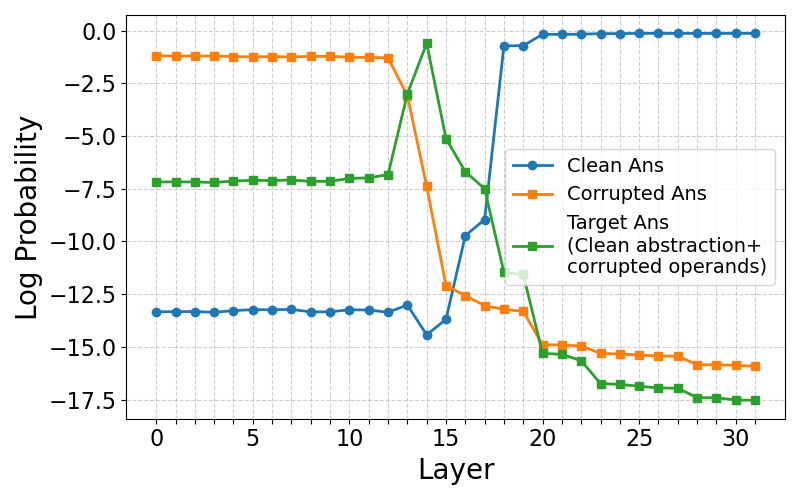}
        \caption{Cl: $a\times b$; Cor: $a\div b$}
        \label{fig:mult_plot3}
    \end{subfigure}
    \hfill
    \begin{subfigure}[b]{0.24\linewidth}
        \centering
        \includegraphics[width=\linewidth]{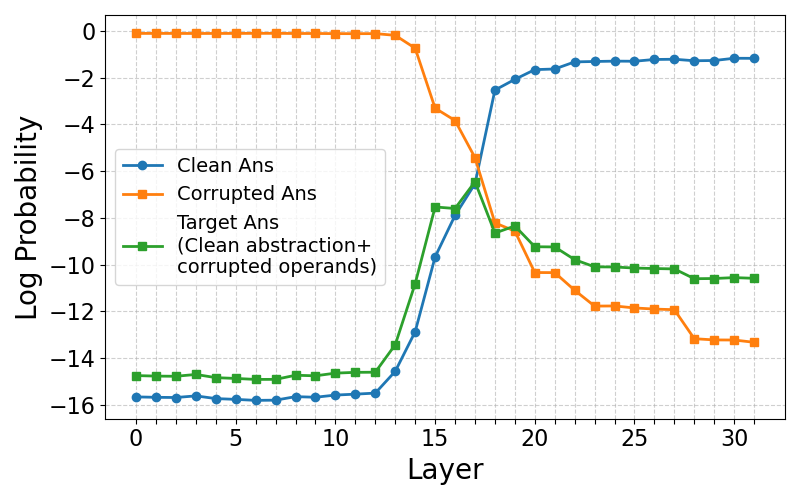}
        \caption{Cl: $a\div b$; Cor: $a\times b$}
        \label{fig:mult_plot4}
    \end{subfigure}
    \begin{subfigure}[b]{0.24\linewidth}
        \centering
        \includegraphics[width=\linewidth]{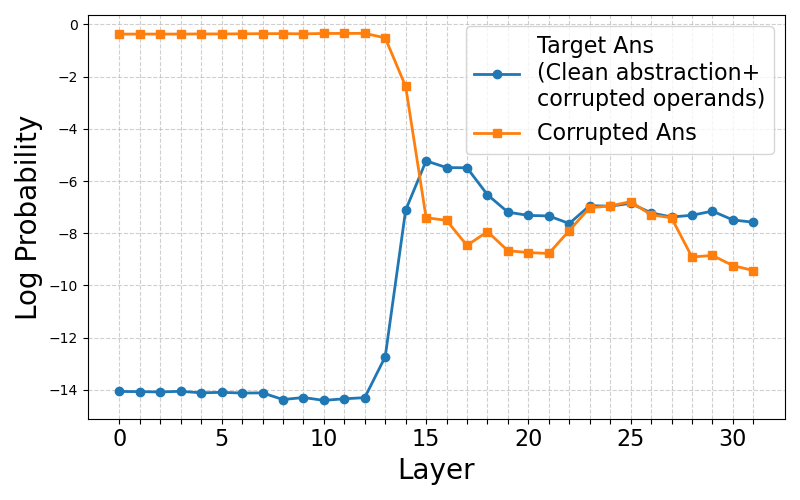}
        \caption{Cl: Paired $x-y$; Cor: $a+b$}
        \label{fig:add_plot1}
    \end{subfigure}
    \hfill
    \begin{subfigure}[b]{0.24\linewidth}
        \centering
        \includegraphics[width=\linewidth]{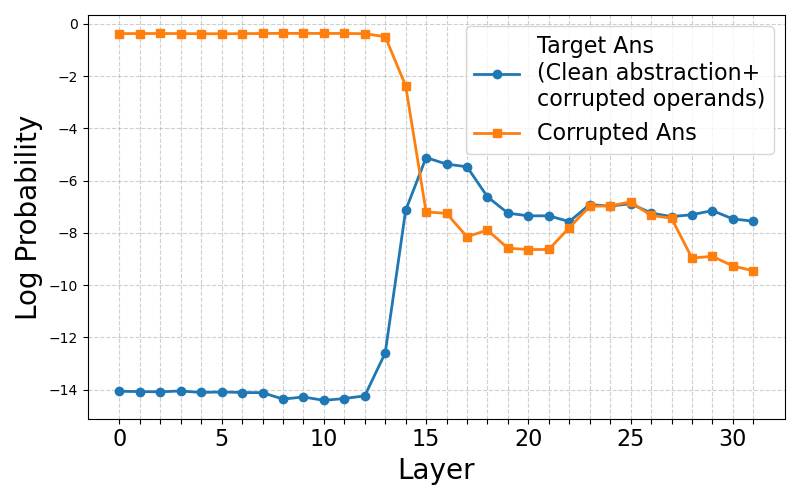}
        \caption{Cl: $x-y$; Cor: $a+b$}
        \label{fig:add_plot2}
    \end{subfigure}
    \hfill
    \begin{subfigure}[b]{0.24\linewidth}
        \centering
        \includegraphics[width=\linewidth]{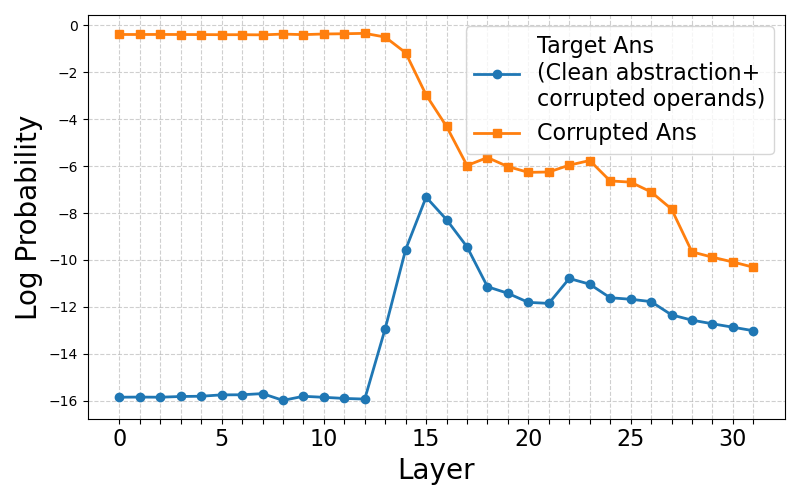}
        \caption{Cl: $x\times y$; Cor: $a+b$}
        \label{fig:add_plot3}
    \end{subfigure}
    \hfill
    \begin{subfigure}[b]{0.24\linewidth}
        \centering
        \includegraphics[width=\linewidth]{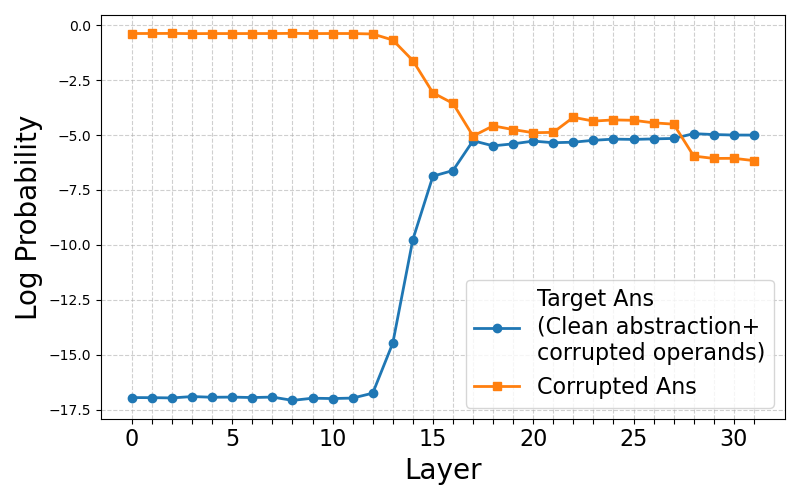}
        \caption{Cl: $x\div y$; Cor: $a+b$}
        \label{fig:add_plot4}
    \end{subfigure}

    \caption{Cross-patching results for Llama-3 8B with corresponding clean and corrupted run. $a,b$ indicate concrete numerical problems, while $x,y$ indicate symbolic problems. \textbf{Top (Numerical Abstraction):} Patching concrete problems with different abstractions shows target log-prob rising at 13 (abstraction onset), peaking at 14 (abstraction formed), then falling as clean operands are introduced. Meanwhile, the clean answer's log-prob rises from 13 (abstraction) and 15 (operand integration), stabilizing at layer 18 (computation). \textbf{Bottom (Symbolic Abstraction):} Patching symbolic problems into concrete addition shows target log-probability rising at layer 13, peaking at 15 (where predictions flip), then declining.}
   
    \label{fig:Llama3-cross-patching}
\end{figure*}

\subsection{Validation and Abstraction Transfer with Cross-Prompt Patching}\label{sec:validation}
We now validate the causal role of the critical layers for abstraction (L13,14) and computation (L15,16 for operands, L18 for execution). We also investigate if the abstraction representations formed at around L13,14 are transferable across problem forms (symbolic/concrete) and templates, and can be composed with subsequent computation stage.


\paragraph{Method} Cross-prompt patching also uses Algorithm~\ref{alg:activation_patching}, but instead of computing patching effects, we track the log-probability of specific tokens across layers in each patched run. This acts as a form of ``knock-out'' intervention: we overwrite a single layer's activations in the corrupted run with clean activations that are hypothesized to contain specific information (e.g., abstraction, operands, computation), and observe whether this information is reflected in the output level. 

To validate the critical layers for each stage, we cross-patch for \textbf{numerical abstraction} (Figure~\ref{fig:interp_method}c), where both the clean and corrupted inputs are numerical problems, but differ in both underlying logic and operands. As shown in Figure~\ref{fig:interp_method}c, the clean run corresponds to $5+3=8$ and the corrupted to $11-7=4$. We patch hidden states from the clean run into the corrupted run and validate our hypothesis: (i) \textit{Abstraction (L13-14)}: At these layers, operands have not been transferred yet, so patching should only transfer the clean logic. If these layers encode addition logic, the model should apply the clean addition operator to the corrupted operands, computing $11+7=18$ in the remaining forward pass. We expect the log-probability of this \textit{target answer} (`18') to rise. (ii) \textit{Operand Transfer (L15–16)}: L15 begins operand transfer and already contains both clean run logic and operand information. Patching them should increase the log-prob of the clean answer (e.g., `8'), while reducing probability of the corrupted (`4') and target answers (`18'). (iii) \textit{Computation (Layer 18)}: By this point, the full ingredients (abstraction and computation) are available. Patching here should fully recover the clean answer (`8'). We expect the log-prob close to 0. To evaluate these effects, we track log-probabilities across layers for the \textbf{target answer} ($18=11+7$, clean logic + corrupted operands) – testing numerical abstraction transfer, the \textbf{clean answer} ($8=5+3$) – testing operand alignment and execution, and the \textbf{corrupted answer} ($4=11-7$).

To investigate if the abstraction representations can be transferred across problem forms (symbolic/numerical) and templates, and if they can be composed with subsequent computation stage, we cross-patch for \textbf{symbolic abstraction} (See Figure~\ref{fig:interp_method}c) -- patching symbolic clean states to numerical corrupted run. Here, clean inputs are symbolic (no concrete numbers), and corrupted inputs are numerical problems with a different underlying logic. This ensures that only abstraction (no operands or 
computation) is transferred from the clean run, unlike numerical abstraction cross-patching. In the example in Figure~\ref{fig:interp_method}c, clean run predicts $x-y$, while the corrupted run corresponds to $5+3=8$. By patching clean states from the symbolic problem to the numerical corrupted run, we examine (i) if \textit{symbolic abstractions are also formed} at around L13-14, despite predicting `x' as the first token, and (ii) if this abstraction ($x-y$), when transferred into numerical corrupted run, can be \textit{used and composed} with corrupted operands ($5,3$) to compute $5-3=2$. To assess this, we track the per-layer log-probabilities of the \textbf{target answer} (clean logic + corrupted operands, $2=5-3$) and \textbf{corrupted answer} ($8=5+3$), and omit the clean answer `x'. If symbolic abstraction transfer occurs, we expect an increase in the target answer log-prob, and a corresponding decrease in the corrupted answer starting around L13. Note that since the symbolic clean states across layers are predicting `x', we expect both answer log-probs to drop.

\paragraph{Results}
Figure~\ref{fig:Llama3-cross-patching}a shows results for \textbf{numerical abstraction} cross-patching results corresponding to the example illustrated in Figure~\ref{fig:interp_method}. As expected, the target answer log-probability (`18') begins rising at L13 (abstraction onset), peaks at L14 (abstraction formed), and drops when clean operands are introduced (L15). The clean answer (`8') log-probability  keeps rising from L13 (abstraction) and continue at 15 (operand integration), stabilizing by L18 (computation). The corrupted answer (`4') log-probability drops after L13. These trends hold across underlying logic (Figure~\ref{fig:Llama3-cross-patching}b-d), confirming the roles of these critical layers as identified earlier. In \textbf{symbolic abstraction} cross-patching (Figure~\ref{fig:Llama3-cross-patching}e-h), we observe consistent behaviour: from L13 onward, the target answer probability increases while the clean answer decreases, eventually flipping. This indicates that (i) abstractions injected via patching are composed with corrupted operands to produce valid outputs, and (ii) abstraction representations at L13–14 are invariant to surface form and problem template. Concretely, comparing Figure~\ref{fig:Llama3-cross-patching}e and Figure~\ref{fig:Llama3-cross-patching}f, where minimally different templates are used in (e) and random templates in (f), we observe near-identical effects in both cases —suggesting abstraction transfer is template-invariant. Furthermore, (g) and (h) show that injecting symbolic \textit{multiplication} and \textit{division} abstractions into concrete \textit{addition} problems still flips the model’s prediction—demonstrating the generality of abstraction transfer. Cross-patching results for other models, and two-operator problems are in Appendix ~\ref{app:cross_patching_results}.

Together, these results provide strong support for the abstract-then-compute hypothesis with critical layers for abstraction (L13,14) and computation (L15,16 for operands and L18 for computation), and further demonstrate that:
(i) abstraction can be transferred and composed with subsequent computation across surface forms (symbolic/concrete) and templates, and
(ii) even at the last position in symbolic problems, when predicting the first output token `x', middle layers already encode abstraction (e.g., the correct operator), indicating that next-token prediction reflects not just immediate token prediction, but also anticipates future outputs.

\section{Conclusion}
Disentangled evaluation reveals that, without CoT, models perform better at abstraction than computation, with the latter bottlenecking final-answer accuracy — challenging the view that poor performance always imply reasoning failure. Mechanistic interpretability uncovers an abstract-then-compute mechanism with transferable abstractions. We argue for disentangled evaluation to more precisely assess model abilities and inform architectural design.

\section{Limitations}
Our study has several limitations. First, we focus solely on English-language datasets; whether the abstract-then-compute mechanism generalizes to other languages remains an open question. Second, our evaluation decomposes mathematical problem-solving into only two stages: abstract formulation and arithmetic computation. Finer-grained breakdowns (e.g., \citet{opedal2024language}) may offer deeper insight. Third, our interpretability analysis is limited to single-step generation, as common techniques (e.g., activation patching, logit attribution) target single-token behavior. Extending these to multi-step reasoning is an ongoing challenge, with recent work like SelfIE \citep{pmlr-v235-chen24ao} provides initial steps. Fourth, while we focus on critical layers involved in abstraction and computation, we leave detailed analysis of components to future work. Finally, due to compute constraints, we analyze models up to 12B parameters. Extending to larger models is left for future studies.

\section*{Acknowledgments} This work is supported by the Mitacs Accelerate Program (Project ID: IT27067) and partially enabled by Mila’s computing resources (mila.quebec). We thank Zichao Li, Cesare Spinoso-Di Piano, and Xiyuan Zou for helpful discussions. Jackie Chi Kit Cheung is supported by the Canada CIFAR AI Chair program. We acknowledge the material support of NVIDIA for providing computational resources.



\bibliography{main}

\appendix
\section{Disentangled Evaluation Details and Additional Results}
\subsection{Symbolic Variant Creation For GSM8K and SVAMP}\label{app: gsm8k-svamp}
\begin{figure}[h!]
    \centering
    \includegraphics[width=1.\linewidth]{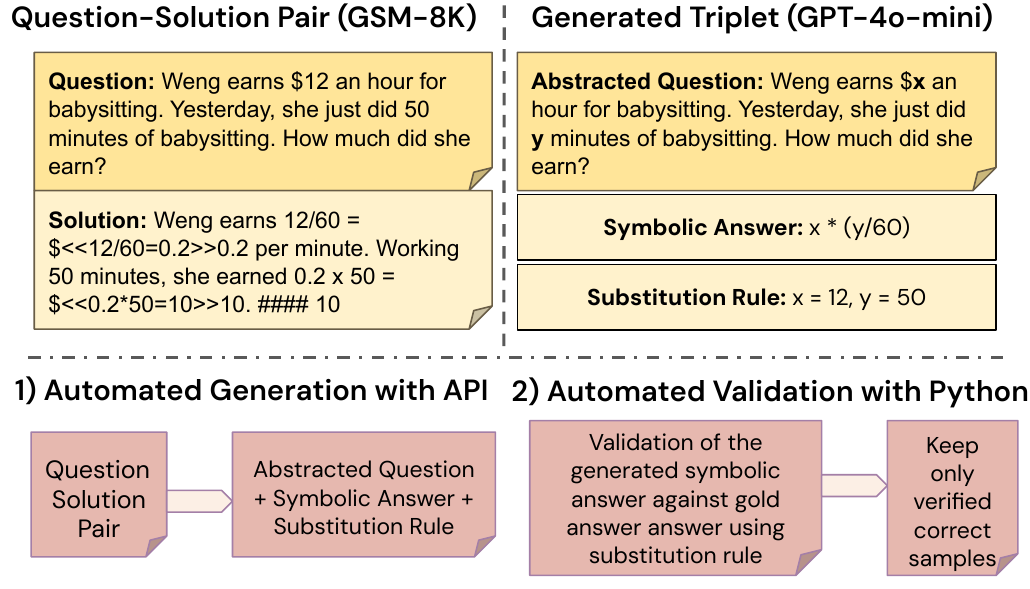}
    \caption{\textbf{Generate-then-validate pipeline}: We use API calls to obtain abstract question-answer-substitution triplets from the concrete question-solution pair from GSM-8K, then validate them against gold answer using \texttt{sympy}. Triplets that fail this check are manully reviewer and corrected.}
    \label{fig:data_construction}
\end{figure}
All our evaluations are conducted on the GSM8K test set and the full SVAMP dataset. To support our evaluation of abstract formulation and arithmetic computation in Section~\ref{sec:eval_results}, we construct symbolic question and expression answer variants for both SVAMP and GSM-8K. Examples are shown in Table~\ref{tab:symbolic_examples}. 

For SVAMP, which already includes both the expression (e.g., 20 × 10) and the final numerical answer (e.g., 200), we create symbolic abstraction variants by replacing all numeric values with symbolic variables (e.g., $x$, $y$) in both the question and the corresponding expression. This preserves the structure and semantics of the original problem while abstracting away from the concrete numbers. For arithmetic computation variant, we use the paired expression.

For GSM-8K, which lacks such annotations, we generate both the symbolic abstraction variant and the numerical expressions using a two-stage generate-then-validate pipeline (Figure~\ref{fig:data_construction}). In the generation stage, we use \texttt{GPT-4o-mini} \cite{openai2024} to produce triplets from original question–solution pairs. Each triplet consists of: (1) a symbolic version of the question, where relevant numbers are replaced with variables while maintaining the semantic content; (2) a symbolic expression that represents the solution in closed-form using those variables; and (3) a substitution rule that maps each variable to its original numeric value. In the validation stage, we verify the correctness of each generated sample. We apply the substitution rule to the symbolic expression, obtaining a numerical expression, then using \texttt{sympy} to evaluate the expression, and compare the resulting numeric answer to the gold answer from GSM-8K. Triplets that fail this check are manually reviewed and corrected. 

\begin{tcolorbox}[float=htb,title=Box 1: Symbolic Evaluation Prompt, colback=gray!5, colframe=gray!80, fonttitle=\bfseries]

\small
Determine whether the following two mathematical expressions are equivalent. The expressions may be written in simplified or unsimplified symbolic form (e.g., \texttt{1/2x + 3}), natural language (e.g., ``Susan made \texttt{1/2x + 3} buttons'') or in LaTeX notation. Consider expressions equivalent if they represent the same mathematical value, even if written differently (e.g., different notation, simplification, or variable order when valid). Respond only with: \texttt{True} or \texttt{False}.\\[0.5em]
\textbf{Example:}\\
1. \( z - (y - x) \)\\
2. \texttt{z - y + x}\\
\textbf{Answer:} \texttt{True}\\[0.5em]

1. \texttt{Susan made 1/2*x buttons}\\
2. \texttt{0.5x}\\
\textbf{Answer:} \texttt{True}\\[0.5em]

1. \texttt{2(y + x)}\\
2. \( M = 2(y + x) \)\\
\textbf{Answer:} \texttt{True}\\[0.5em]

1. \texttt{xz * ((1 - y)/100)}\\
2. \texttt{x × z - (y/100) × (x × z)}\\
\textbf{Answer:} \texttt{True}\\[0.5em]

\textbf{Now evaluate:}\\
1. \{\texttt{symbolic\_gold\_answer}\}\\
2. \{\texttt{abstract\_generated\_answer}\}\\
\textbf{Answer:}
\end{tcolorbox}

\begin{table*}[h]
\centering
\small
\begin{tabular}{@{}llcc@{}}
\toprule
\textbf{Gold Answer} & \textbf{Model Generation} & \textbf{Our Eval} & \textbf{GPT-4o-mini Eval} \\
\midrule
$u*(x+y+z)$ & $xu + yu + zu$ & True & True \\
$x + x*(1/y)$ & $x + (x/y)$ & True & True \\
$0.5(x+yz)$ & $z * (y + 1) * x / 2$ & False & False \\
$(y+z)/x$ & $xz - y = xy$ & False & False \\
$xz*((1-y)/100)$ & $(x * (1 - y/100) * z)$ & False & True \\
$(12/x)*y$ & $y * 12$ & False & True \\
\bottomrule
\end{tabular}
\caption{Comparison of gold answers, model generations, our annotated correctness, and GPT-4o-mini evaluation on a held-out set of 120 samples.}
\label{tab:evaluation_comparison}
\end{table*}

\begin{table*}[h!]
\centering
\small
\begin{tabular}{p{1.5cm}p{7cm} p{3cm} p{2.5cm}}
\toprule
\textbf{Dataset} & \textbf{Symbolic Question} & \textbf{Answer} & \textbf{Substitution} \\
\midrule
GSM8K & I have x liters of orange drink that are y\% water and I wish to add it to z liters of pineapple drink that is u\% water. But as I pour it, I spill v liters of the orange drink. How much water is in the remaining w liters? & $(y \cdot (x - v) + u \cdot z) / 100$ & $x=10,\ y=\tfrac{2}{3},\ z=15,\ u=\tfrac{3}{5},\ v=1,\ w=24$ \\
\addlinespace
GSM8K &Jerry has a flock of chickens. The red chickens produce x eggs a day, and the white chickens produce y eggs a day. Every day Jerry collects z eggs. If he has u more white chickens than red chickens, how many red chickens does he have? & $(z - u \cdot y) / (x + y)$ & $x=3,\ y=5,\ z=42,\ u=2$ \\
\addlinespace
GSM8K &Adrian's age is x times the age of Harriet, and Harriet is y the age of Zack. Calculate the average age of the three in three years if Harriet is z years old now. & $(x*z + z + (z/y) + 9)/3$ & $x=3,\ y=\tfrac{1}{2},\ z=21$ \\
\addlinespace
SVAMP & Each pack of DVDs costs x dollars. If there is a discount of y dollars on each pack & $x - y$ & $x=76,\ y=25$ \\
\addlinespace
SVAMP & An industrial machine worked for x minutes. It can make y shirts a minute. & $x \cdot y$ & $x=4,\ y=5$ \\
\addlinespace
SVAMP & Paco had x salty cookies and y sweet cookies. He ate z sweet cookies and u salty cookies. How many salty cookies did Paco have left? & $x - u$ & $x=26,\ y=17,\ z=14,\ u=9$ \\
\bottomrule
\end{tabular}
\caption{Constructed symbolic examples from GSM8K and SVAMP datasets.}
\label{tab:symbolic_examples}
\end{table*}

\subsection{Evaluation Details}\label{app:eval_details}
In this section, we detail the evaluation of the four settings. First, we show the instructions used in each settings in Table~\ref{tab:experiments} with and without CoT. The prompt used in each setting is then a concatenation of the instruction and the question. 

\begin{table*}[h!]
\centering
\scriptsize
\caption{Prompting Strategies, Problem Variants and Instructions}
\label{tab:experiments}
\renewcommand{\arraystretch}{1.15}
\begin{tabular}{p{3cm}p{1cm}p{4.5cm}p{3.3cm}p{1.5cm}}
\toprule
\textbf{Setting} & \textbf{Strategy} & \textbf{Instruction} & \textbf{Question} & \textbf{Answer}  \\
\midrule
Original & No CoT & Please answer the question directly WITHOUT showing the reasoning process, you MUST write the answer as \textbf{an integer} after `\#\#\#\#', without including the equation or units. & Weng earns \$12 an hour for babysitting. Yesterday, she just did 50 minutes of babysitting. How much did she earn? & 10 \\
Original & CoT & \textbf{Let's think step by step}, you MUST write the answer as \textbf{an integer} after `\#\#\#\#' without including the units. Write the answer at the end. & Weng earns \$12 an hour for babysitting. Yesterday, she just did 50 minutes of babysitting. How much did she earn? & 10 \\
\midrule
Arithmetic Computation & No CoT & Please answer the question directly WITHOUT showing the reasoning process, you MUST write the answer as \textbf{an integer} after `\#\#\#\#' & What is the value of 12 * (50/60)? & 10  \\
Arithmetic Computation & CoT& \textbf{Let's think step by step}, you MUST write the answer as \textbf{an integer} after `\#\#\#\#' . Write the answer at the end. & What is the value of 12 * (50/60)? & 10  \\
\midrule
Numerical Abstraction & No CoT & Please answer the question directly without showing the reasoning process, you MUST write the \textbf{expression} with appropriate round brackets after `\#\#\#\#', without including the units, and you DO NOT need to simplify the expression. & Weng earns \$12 an hour for babysitting. Yesterday, she just did 50 minutes of babysitting. How much did she earn? & $12 * (50/60)$  \\
Numerical Abstraction & CoT  & \textbf{Let's think step by step}, at the end, you MUST write the \textbf{expression} with appropriate parenthesis after `\#\#\#\#', without including the units, but you DO NOT need to simplify the expression. & Weng earns \$12 an hour for babysitting. Yesterday, she just did 50 minutes of babysitting. How much did she earn? & $12 * (50/60)$ \\
\midrule
Symbolic Abstraction & No CoT &  Please answer the question directly WITHOUT showing the reasoning process, you MUST write the \textbf{expression} with appropriate round brackets after `\#\#\#\#' without including the units, and you DO NOT need to simplify the expression. & Weng earns \$x an hour for babysitting. Yesterday, she just did y minutes of babysitting. How much did she earn? & $x * (y/60)$\\
Symbolic Abstraction & CoT & \textbf{Let's think step by step}, at the end, you MUST write the \textbf{expression} with appropriate round brackets after `\#\#\#\#' without including the units, but you DO NOT need to simplify the expression. & Weng earns \$x an hour for babysitting. Yesterday, she just did y minutes of babysitting. How much did she earn? & $x * (y/60)$\\
\bottomrule
\end{tabular}
\end{table*}

For the \textit{Original} and \textit{Arithmetic Computation} settings, where the expected output is a final integer answer, we extract the answer following the token \texttt{\#\#\#\#}, remove any accompanying units, and normalize formatting (e.g., removing commas, dollar signs, percentage symbols, and units like `g`) before comparing it with the gold answer.

For the \textit{Numerical Abstraction} setting, where answers are expected to be \textit{numerical expressions}, we first convert LaTeX-style expressions to Python syntax (when written in Markdown form), then evaluate them using \texttt{sympy} to check equivalence with the gold expression.

In the \textit{Symbolic Abstraction} setting, where outputs are \textit{symbolic expressions}, we use \texttt{gpt-4o-mini} as an automated evaluator. The prompting to \texttt{gpt-4o-mini} is shown in Box 1, and responses are generated with temperature set to 0. To validate this method, we annotated a held-out set of 120 samples manually for correctness, and compared our annotations with the \texttt{gpt-4o-mini} 
evaluator’s decisions. We find that \texttt{gpt-4o-mini} achieves \textbf{94\% agreement} with our judgment in identifying symbolic expression equivalence. Example comparisons are shown in Table~\ref{tab:evaluation_comparison}.


\subsection{Additional Result of Disentangled Evaluation Without CoT}\label{app:additional_eval_results}

\begin{figure*}[h]
    \centering
    \includegraphics[width=\linewidth]{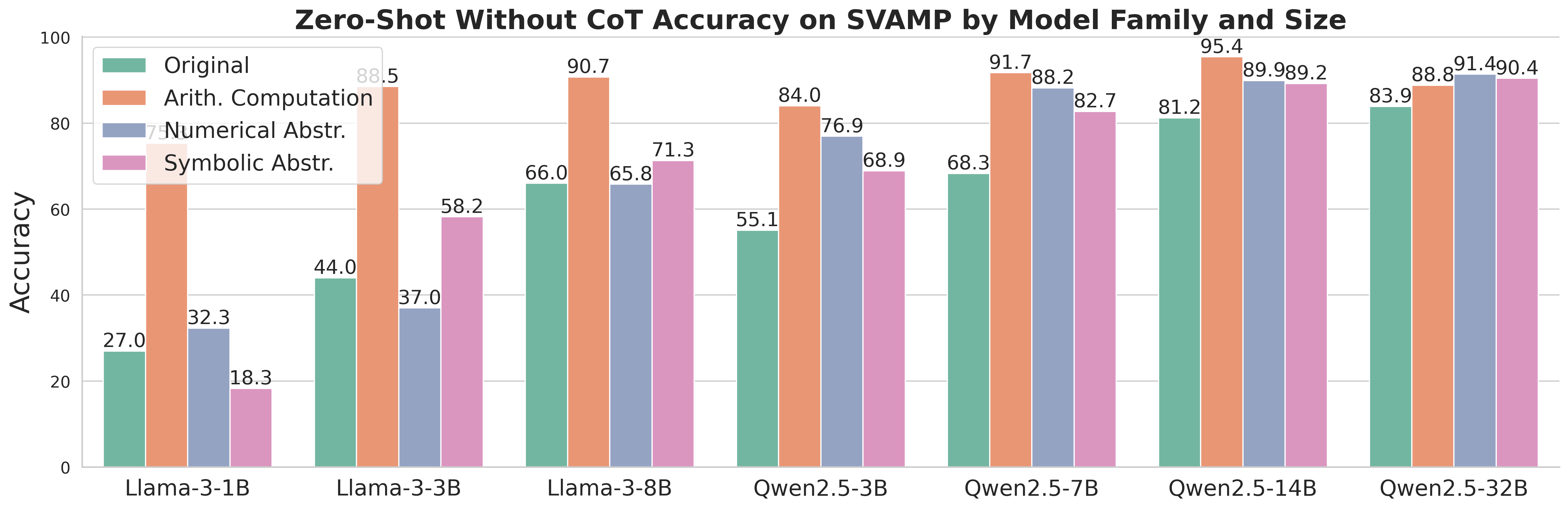}
    \caption{Model zero-shot \textbf{without CoT} performance on SVAMP.}
    \label{fig:computation_zs_svamp}
\end{figure*}

We report zero-shot, no-CoT performance on SVAMP in Figure~\ref{fig:computation_zs_svamp}. Compared to GSM8K, SVAMP is a significantly simpler benchmark consisting of math word problems that require only a single reasoning step --- namely, a single arithmetic operation. As with GSM8K, models perform better on the abstraction variants than in the original setting, though the performance gap is smaller due to the task's simplicity.

Interestingly, we observe a notable difference from GSM8K: across all model sizes, even small models such as \textsc{LLaMA}~1B and 3B perform well on the \textit{Arithmetic Computation} variant, often outperforming both the abstraction variants and the original setting. This suggests that computing one-step expressions (e.g., $5 - 3$) is less challenging than deriving an abstract formulation with only one step. However, in tasks involving multiple steps, abstraction becomes comparatively easier than executing the full computation correctly, as shown in the case of GSM8K. This highlights how model capabilities depend not just on the skill type but also on the complexity of the required operation.


\begin{figure}[th]
    \centering
    \includegraphics[width=1.0\linewidth]{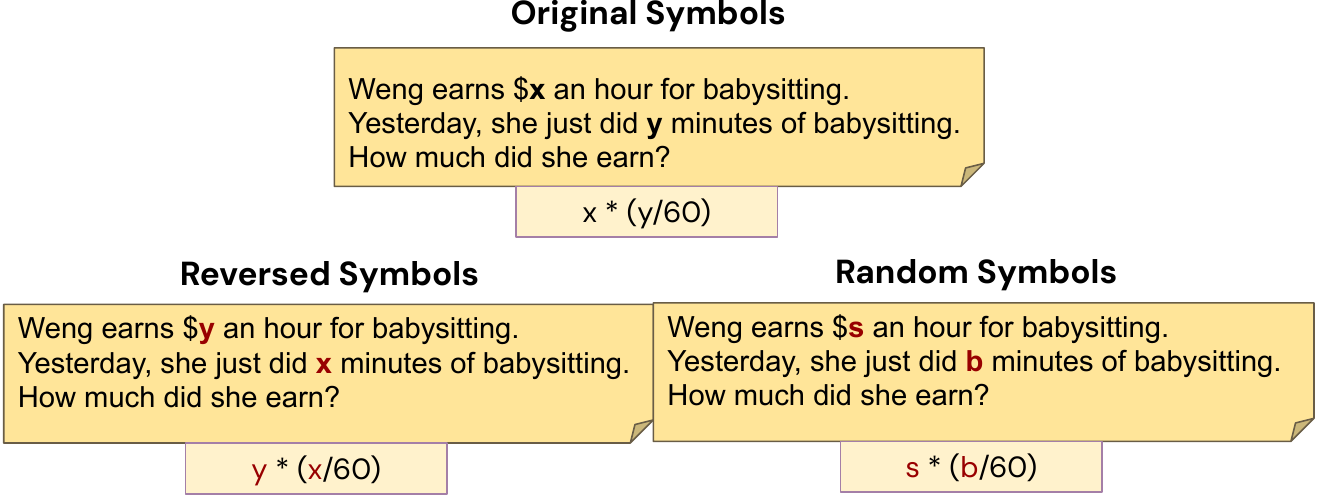}
    \caption{Experiment configurations for the ablation study on symbol choices and symbol order.}
    \label{fig:symbol_abl}
\end{figure}

\subsection{Additional Resuls of Disentangled Evaluation With CoT}\label{app:cot_ablation}
\begin{figure*}[h]
    \centering
    \includegraphics[width=\linewidth]{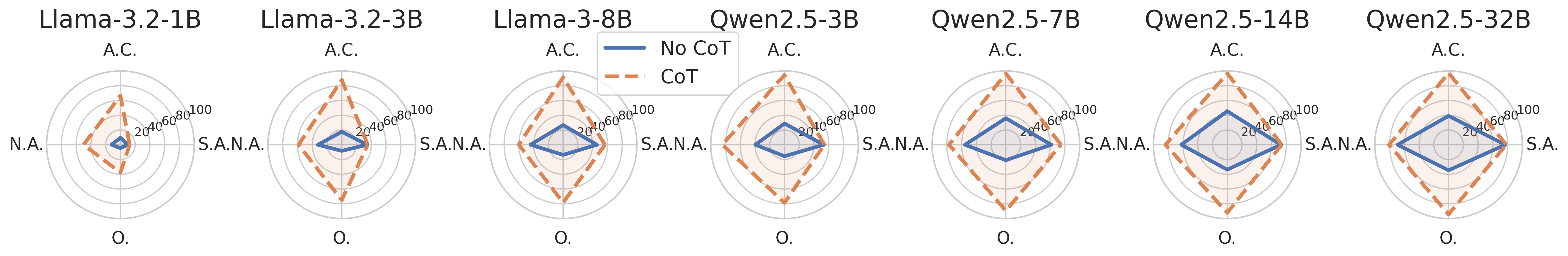}
    \caption{Model zero-shot \textbf{with and without CoT} performance on GSM8K. A.C.: Arithmetic Computation; N.A.: Numerical Abstraction; O.: Original; S.A.: Symbolic Abstraction.}
    \label{fig:gsm8k_cot_effect}
\end{figure*}

\begin{figure*}[h]
    \centering
    \includegraphics[width=\linewidth]{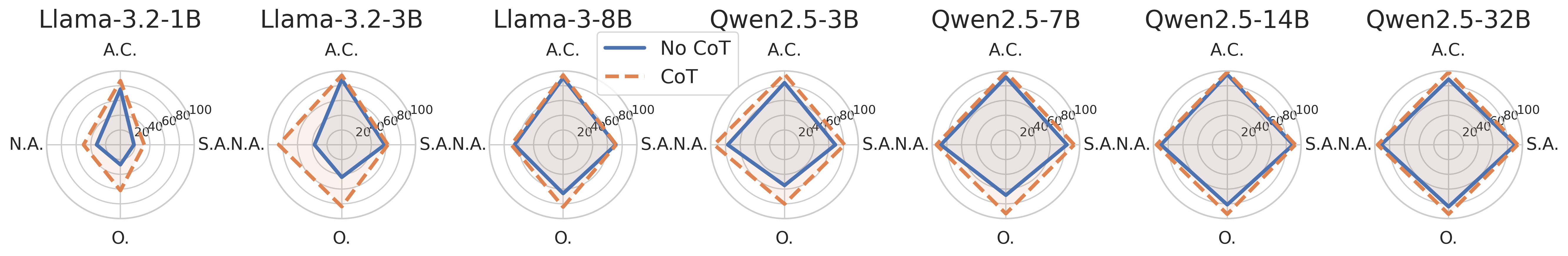}
    \caption{Model zero-shot \textbf{with and without CoT} performance on SVAMP. A.C.: Arithmetic Computation; N.A.: Numerical Abstraction; O.: Original; S.A.: Symbolic Abstraction.}
    \label{fig:svamp_cot_effect}
\end{figure*}

We present the full results on GSM8K for Llama family and Qwen family in Figure~\ref{fig:gsm8k_cot_effect}, and full results on SVAMP for Llama family and Qwen family in Figure~\ref{fig:svamp_cot_effect}.

\subsection{Ablation on Symbolic Abstraction Variant}\label{app:symbolic_ablation}
To assess the reliability and external validity of the symbolic abstraction evaluation, we perform ablations over symbol order and symbol choice. As illustrated in Figure~\ref{fig:symbol_abl}, we compare three settings: \begin{itemize}
    \item \textbf{Original Symbols}: Variables are consistently represented using a fixed set of letters in order—$x$, $y$, $z$, $u$, $v$, $w$, $p$, $q$, $r$, $s$, $t$—e.g., $x \times (y/60)$.
    
    \item \textbf{Reversed Symbols}: The same set of symbols is used, but the order is reversed (e.g., $y \times (x/60)$), preserving the semantic and structural content of the problem while changing the superficial presentation.
    
    \item \textbf{Random Symbols}: Each original symbol is replaced with a randomly sampled letter from the alphabet, unique to each dataset. This preserves the structure of the expression while removing any consistent identity cues. The mappings are as follows: \texttt{\{'a': 'h', 'd': 'i', 'm': 's', 'n': 'r', 'p': 'e', 'q': 'l', 'r': 'c', 's': 'v', 't': 'j', 'u': 'm', 'v': 't', 'w': 'o', 'x': 'u', 'y': 'p', 'z': 'b', 'Z': 'f'\}}
\end{itemize}

In Table~\ref{tab:symbolic_ablation_gsm8k}, we observe mild performance degradation with symbol perturbations on both models, (e.g., three-point drop with Reversed and another two points with Random), but models retain strong accuracy compared to the Original setting. This suggests that Symbolic Abstraction is relatively robust to surface-level symbol changes. 


\begin{table}[h!]
\centering
\small
\begin{tabular}{@{}lcc|cc@{}}
\toprule
\textbf{Setting} & \multicolumn{2}{c|}{\textbf{Llama 8B}} & \multicolumn{2}{c}{\textbf{Qwen 7B}} \\
                 & \textbf{No CoT} & \textbf{CoT} & \textbf{No CoT} & \textbf{CoT} \\
\midrule
original         & 45.7 & 56.7 & 61.5 & 74.7 \\
\midrule
reverse          & 42.8 & 51.8 & 61.9 & 74.8 \\
random           & 41.0 & 53.1 & 58.0 & 71.9 \\
\bottomrule
\end{tabular}
\caption{Results of ablation study on symbol choices and symbol order, with and without CoT under zero-shot setting on GSM8K.}
\label{tab:symbolic_ablation_gsm8k}
\end{table}

\begin{table*}
\centering
\small
\begin{tabular}{p{1.5cm}p{14cm}}
\toprule
\textbf{Subset} & \textbf{Example Data} \\
\midrule
$(+, -)$ & $(+)$ [name] owns {x} stuffed animals. A relative sends them {y} more stuffed animals. How many stuffed animals does [name] have now? \newline  $(-)$ [name] owns {x} stuffed animals. They give {y} stuffed animals to a younger sibling. How many stuffed animals does [name] have now?\\
\addlinespace

$(+, -)$ & $(+)$ [name] finds {x} seashells at the beach. The next day they find {y} more seashells. How many seashells does [name] have now? \newline $(-)$ [name] finds {x} seashells at the beach. The tide washes away {y} seashells. How many seashells does [name] have now? \\
\addlinespace
$(+, -)$ & $(+)$ The storage has {x} gigabytes free. [name] saves {y} gigabytes of photos. How much space remains? \newline $(-)$ The storage has {x} gigabytes free. Cloud storage adds {y} gigabytes. What is the new capacity?\\
\addlinespace

\midrule
\addlinespace
$(\times, \div)$ & $(\times)$ The glacier recedes {x} inches daily. How much will it shrink after {y} days? \newline $(\div)$ The glacier retreated {x} inches over {y} days. What was the average daily recession?\\

\addlinespace
$(\times, \div)$ & $(\times)$ Each server rack uses x kilowatts. What's the total power for y racks? \newline $(\div)$ The data center used x kilowatts across y racks. What was the average per rack?\\

\addlinespace
$(\times, \div)$ & $(\times)$ The spaceship's shield blocks {x} radiation units hourly. How much radiation can it block in {y} hours? \newline $(\div)$ The shield blocked {x} units over {y} hours. What was its average protection rate?\\

\midrule

Two operations & $(x + y + z)$ [name] collects {x} stamps, buys {y} more, and inherits {z}. Total stamps?
\newline $(x + y - z)$ [name] has {x} stamps, acquires {y} more, but loses {z}. How many left?
\newline $(x - y + z)$ [name] owns {x} stamps, sells {y}, but trades for {z}. How many now?
\newline $(x - y - z)$ [name] has {x} stamps, donates {y}, and ruins {z}. How many remain?\\



\bottomrule
\end{tabular}
\caption{Interpretability dataset examples.}
\label{tab:interpretability_data_examples}
\end{table*}

\section{Intepretability Results}
\subsection{Interpretability Data Construction}\label{sec:data_interpretability}
To construct a dataset suitable for mechanistic interpretability, we focus on simpler math word problems that require only one or two reasoning steps with one or two basic arithmetic operations (addition, subtraction, multiplication, or division). We deliberately avoid more complex multi-step problems, as model performance on such tasks tends to be poor, potentially confounding interpretability analyses.

For each pair of arithmetic operations—\((x + y, x - y)\) and \((x \times y, x \div y)\) and \((x + y +z, x + y -z, x - y +z, x - y -z)\) — we use a proprietary model to generate 150 template pairs, totaling 1200 templates. These templates are minimally different in semantics but vary across a broad range of topics, verb choices, names, and syntactic structures. Examples are presented in Table~\ref{tab:interpretability_data_examples}. For instance, a representative pair might include:

\begin{itemize}
    \item \texttt{[name]} has \(\{x\}\) apples. They get \(\{y\}\) more apples. How many apples does \texttt{[name]} have now? \hfill (corresponding to \(x + y\))
    \item \texttt{[name]} has \(\{x\}\) apples. They give away \(\{y\}\) apples. How many apples does \texttt{[name]} have now? \hfill (corresponding to \(x - y\))
\end{itemize}

Each template is instantiated by replacing the \texttt{[name]} placeholder with a randomly selected name from a curated list of 30 English first names, shown below:

\begin{quote}
\texttt{James}, \texttt{Emma}, \texttt{William}, \texttt{Olivia}, \texttt{Benjamin}, \texttt{Charlotte}, \texttt{Henry}, \texttt{Amelia}, \texttt{Alexander}, \texttt{Ava}, \texttt{Samuel}, \texttt{Sophia}, \texttt{Jacob}, \texttt{Mia}, \texttt{Daniel}, \texttt{Lily}, \texttt{Michael}, \texttt{Grace}, \texttt{Ethan}, \texttt{Ella}, \texttt{Jack}, \texttt{Chloe}, \texttt{Lucas}, \texttt{Harper}, \texttt{Thomas}, \texttt{Zoe}, \texttt{Matthew}, \texttt{Nora}, \texttt{Nathan}, \texttt{Isla}.
\end{quote}

The numerical placeholders \(\{x\}\) and \(\{y\}\) are populated with integers \(\leq 50\), to avoid detokenization issues during model processing.

\subsection{Logit Attribution and Activation Patching Additional Results}\label{app: logit_lens_results}
\paragraph{Other Models} 
We observe a similar \textit{abstract-then-compute} mechanism in other models, including Qwen 2.5 7B and Qwen 2.5 14B. In Qwen 2.5 7B, the abstraction stage occurs around layers 18--20, with the computation stage beginning around layers 22--23. In Qwen 2.5 14B, abstraction takes place around layers 29--32, followed by computation starting at layer 36.

\begin{figure*}[ht]
    \centering
    \begin{minipage}{0.32\textwidth}
        \includegraphics[width=\linewidth]{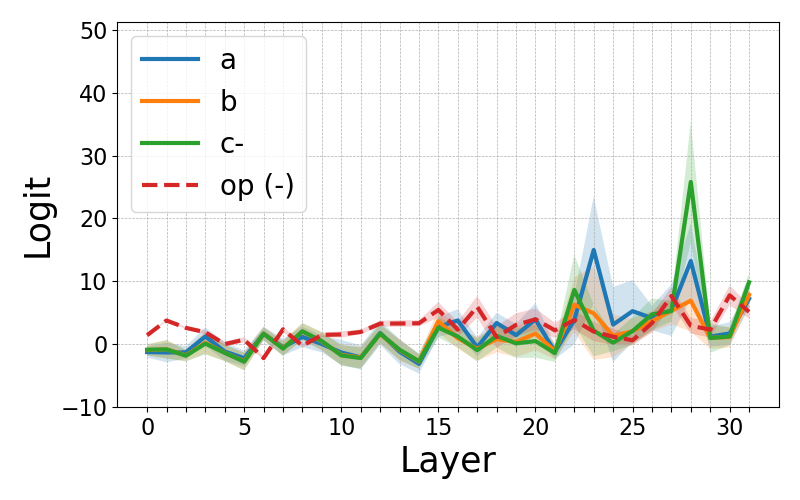}
        \subcaption{Attn}
    \end{minipage}
    \begin{minipage}{0.32\textwidth}
        \includegraphics[width=\linewidth]{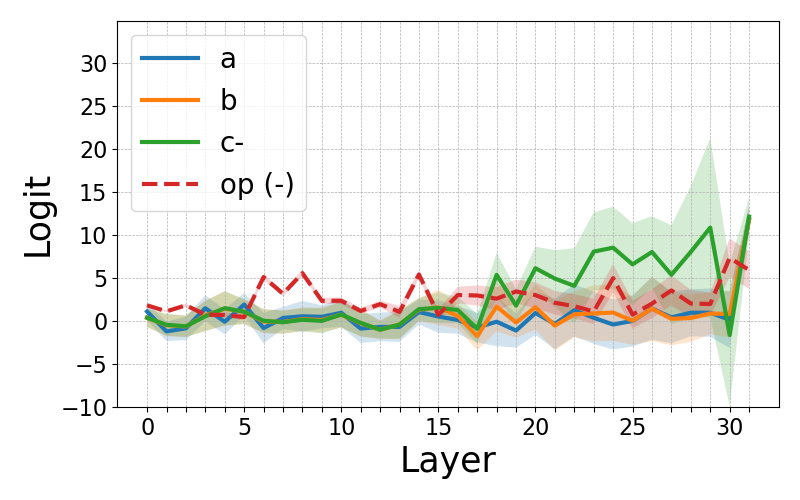}
        \subcaption{MLP}
    \end{minipage}
    \begin{minipage}{0.32\textwidth}
        \includegraphics[width=\linewidth]{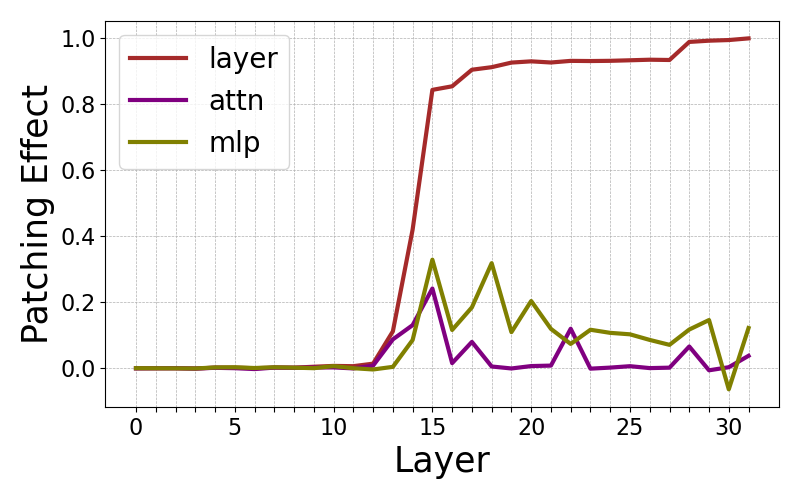}
        \subcaption{Abstraction}
    \end{minipage}


    \begin{minipage}{0.32\textwidth}
        \includegraphics[width=\linewidth]{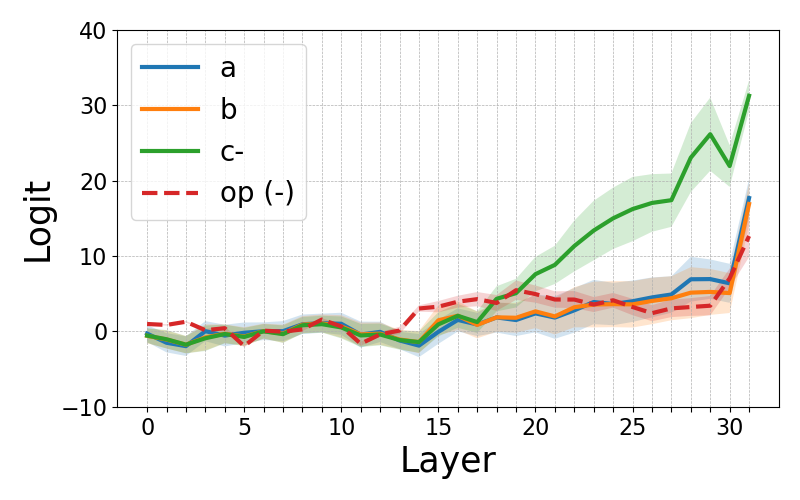}
        \subcaption{Block}
    \end{minipage}
    \begin{minipage}{0.32\textwidth}
        \includegraphics[width=\linewidth]{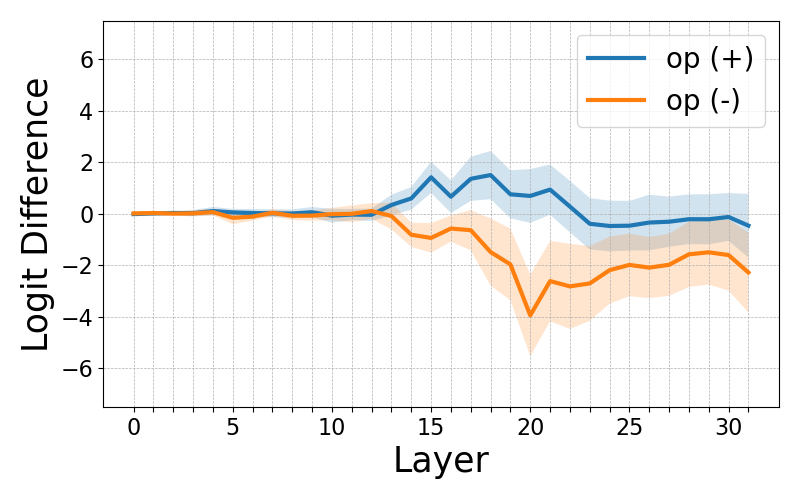}
        \subcaption{Resid Mid}
    \end{minipage}
    \begin{minipage}{0.32\textwidth}
        \includegraphics[width=\linewidth]{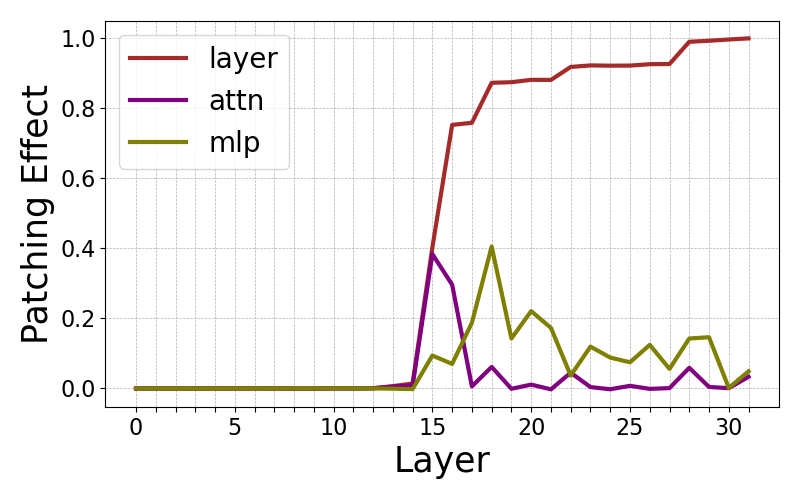}
        \subcaption{Computation}
    \end{minipage}

    \caption{Visualizations of internal computations at last token position in \textbf{Llama-3 8B} for \textbf{subtraction} math word problems: (a, b, d, e) for logit attribution results, (c, d) activation patching for results.}
    \label{fig:llama-subtraction}
\end{figure*}

\begin{figure*}[ht]
    \centering
    \begin{minipage}{0.32\textwidth}
        \includegraphics[width=\linewidth]{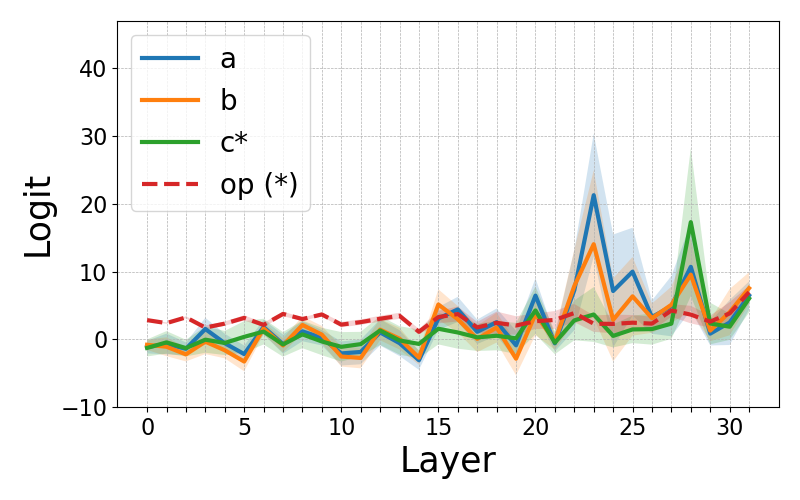}
        \subcaption{Attn}
    \end{minipage}
    \begin{minipage}{0.32\textwidth}
        \includegraphics[width=\linewidth]{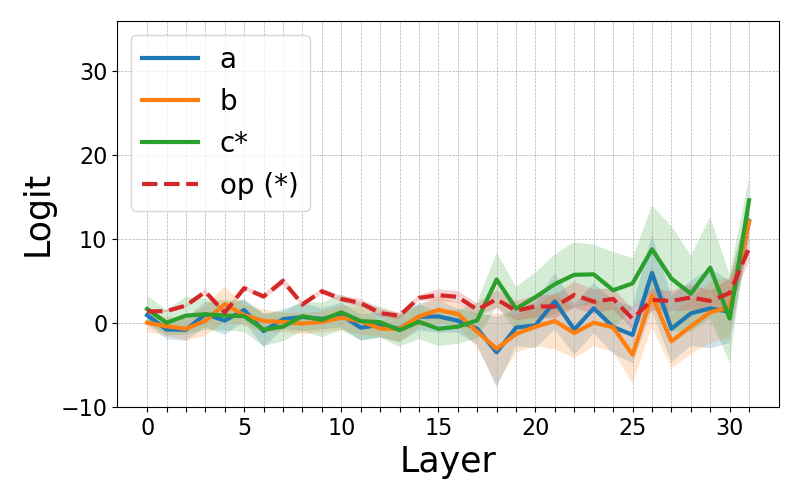}
        \subcaption{MLP}
    \end{minipage}
    \begin{minipage}{0.32\textwidth}
        \includegraphics[width=\linewidth]{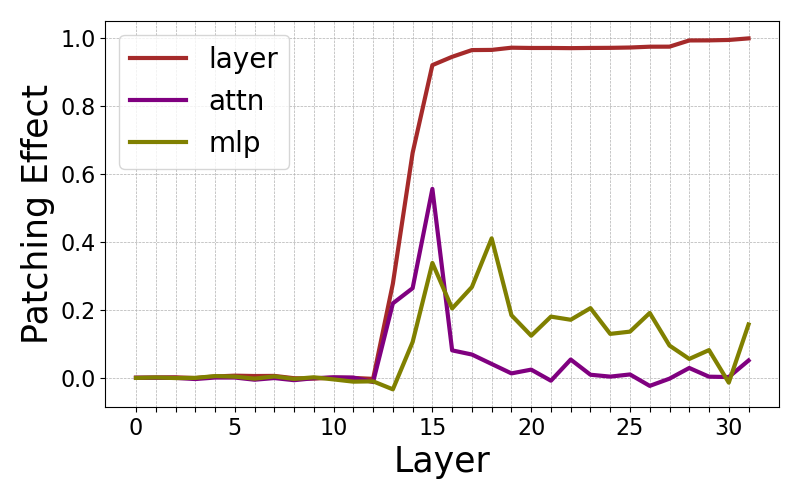}
        \subcaption{Abstraction}
    \end{minipage}


    \begin{minipage}{0.32\textwidth}
        \includegraphics[width=\linewidth]{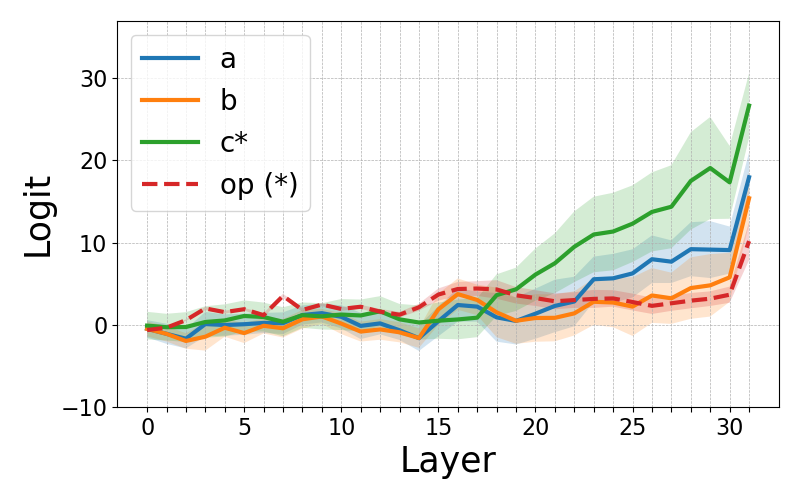}
        \subcaption{Block}
    \end{minipage}
    \begin{minipage}{0.32\textwidth}
        \includegraphics[width=\linewidth]{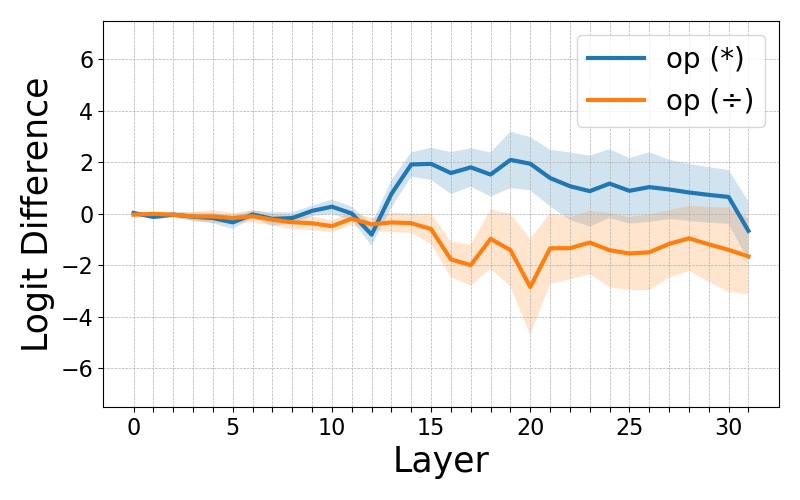}
        \subcaption{Resid Mid}
    \end{minipage}
    \begin{minipage}{0.32\textwidth}
        \includegraphics[width=\linewidth]{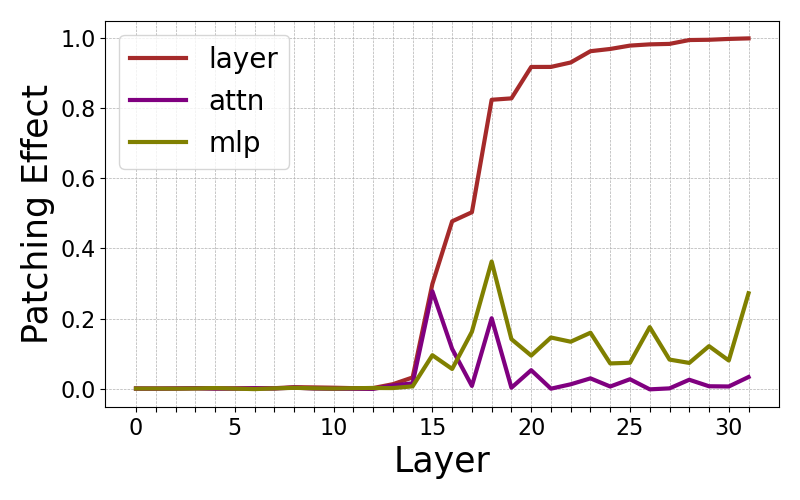}
        \subcaption{Computation}
    \end{minipage}

    \caption{Visualizations of internal computations at last token position in \textbf{Llama-3 8B} for \textbf{multiplication} math word problems: (a, b, d, e) for logit attribution results, (c, d) activation patching for results.}
    \label{fig:llama-multiplication}
\end{figure*}

\begin{figure*}[ht]
    \centering
    \begin{minipage}{0.32\textwidth}
        \includegraphics[width=\linewidth]{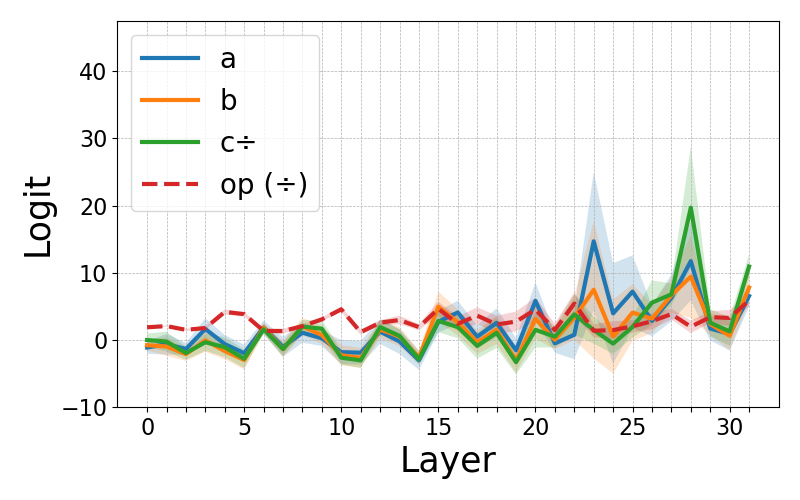}
        \subcaption{Attn}
    \end{minipage}
    \begin{minipage}{0.32\textwidth}
        \includegraphics[width=\linewidth]{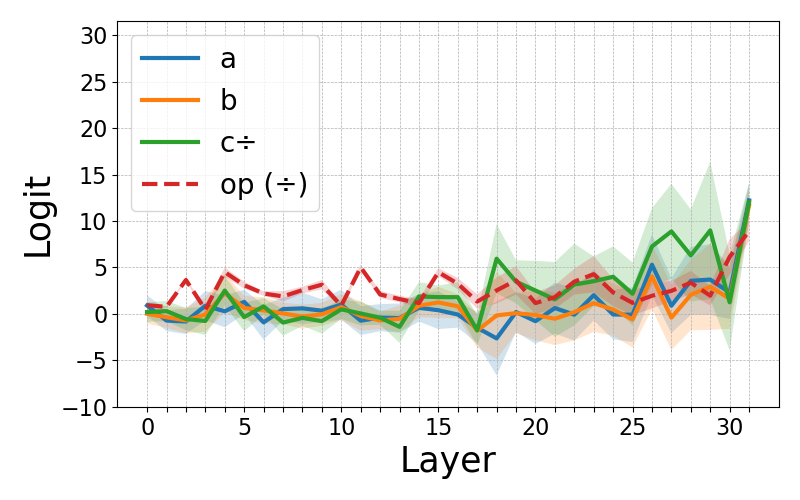}
        \subcaption{MLP}
    \end{minipage}
    \begin{minipage}{0.32\textwidth}
        \includegraphics[width=\linewidth]{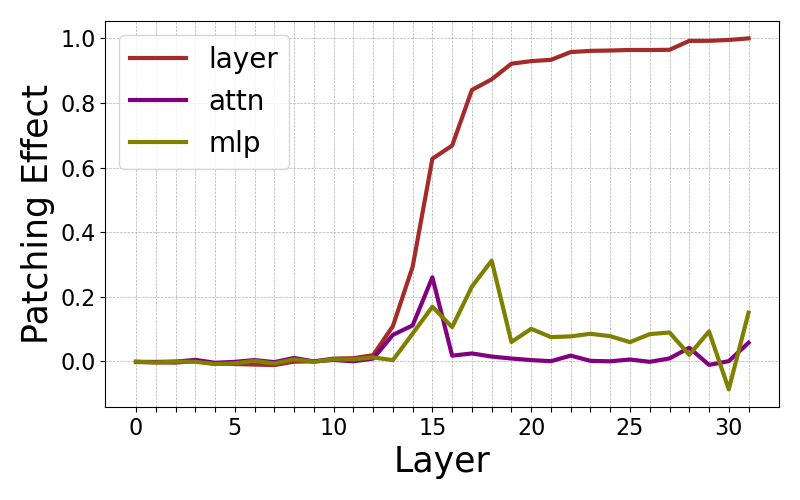}
        \subcaption{Abstraction}
    \end{minipage}


    \begin{minipage}{0.32\textwidth}
        \includegraphics[width=\linewidth]{images/multiplication_resid_final_avg_llama-8b-instruct.png}
        \subcaption{Block}
    \end{minipage}
    \begin{minipage}{0.32\textwidth}
        \includegraphics[width=\linewidth]{images/multiplication-division_resid_mid_diff_llama-8b-instruct.png}
        \subcaption{Resid Mid}
    \end{minipage}
    \begin{minipage}{0.32\textwidth}
        \includegraphics[width=\linewidth]{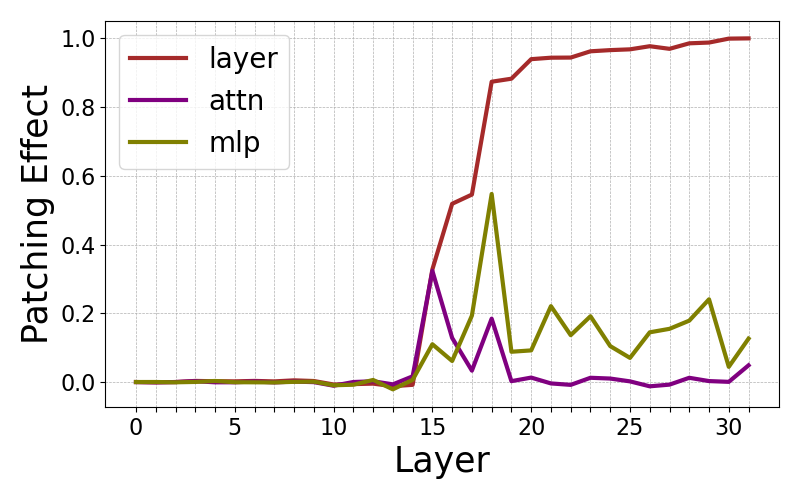}
        \subcaption{Computation}
    \end{minipage}

    \caption{Visualizations of internal computations at last token position in \textbf{Llama-3 8B} for \textbf{division} math word problems: (a, b, d, e) for logit attribution results, (c, d) activation patching for results.}
    \label{fig:llama-division}
\end{figure*}

\begin{figure*}[ht]
    \centering
    \begin{minipage}{0.32\textwidth}
        \includegraphics[width=\linewidth]{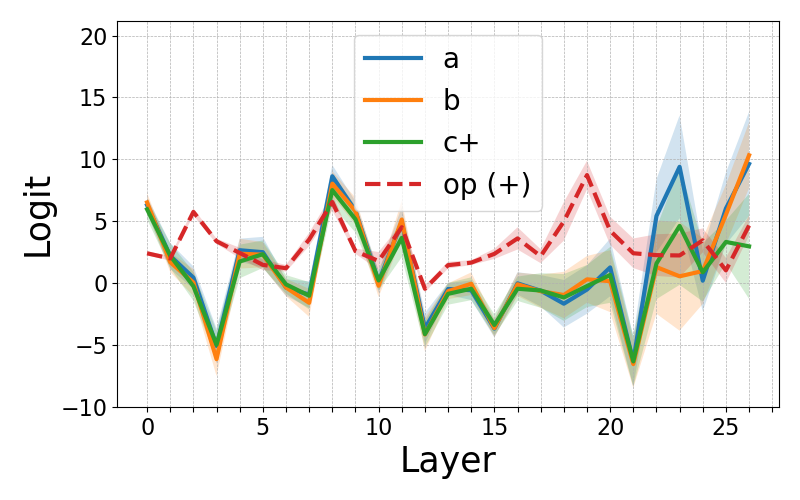}
        \subcaption{Attn}
    \end{minipage}
    \begin{minipage}{0.32\textwidth}
        \includegraphics[width=\linewidth]{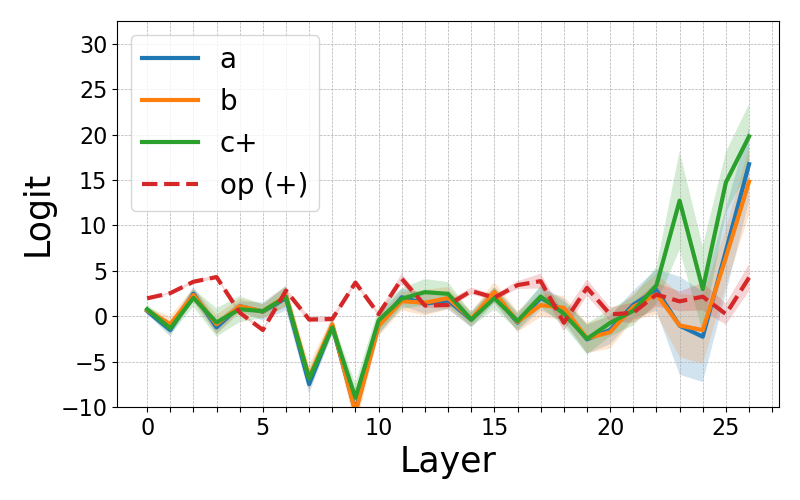}
        \subcaption{MLP}
    \end{minipage}
    \begin{minipage}{0.32\textwidth}
        \includegraphics[width=\linewidth]{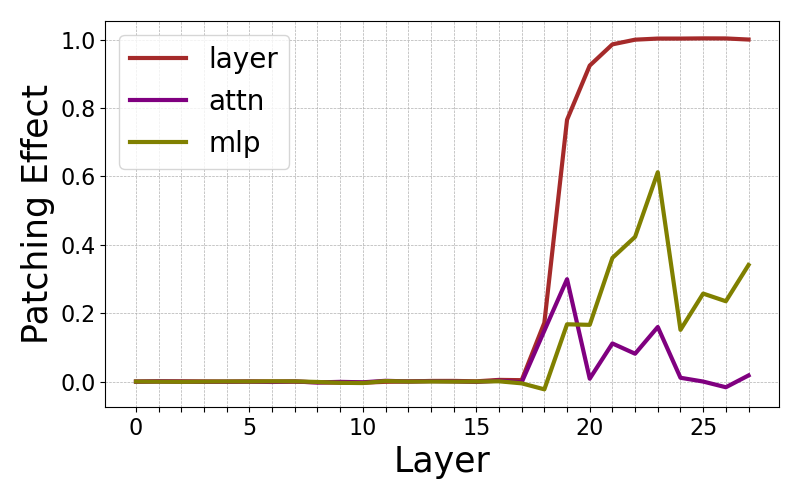}
        \subcaption{Abstraction}
    \end{minipage}


    \begin{minipage}{0.32\textwidth}
        \includegraphics[width=\linewidth]{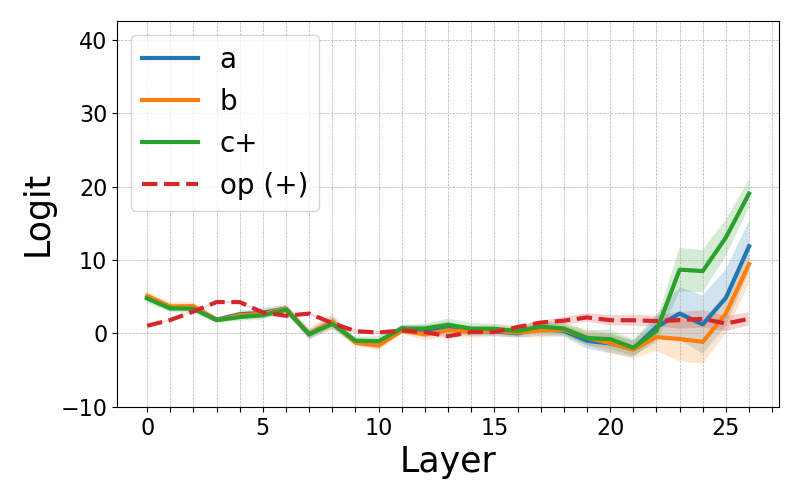}
        \subcaption{Block}
    \end{minipage}
    \begin{minipage}{0.32\textwidth}
        \includegraphics[width=\linewidth]{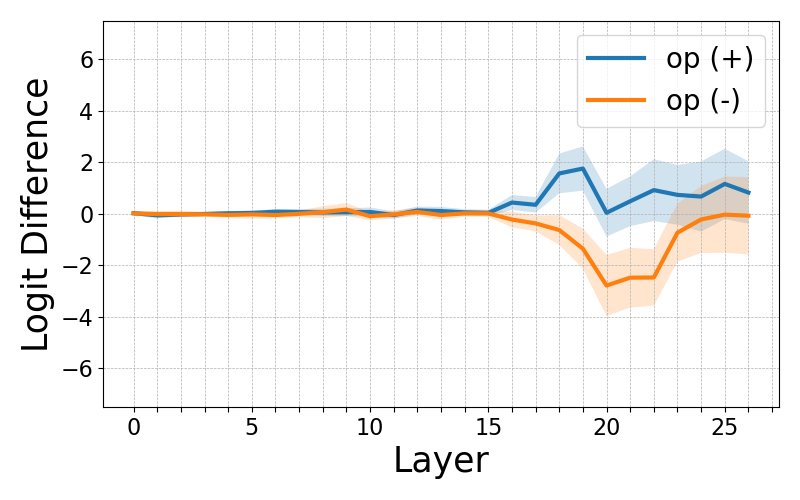}
        \subcaption{Resid Mid}
    \end{minipage}
    \begin{minipage}{0.32\textwidth}
        \includegraphics[width=\linewidth]{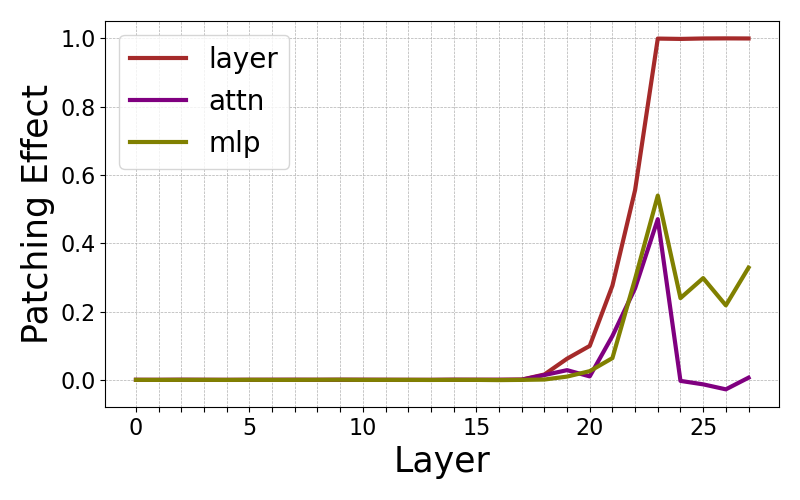}
        \subcaption{Computation}
    \end{minipage}

    \caption{Visualizations of internal computations at last token position in \textbf{Qwen 2.5 7B} for \textbf{addition} math word problems: (a, b, d, e) for logit attribution results, (c, d) activation patching for results.}
    \label{fig:7b-addition}
\end{figure*}

\begin{figure*}[ht]
    \centering
    \begin{minipage}{0.32\textwidth}
        \includegraphics[width=\linewidth]{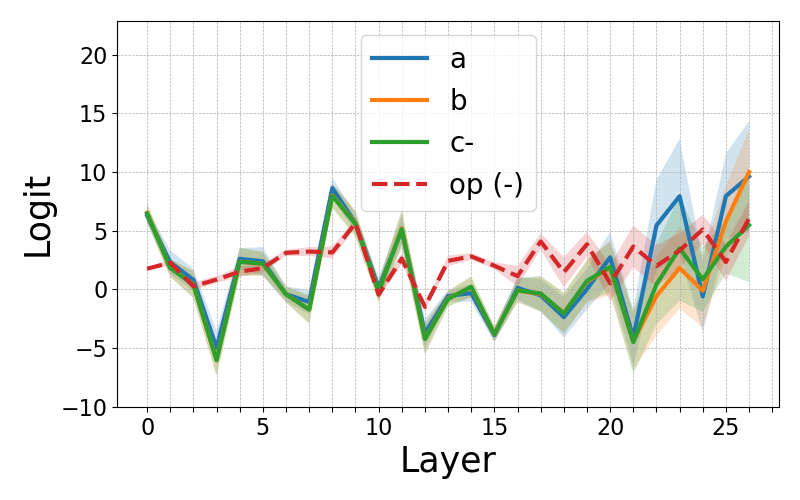}
        \subcaption{Attn}
    \end{minipage}
    \begin{minipage}{0.32\textwidth}
        \includegraphics[width=\linewidth]{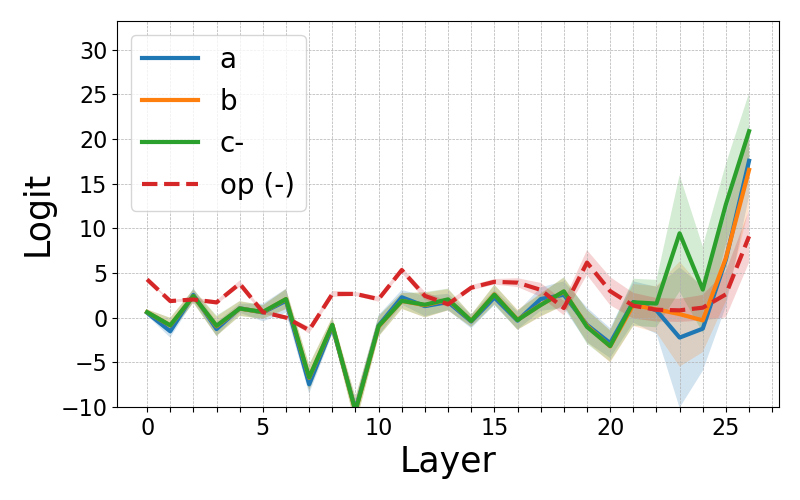}
        \subcaption{MLP}
    \end{minipage}
    \begin{minipage}{0.32\textwidth}
        \includegraphics[width=\linewidth]{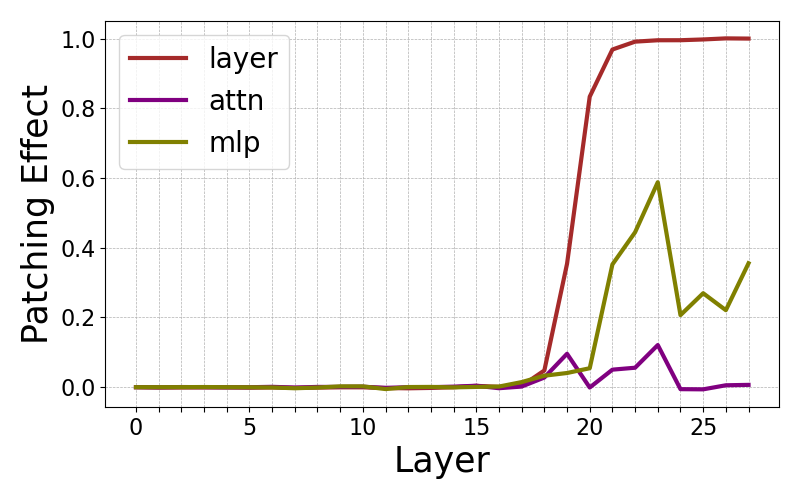}
        \subcaption{Abstraction}
    \end{minipage}


    \begin{minipage}{0.32\textwidth}
        \includegraphics[width=\linewidth]{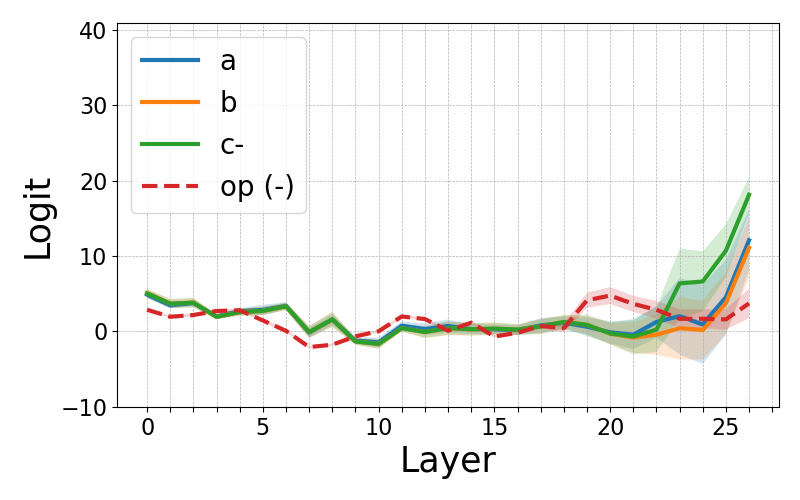}
        \subcaption{Block}
    \end{minipage}
    \begin{minipage}{0.32\textwidth}
        \includegraphics[width=\linewidth]{images/addition-substraction_resid_mid_diff_qwen-7b-instruct.png}
        \subcaption{Resid Mid}
    \end{minipage}
    \begin{minipage}{0.32\textwidth}
        \includegraphics[width=\linewidth]{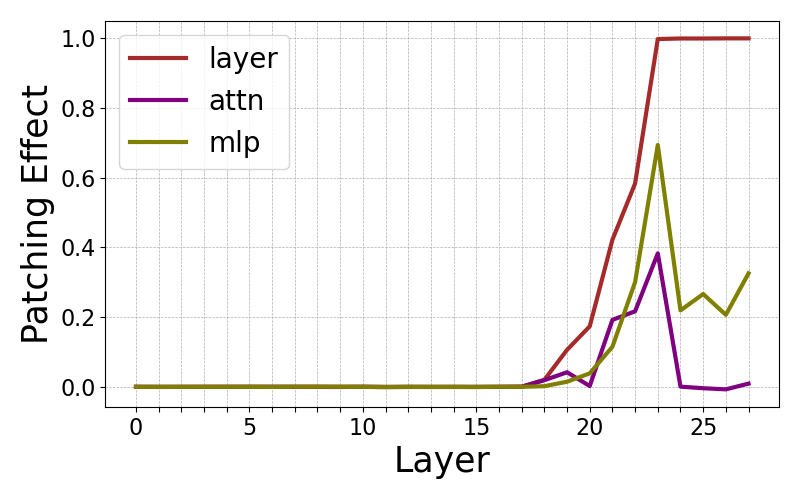}
        \subcaption{Computation}
    \end{minipage}

    \caption{Visualizations of internal computations at last token position in \textbf{Qwen 2.5 7B} for \textbf{subtraction} math word problems: (a, b, d, e) for logit attribution results, (c, d) activation patching for results.}
    \label{fig:7b-subtraction}
\end{figure*}

\begin{figure*}[ht]
    \centering
    \begin{minipage}{0.32\textwidth}
        \includegraphics[width=\linewidth]{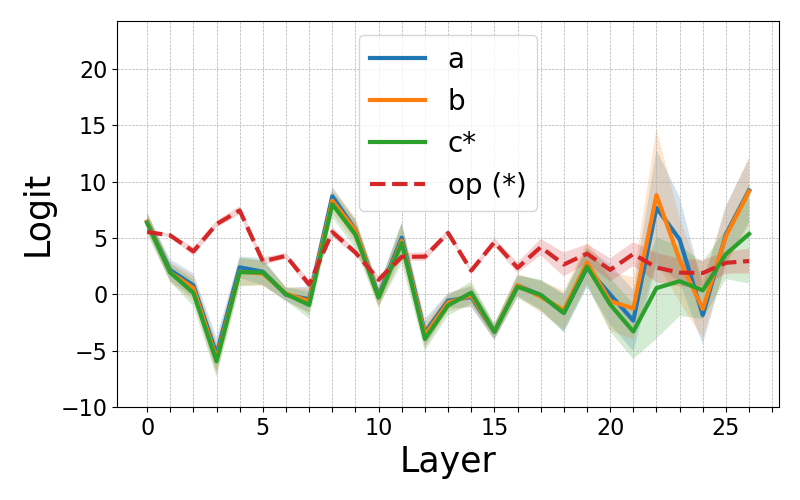}
        \subcaption{Attn}
    \end{minipage}
    \begin{minipage}{0.32\textwidth}
        \includegraphics[width=\linewidth]{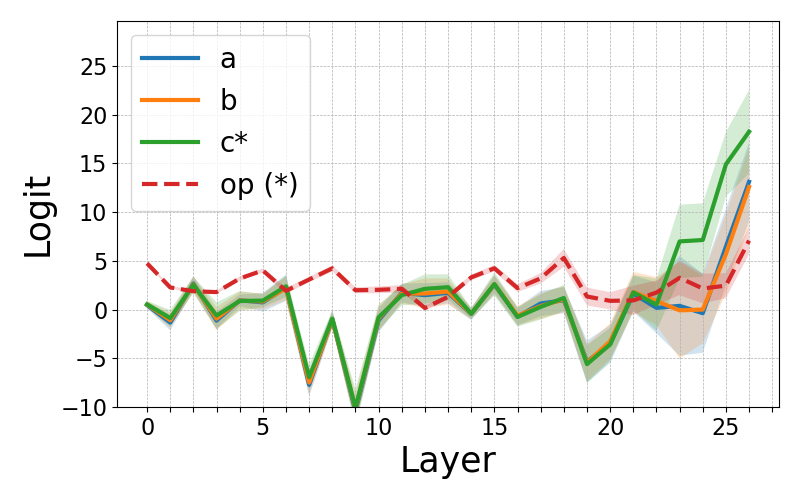}
        \subcaption{MLP}
    \end{minipage}
    \begin{minipage}{0.32\textwidth}
        \includegraphics[width=\linewidth]{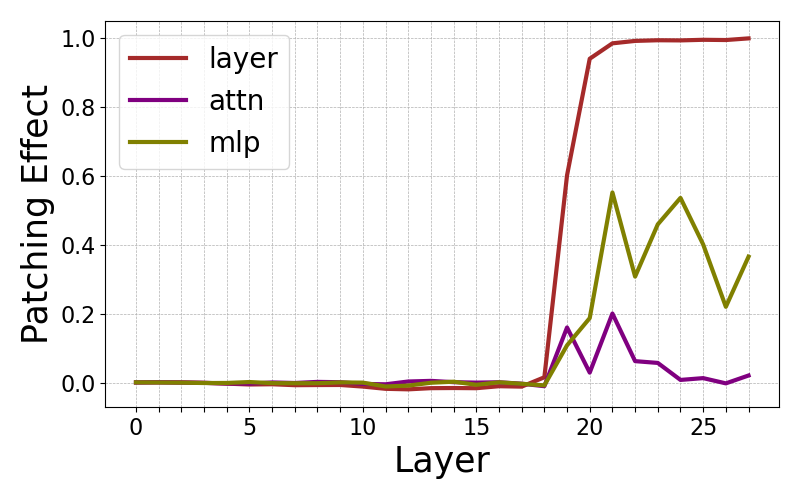}
        \subcaption{Abstraction}
    \end{minipage}


    \begin{minipage}{0.32\textwidth}
        \includegraphics[width=\linewidth]{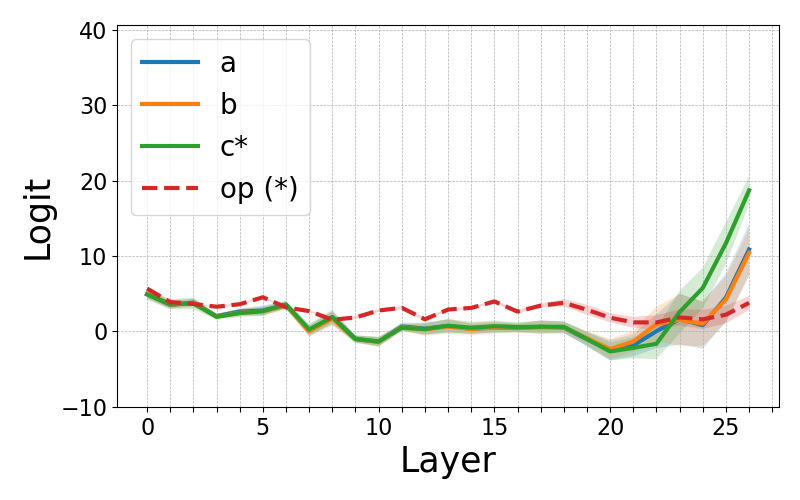}
        \subcaption{Block}
    \end{minipage}
    \begin{minipage}{0.32\textwidth}
        \includegraphics[width=\linewidth]{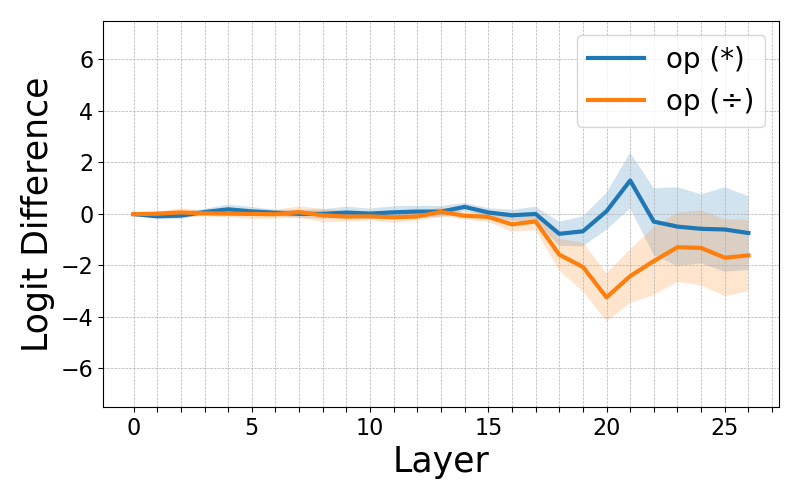}
        \subcaption{Resid Mid}
    \end{minipage}
    \begin{minipage}{0.32\textwidth}
        \includegraphics[width=\linewidth]{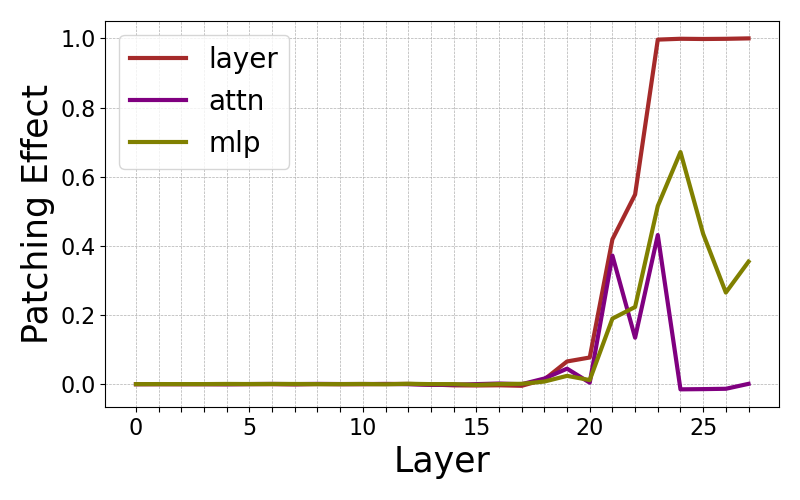}
        \subcaption{Computation}
    \end{minipage}

    \caption{Visualizations of internal computations at last token position in \textbf{Qwen 2.5 7B} for \textbf{multiplication} math word problems: (a, b, d, e) for logit attribution results, (c, d) activation patching for results.}
    \label{fig:7b-multiplication}
\end{figure*}

\begin{figure*}[ht]
    \centering
    \begin{minipage}{0.32\textwidth}
        \includegraphics[width=\linewidth]{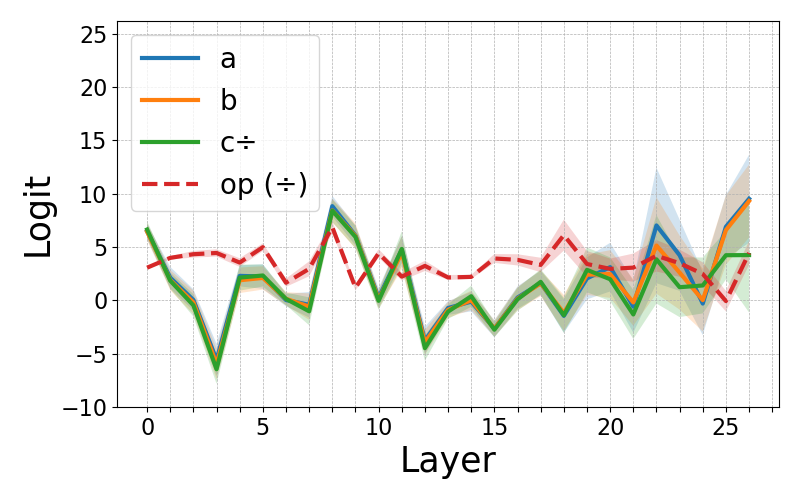}
        \subcaption{Attn}
    \end{minipage}
    \begin{minipage}{0.32\textwidth}
        \includegraphics[width=\linewidth]{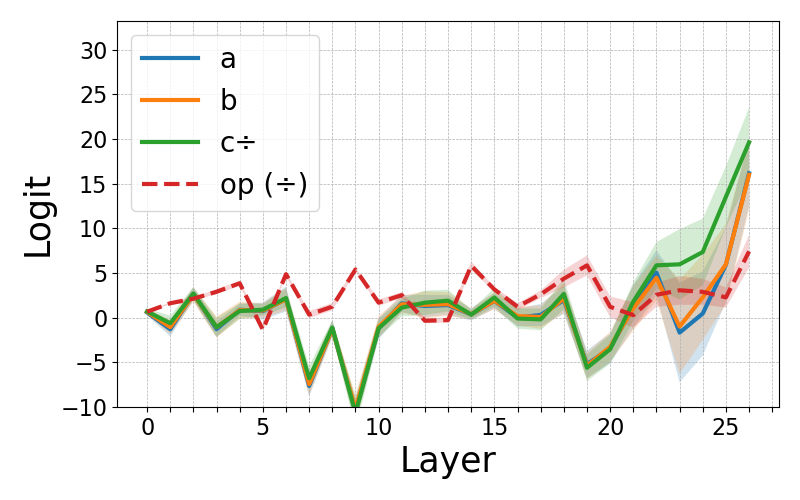}
        \subcaption{MLP}
    \end{minipage}
    \begin{minipage}{0.32\textwidth}
        \includegraphics[width=\linewidth]{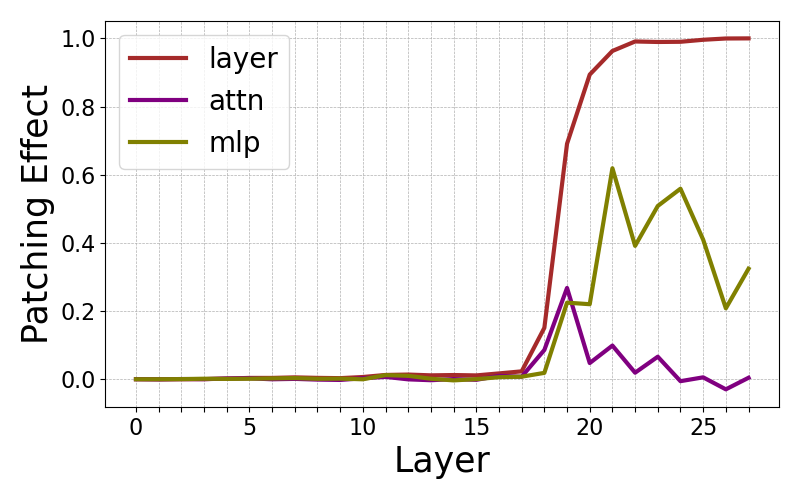}
        \subcaption{Abstraction}
    \end{minipage}


    \begin{minipage}{0.32\textwidth}
        \includegraphics[width=\linewidth]{images/multiplication_resid_final_avg_qwen-7b-instruct.png}
        \subcaption{Block}
    \end{minipage}
    \begin{minipage}{0.32\textwidth}
        \includegraphics[width=\linewidth]{images/multiplication-division_resid_mid_diff_qwen-7b-instruct.png}
        \subcaption{Resid Mid}
    \end{minipage}
    \begin{minipage}{0.32\textwidth}
        \includegraphics[width=\linewidth]{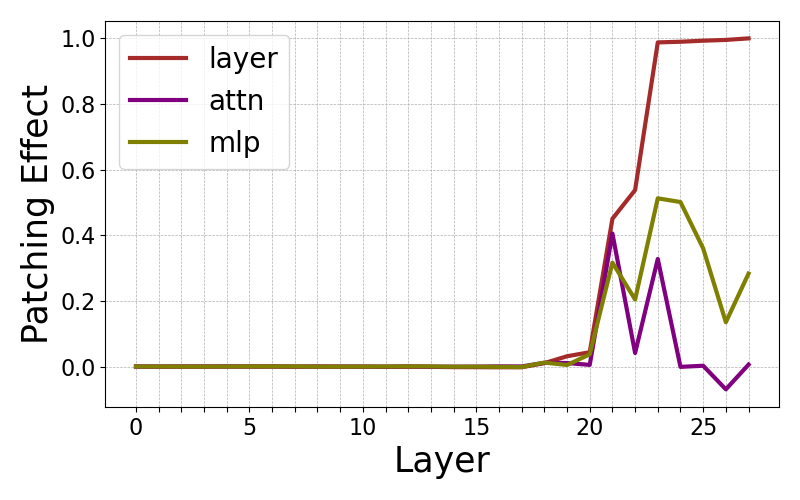}
        \subcaption{Computation}
    \end{minipage}

    \caption{Visualizations of internal computations at last token position in \textbf{Qwen 2.5 7B} for \textbf{division} math word problems: (a, b, d, e) for logit attribution results, (c, d) activation patching for results.}
    \label{fig:7b-division}
\end{figure*}

\begin{figure*}[ht]
    \centering
    \begin{minipage}{0.32\textwidth}
        \includegraphics[width=\linewidth]{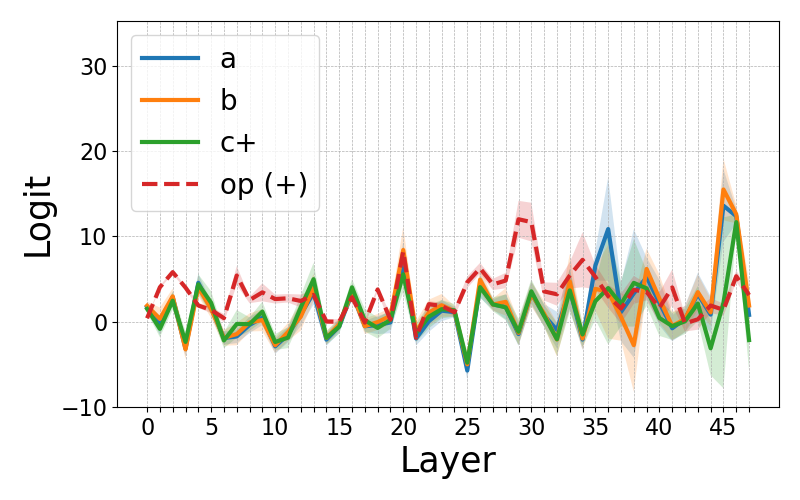}
        \subcaption{Attn}
    \end{minipage}
    \begin{minipage}{0.32\textwidth}
        \includegraphics[width=\linewidth]{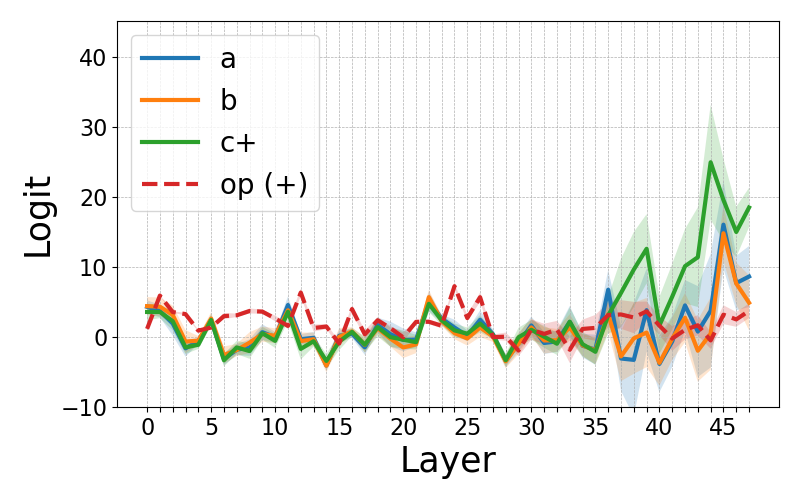}
        \subcaption{MLP}
    \end{minipage}
    \begin{minipage}{0.32\textwidth}
        \includegraphics[width=\linewidth]{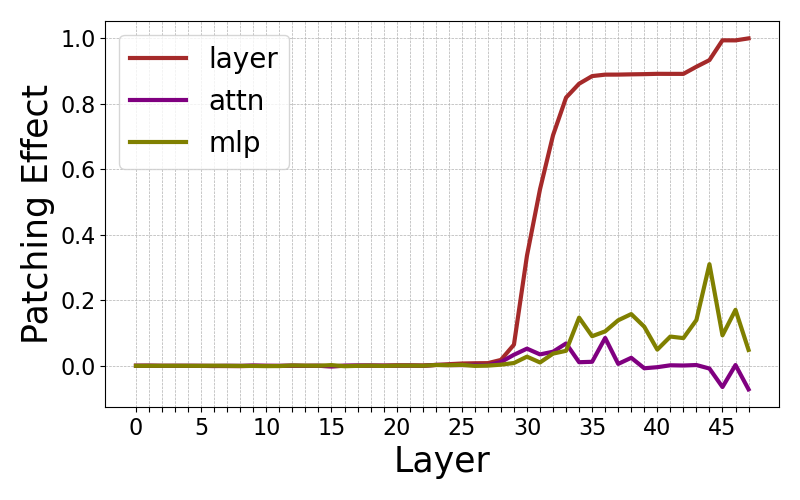}
        \subcaption{Abstraction}
    \end{minipage}


    \begin{minipage}{0.32\textwidth}
        \includegraphics[width=\linewidth]{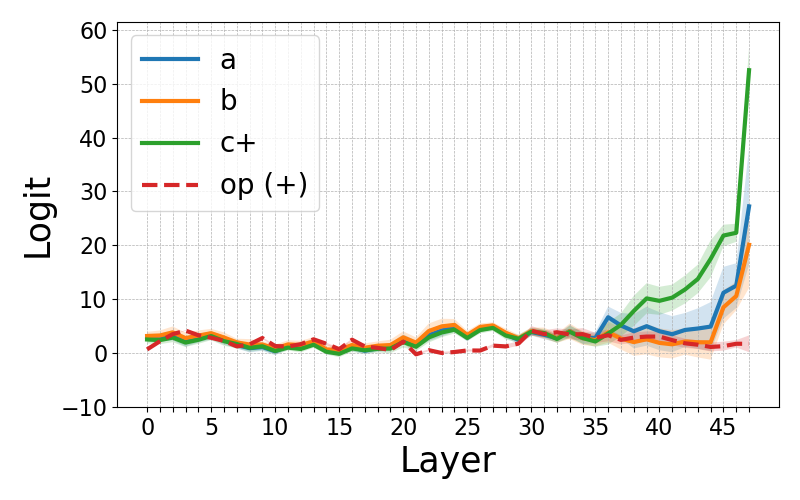}
        \subcaption{Block}
    \end{minipage}
    \begin{minipage}{0.32\textwidth}
        \includegraphics[width=\linewidth]{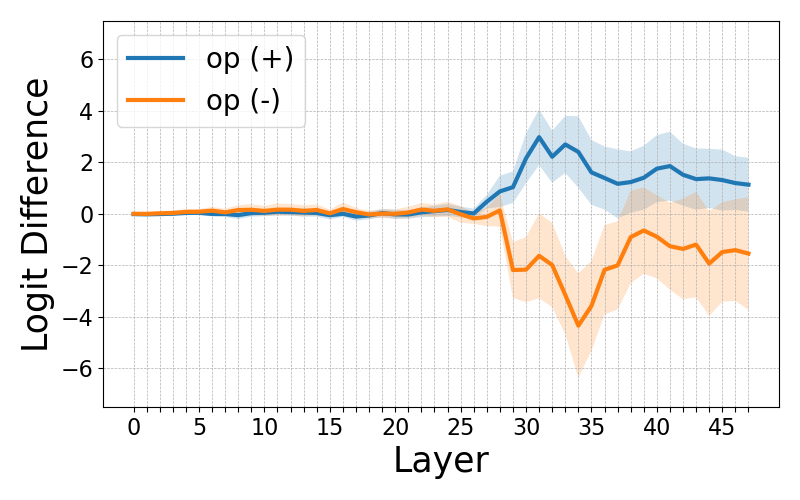}
        \subcaption{Resid Mid}
    \end{minipage}
    \begin{minipage}{0.32\textwidth}
        \includegraphics[width=\linewidth]{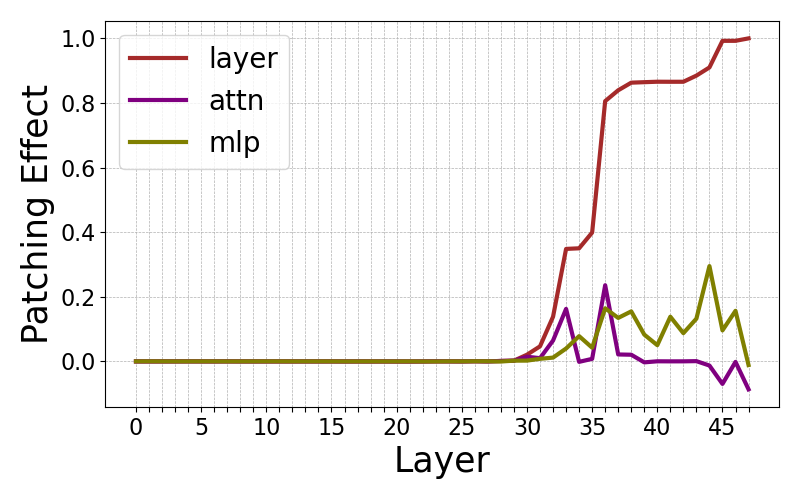}
        \subcaption{Computation}
    \end{minipage}

    \caption{Visualizations of internal computations at last token position in \textbf{Qwen 2.5 14B} for \textbf{addition} math word problems: (a, b, d, e) for logit attribution results, (c, d) activation patching for results.}
    \label{fig:14b-addition}
\end{figure*}

\begin{figure*}[ht]
    \centering
    \begin{minipage}{0.32\textwidth}
        \includegraphics[width=\linewidth]{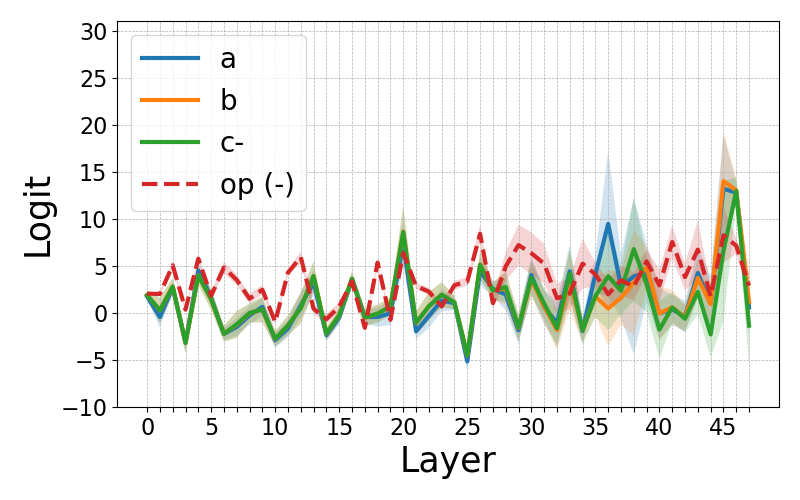}
        \subcaption{Attn}
    \end{minipage}
    \begin{minipage}{0.32\textwidth}
        \includegraphics[width=\linewidth]{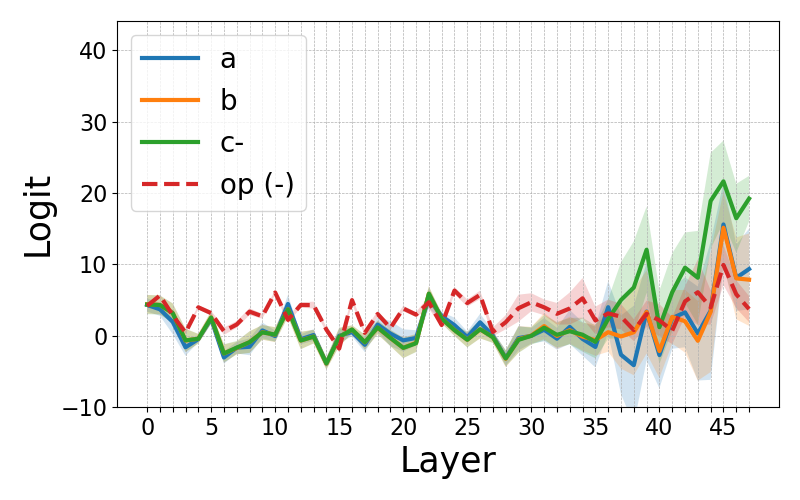}
        \subcaption{MLP}
    \end{minipage}
    \begin{minipage}{0.32\textwidth}
        \includegraphics[width=\linewidth]{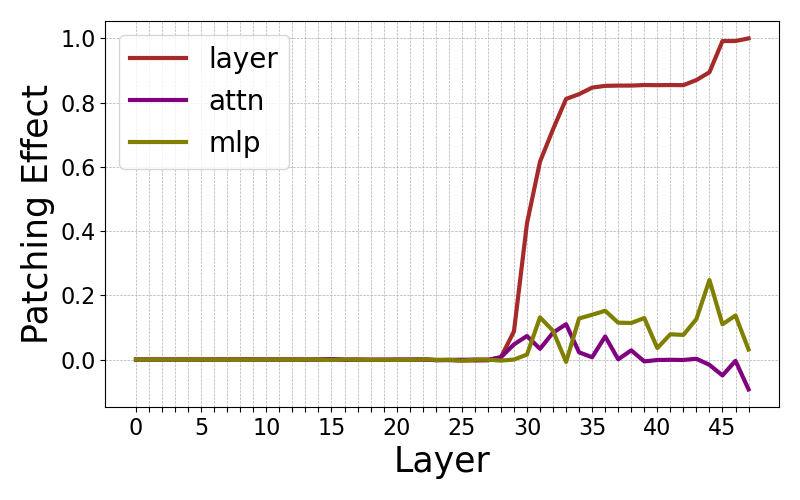}
        \subcaption{Abstraction}
    \end{minipage}


    \begin{minipage}{0.32\textwidth}
        \includegraphics[width=\linewidth]{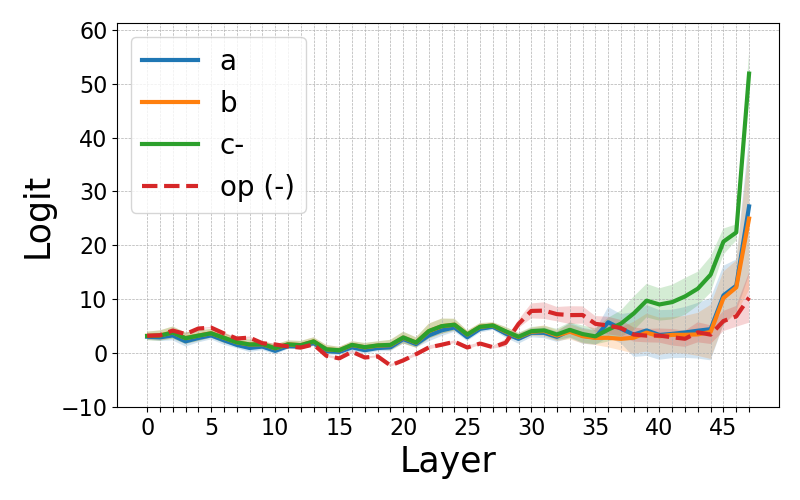}
        \subcaption{Block}
    \end{minipage}
    \begin{minipage}{0.32\textwidth}
        \includegraphics[width=\linewidth]{images/addition-substraction_resid_mid_diff_qwen-14b-instruct.png}
        \subcaption{Resid Mid}
    \end{minipage}
    \begin{minipage}{0.32\textwidth}
        \includegraphics[width=\linewidth]{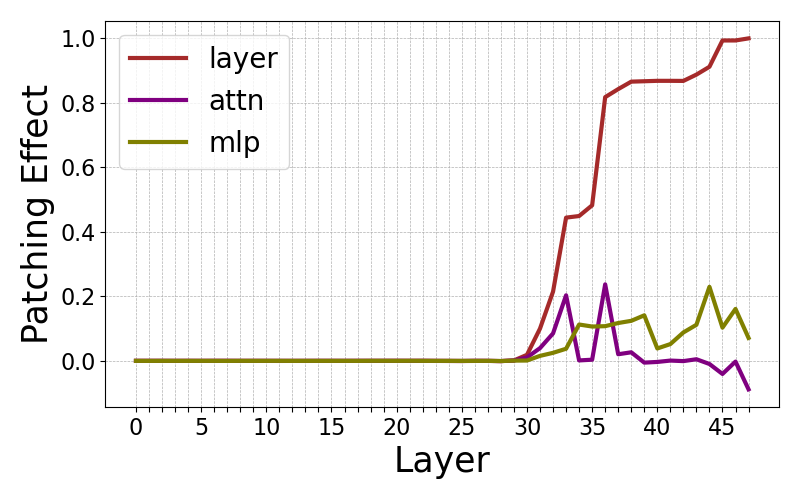}
        \subcaption{Computation}
    \end{minipage}

    \caption{Visualizations of internal computations at last token position in \textbf{Qwen 2.5 14B} for \textbf{subtraction} math word problems: (a, b, d, e) for logit attribution results, (c, d) activation patching for results.}
    \label{fig:14b-subtraction}
\end{figure*}

\begin{figure*}[ht]
    \centering
    \begin{minipage}{0.32\textwidth}
        \includegraphics[width=\linewidth]{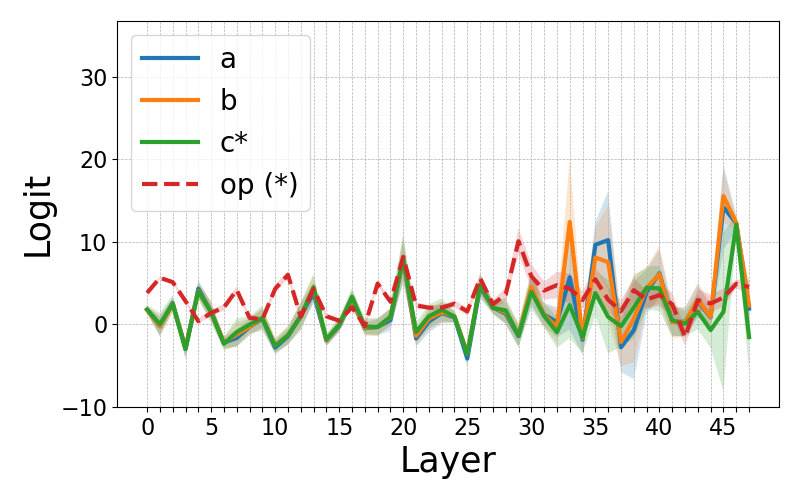}
        \subcaption{Attn}
    \end{minipage}
    \begin{minipage}{0.32\textwidth}
        \includegraphics[width=\linewidth]{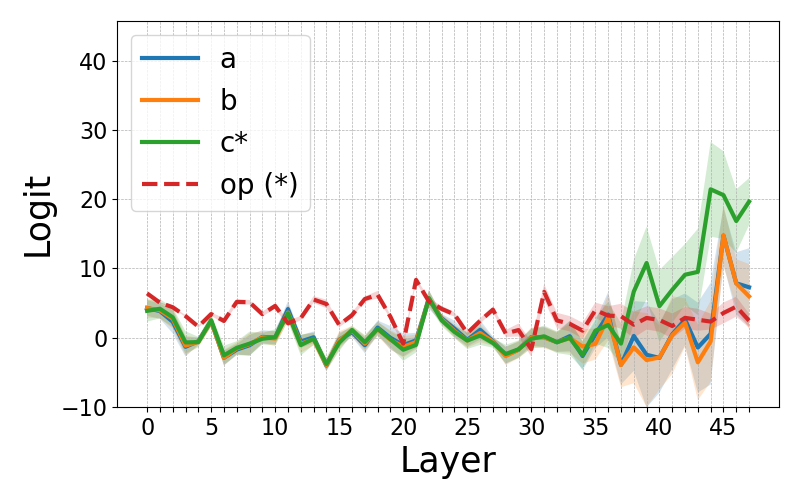}
        \subcaption{MLP}
    \end{minipage}
    \begin{minipage}{0.32\textwidth}
        \includegraphics[width=\linewidth]{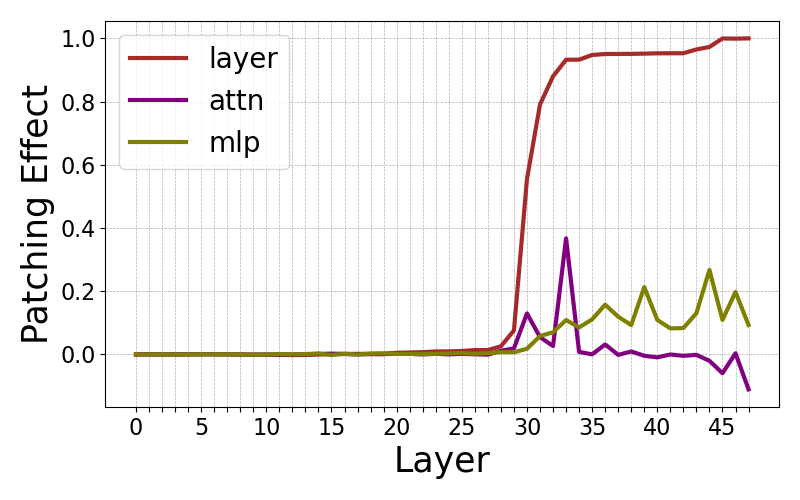}
        \subcaption{Abstraction}
    \end{minipage}


    \begin{minipage}{0.32\textwidth}
        \includegraphics[width=\linewidth]{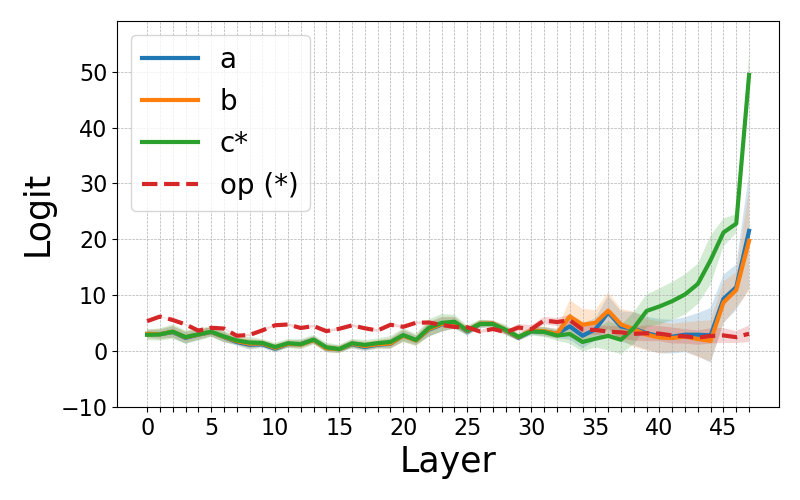}
        \subcaption{Block}
    \end{minipage}
    \begin{minipage}{0.32\textwidth}
        \includegraphics[width=\linewidth]{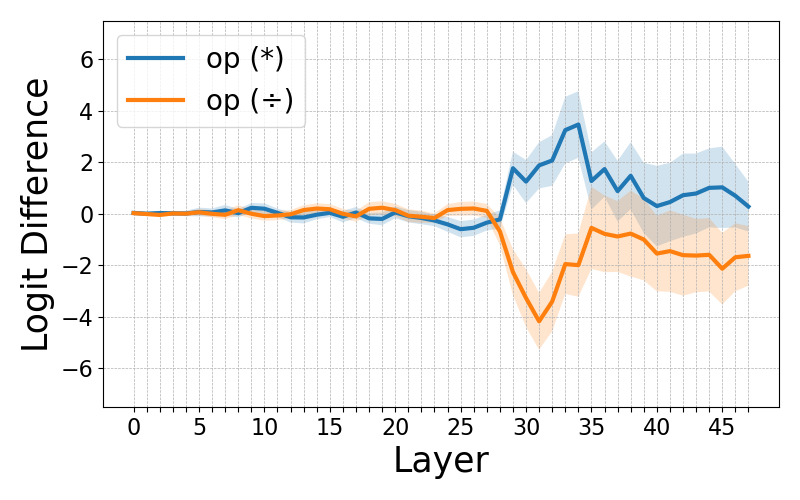}
        \subcaption{Resid Mid}
    \end{minipage}
    \begin{minipage}{0.32\textwidth}
        \includegraphics[width=\linewidth]{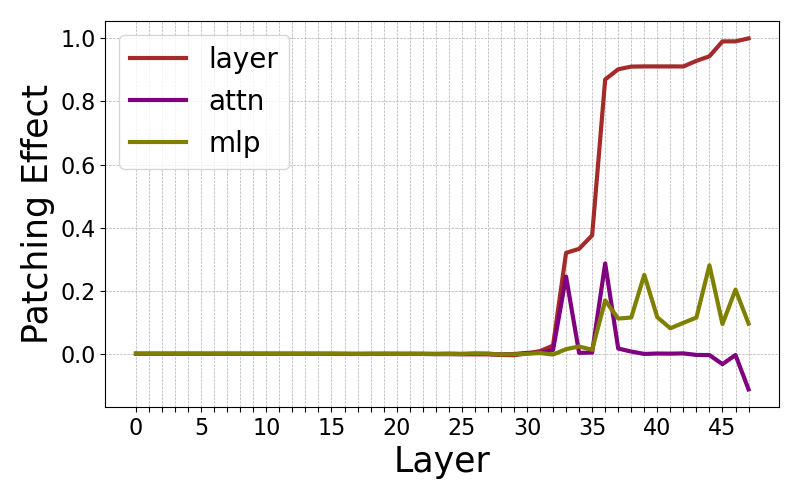}
        \subcaption{Computation}
    \end{minipage}

    \caption{Visualizations of internal computations at last token position in \textbf{Qwen 2.5 14B} for \textbf{multiplication} math word problems: (a, b, d, e) for logit attribution results, (c, d) activation patching for results.}
    \label{fig:14b-multiplication}
\end{figure*}

\begin{figure*}[ht]
    \centering
    \begin{minipage}{0.32\textwidth}
        \includegraphics[width=\linewidth]{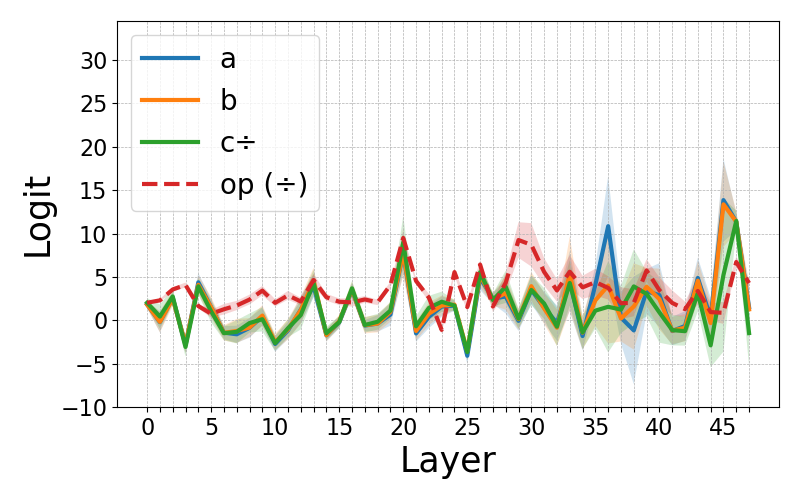}
        \subcaption{Attn}
    \end{minipage}
    \begin{minipage}{0.32\textwidth}
        \includegraphics[width=\linewidth]{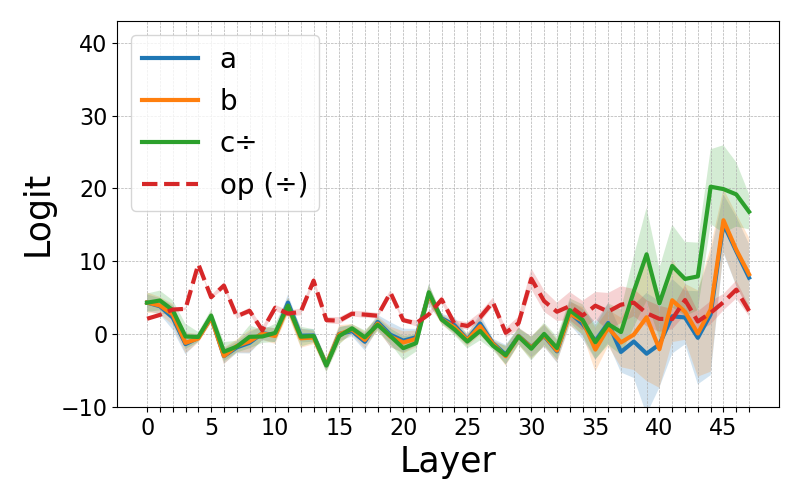}
        \subcaption{MLP}
    \end{minipage}
    \begin{minipage}{0.32\textwidth}
        \includegraphics[width=\linewidth]{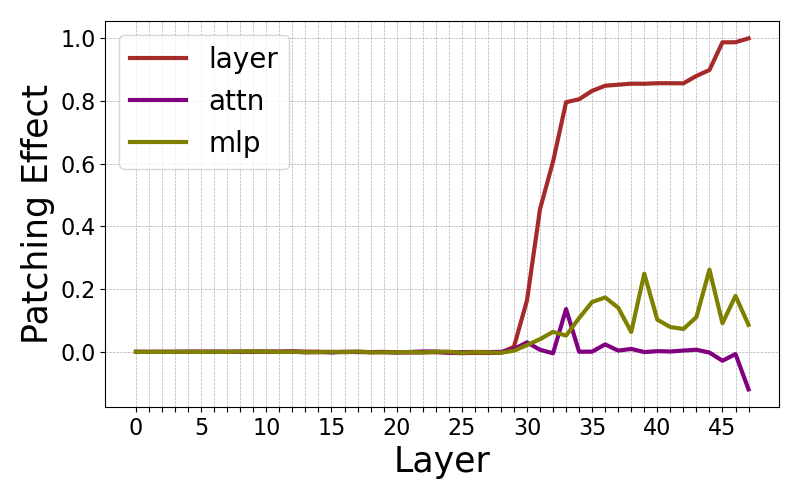}
        \subcaption{Abstraction}
    \end{minipage}


    \begin{minipage}{0.32\textwidth}
        \includegraphics[width=\linewidth]{images/multiplication_resid_final_avg_qwen-14b-instruct.png}
        \subcaption{Block}
    \end{minipage}
    \begin{minipage}{0.32\textwidth}
        \includegraphics[width=\linewidth]{images/multiplication-division_resid_mid_diff_qwen-14b-instruct.png}
        \subcaption{Resid Mid}
    \end{minipage}
    \begin{minipage}{0.32\textwidth}
        \includegraphics[width=\linewidth]{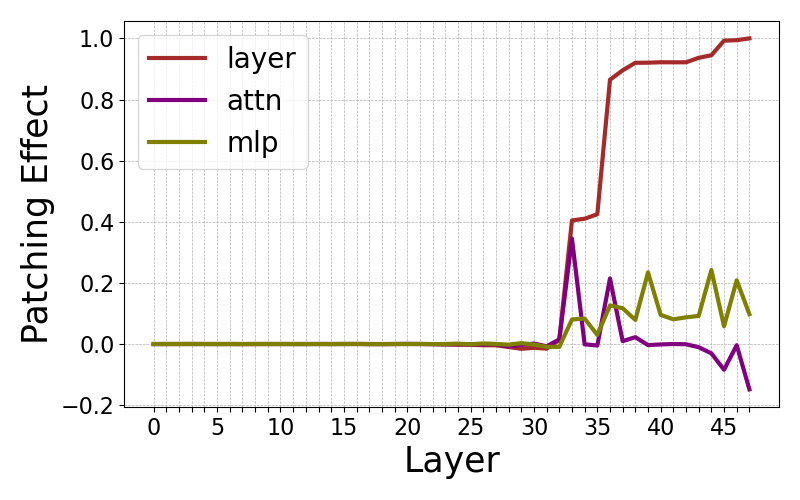}
        \subcaption{Computation}
    \end{minipage}

    \caption{Visualizations of internal computations at last token position in \textbf{Qwen 2.5 14B} for \textbf{division} math word problems: (a, b, d, e) for logit attribution results, (c, d) activation patching for results. We label the starting layer of abstraction, operand moving and computation in pink, blue and green, respectively.}
    \label{fig:14b-division}
\end{figure*}

For additional interpretability results using logit lens and activation patching:
\begin{itemize}
    \item See Figure~\ref{fig:llama-subtraction}, Figure~\ref{fig:llama-multiplication}, and Figure~\ref{fig:llama-division} for Llama-3 8B on subtraction, multiplication, and division.
    \item See Figure~\ref{fig:7b-addition}, Figure~\ref{fig:7b-subtraction}, Figure~\ref{fig:7b-multiplication}, and Figure~\ref{fig:7b-division} for Qwen 2.5 7B on all four operations: addition, subtraction, multiplication, and division.
    \item See Figure~\ref{fig:14b-addition}, Figure~\ref{fig:14b-subtraction}, Figure~\ref{fig:14b-multiplication}, and Figure~\ref{fig:14b-division} for Qwen 2.5 14B on the same set of arithmetic tasks.
\end{itemize}





\paragraph{Two-Operator Dataset}
For two operator dataset, we only report results for Qwen 2.5 7B and Qwen 2.5 14B, because Llama-3 8B only achieve 16.5\% accuracy on this dataset.

\begin{figure*}[ht]
    \centering
    \begin{minipage}{0.32\textwidth}
        \includegraphics[width=\linewidth]{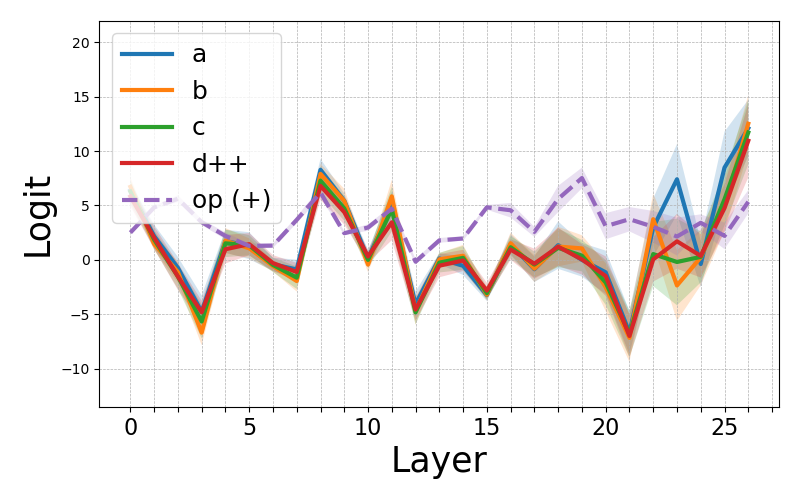}
        \subcaption{Attn}
    \end{minipage}
    \begin{minipage}{0.32\textwidth}
        \includegraphics[width=\linewidth]{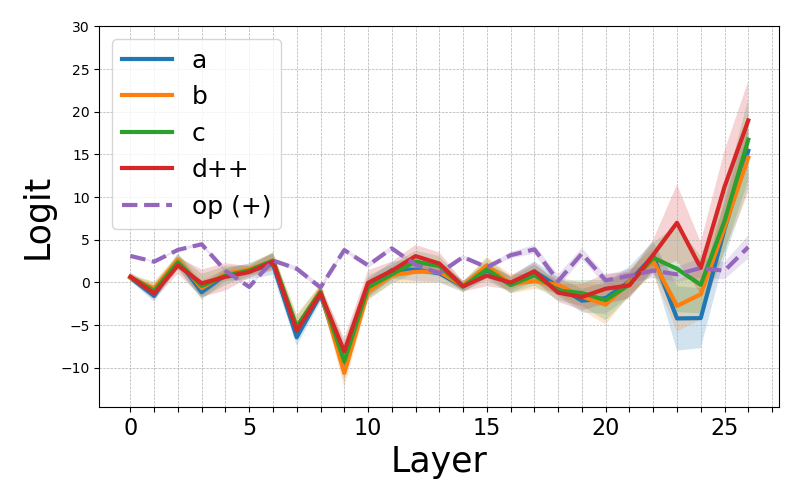}
        \subcaption{MLP}
    \end{minipage}
    \begin{minipage}{0.32\textwidth}
        \includegraphics[width=\linewidth]{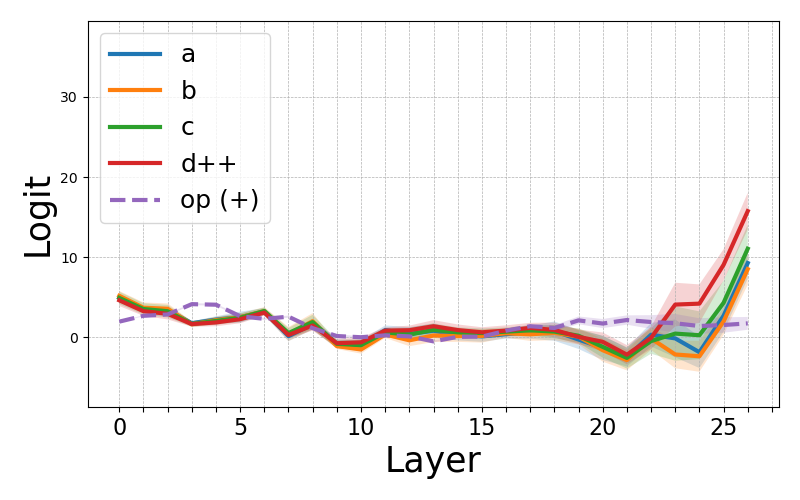}
        \subcaption{Resid Final}
    \end{minipage}


    \begin{minipage}{0.32\textwidth}
        \includegraphics[width=\linewidth]{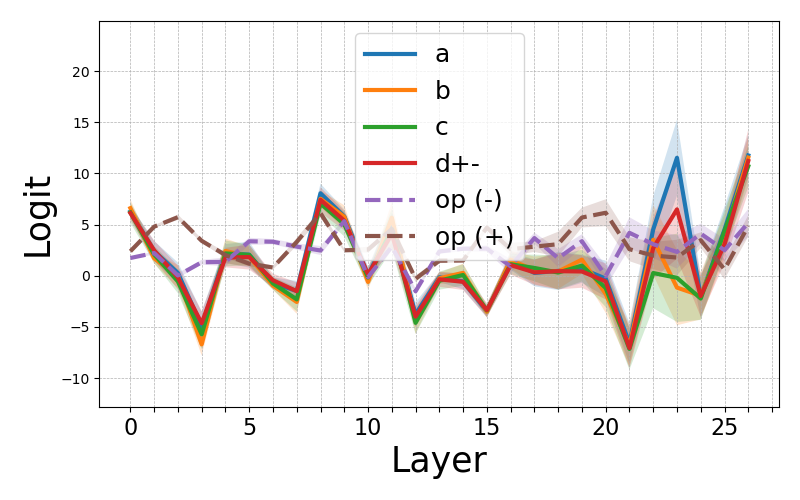}
        \subcaption{Attn}
    \end{minipage}
    \begin{minipage}{0.32\textwidth}
        \includegraphics[width=\linewidth]{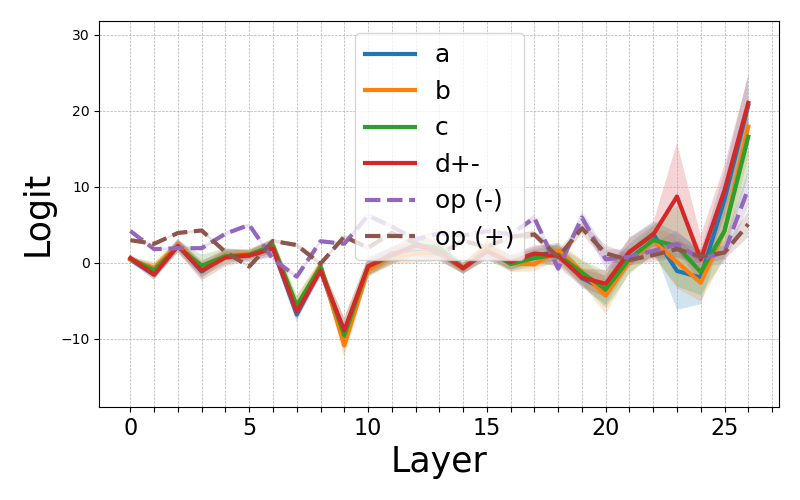}
        \subcaption{MLP}
    \end{minipage}
    \begin{minipage}{0.32\textwidth}
        \includegraphics[width=\linewidth]{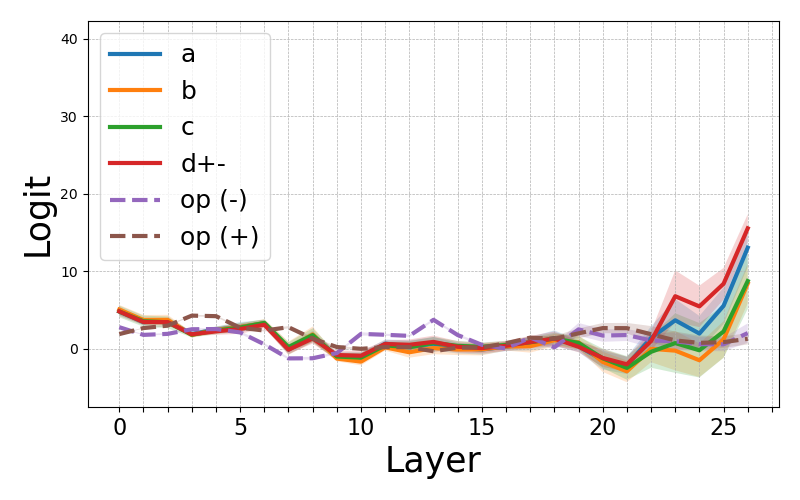}
        \subcaption{Resid Final}
    \end{minipage}

    \begin{minipage}{0.32\textwidth}
        \includegraphics[width=\linewidth]{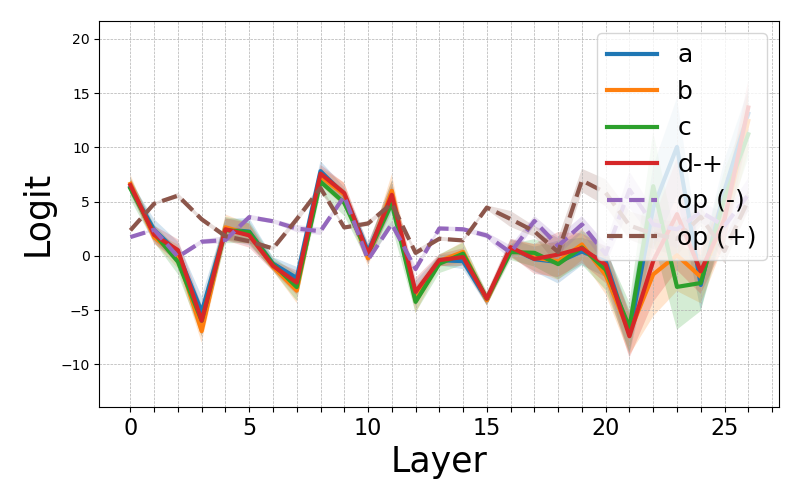}
        \subcaption{Attn}
    \end{minipage}
    \begin{minipage}{0.32\textwidth}
        \includegraphics[width=\linewidth]{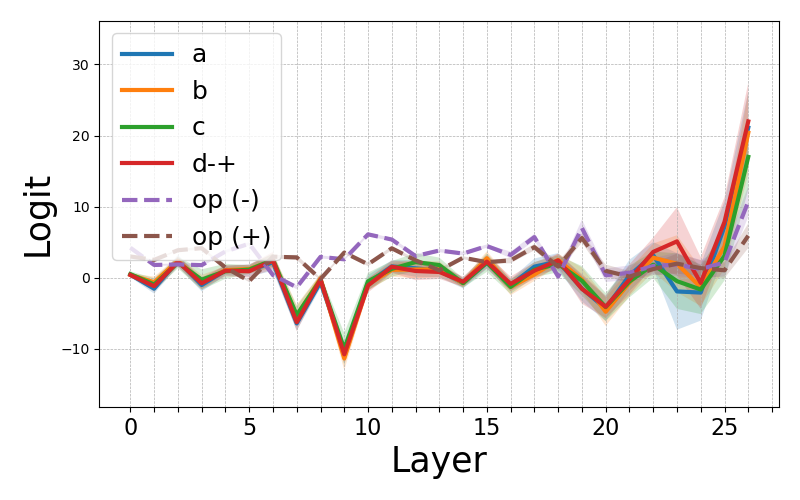}
        \subcaption{MLP}
    \end{minipage}
    \begin{minipage}{0.32\textwidth}
        \includegraphics[width=\linewidth]{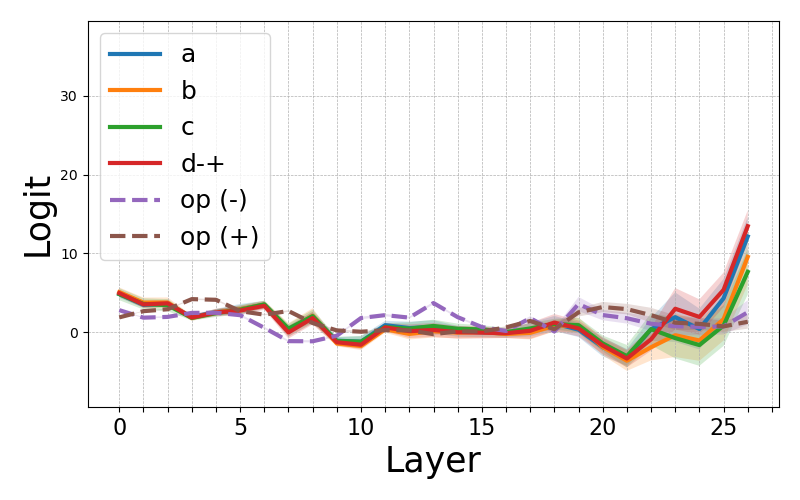}
        \subcaption{Resid Final}
    \end{minipage}

    \begin{minipage}{0.32\textwidth}
        \includegraphics[width=\linewidth]{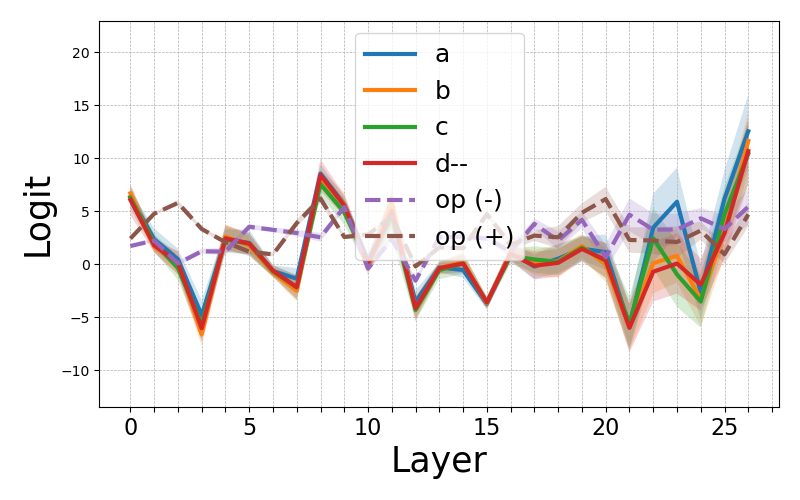}
        \subcaption{Attn}
    \end{minipage}
    \begin{minipage}{0.32\textwidth}
        \includegraphics[width=\linewidth]{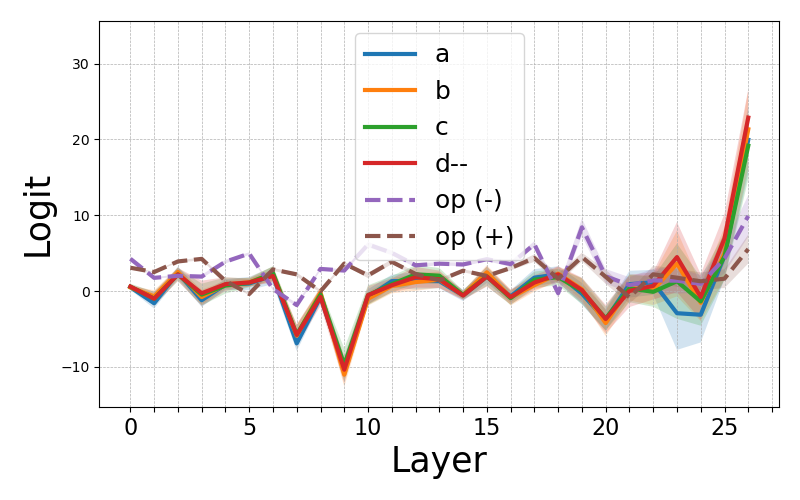}
        \subcaption{MLP}
    \end{minipage}
    \begin{minipage}{0.32\textwidth}
        \includegraphics[width=\linewidth]{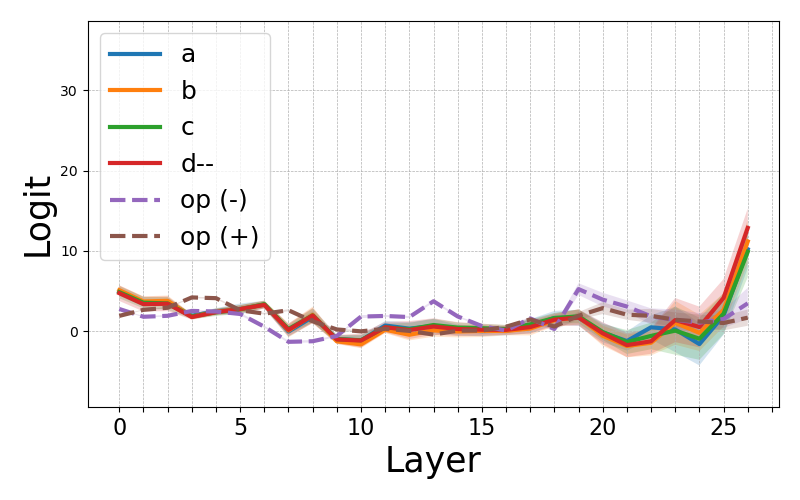}
        \subcaption{Resid Final}
    \end{minipage}

    \caption{Visualizations of internal computations at last token position in \textbf{Qwen 2.5 7B} for \textbf{two-operation} math word problems. \textbf{First row:} for $a + b + c$. \textbf{Second row:} for $a + b - c$. \textbf{Third row:} for $a - b + c$. \textbf{Fourth row:} for $a - b - c$.}
    \label{fig:qwen7b-2op-logit}
\end{figure*}

\begin{figure*}[ht]
    \centering
    \begin{minipage}{0.32\textwidth}
        \includegraphics[width=\linewidth]{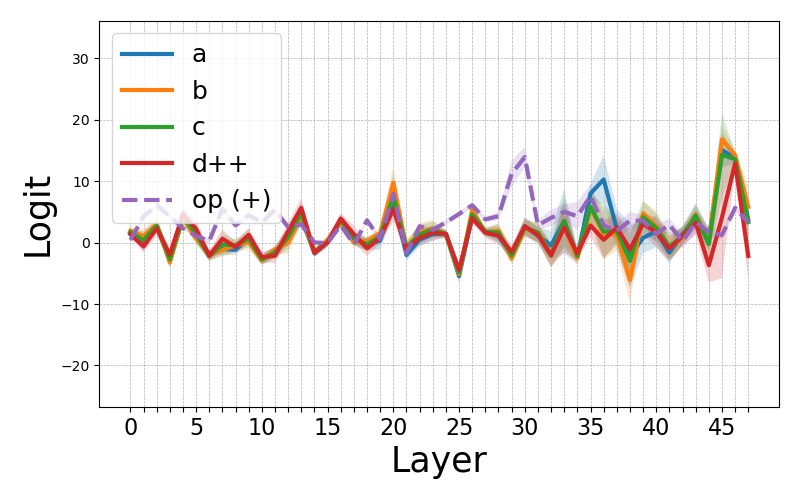}
        \subcaption{Attn}
    \end{minipage}
    \begin{minipage}{0.32\textwidth}
        \includegraphics[width=\linewidth]{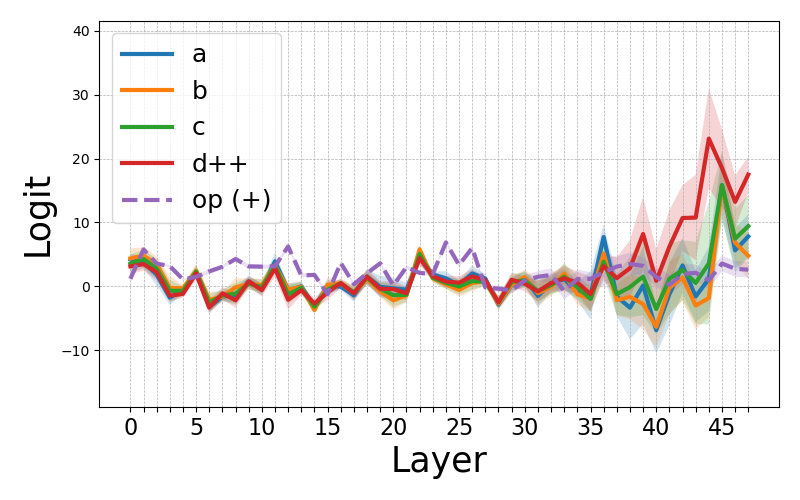}
        \subcaption{MLP}
    \end{minipage}
    \begin{minipage}{0.32\textwidth}
        \includegraphics[width=\linewidth]{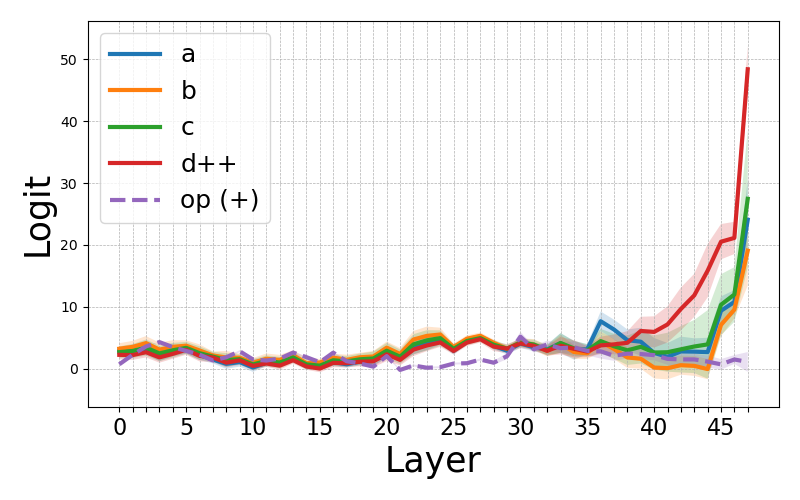}
        \subcaption{Resid Final}
    \end{minipage}


    \begin{minipage}{0.32\textwidth}
        \includegraphics[width=\linewidth]{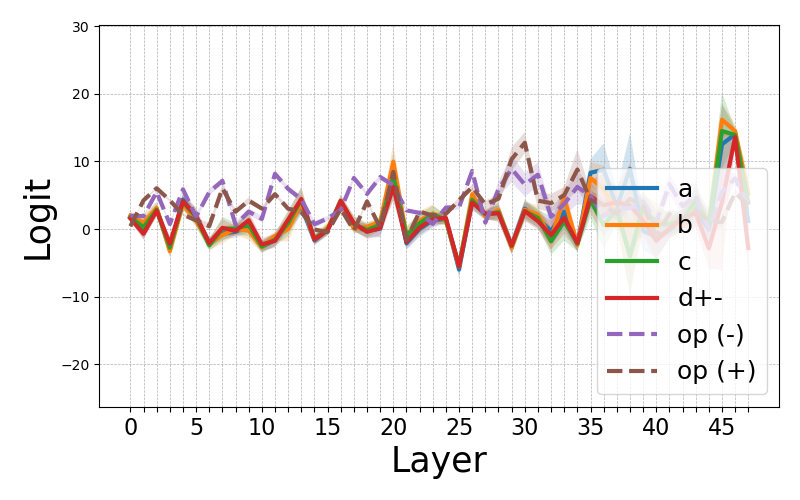}
        \subcaption{Attn}
    \end{minipage}
    \begin{minipage}{0.32\textwidth}
        \includegraphics[width=\linewidth]{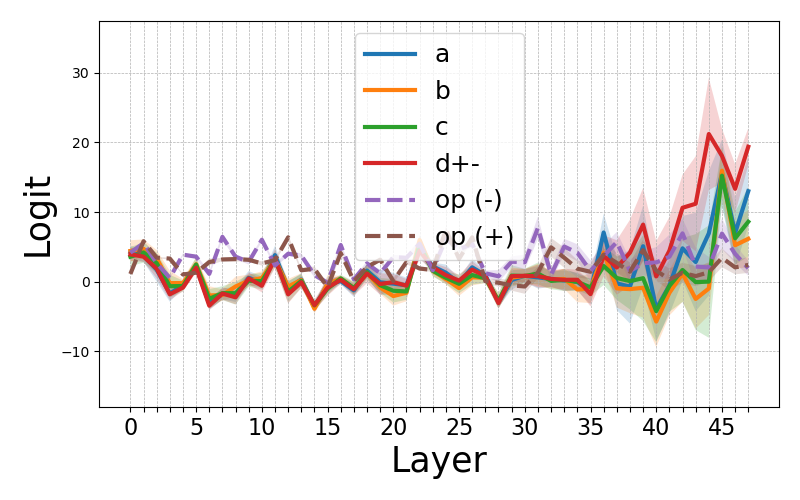}
        \subcaption{MLP}
    \end{minipage}
    \begin{minipage}{0.32\textwidth}
        \includegraphics[width=\linewidth]{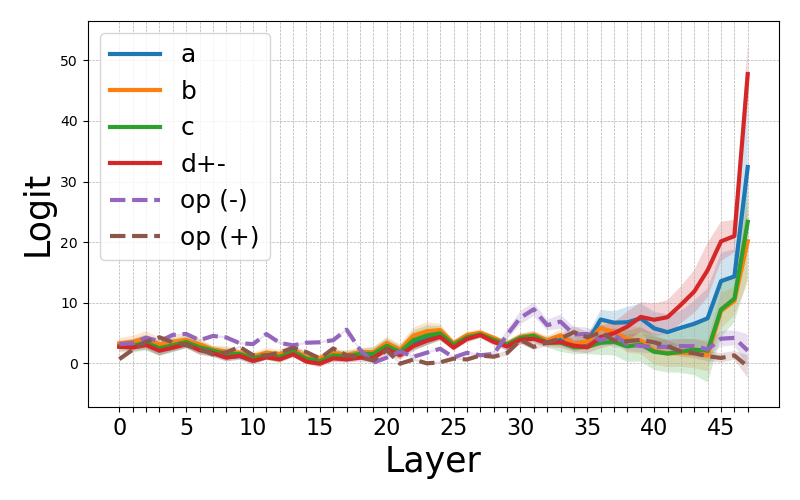}
        \subcaption{Resid Final}
    \end{minipage}

    \begin{minipage}{0.32\textwidth}
        \includegraphics[width=\linewidth]{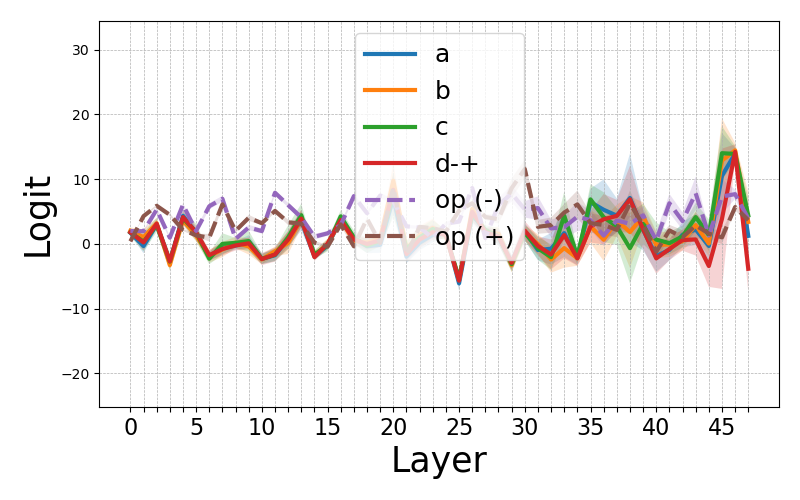}
        \subcaption{Attn}
    \end{minipage}
    \begin{minipage}{0.32\textwidth}
        \includegraphics[width=\linewidth]{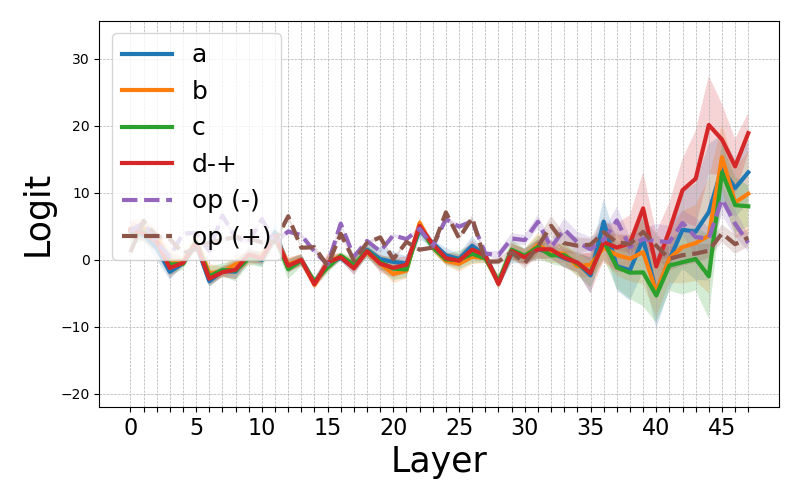}
        \subcaption{MLP}
    \end{minipage}
    \begin{minipage}{0.32\textwidth}
        \includegraphics[width=\linewidth]{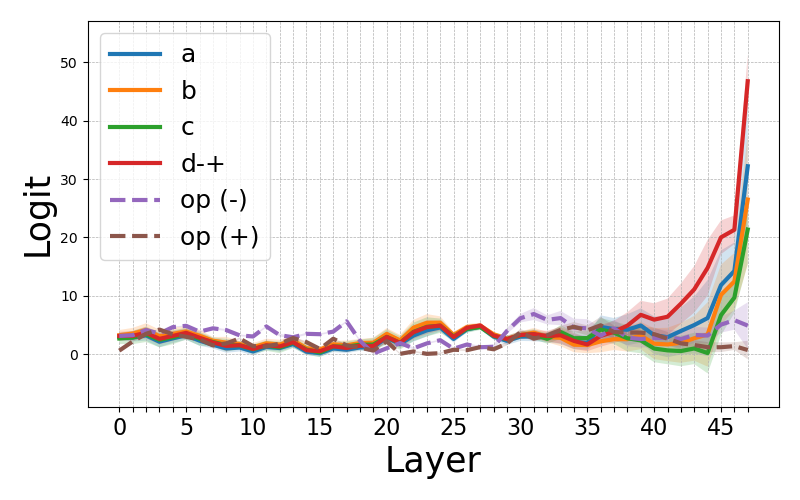}
        \subcaption{Resid Final}
    \end{minipage}

    \begin{minipage}{0.32\textwidth}
        \includegraphics[width=\linewidth]{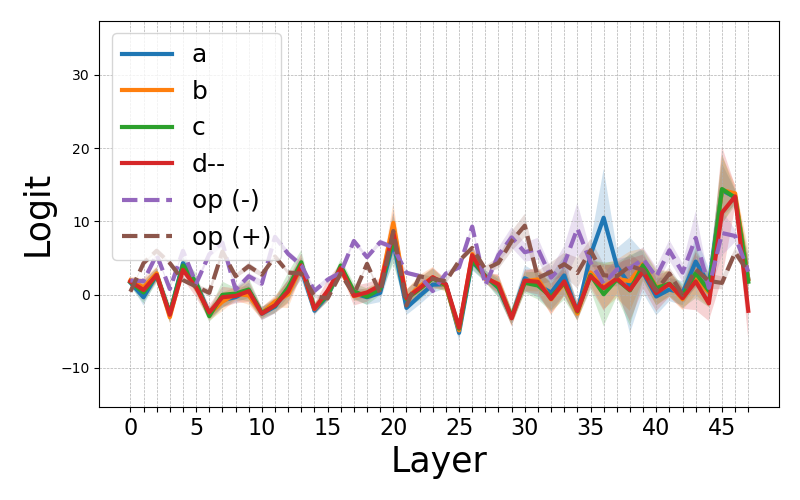}
        \subcaption{Attn}
    \end{minipage}
    \begin{minipage}{0.32\textwidth}
        \includegraphics[width=\linewidth]{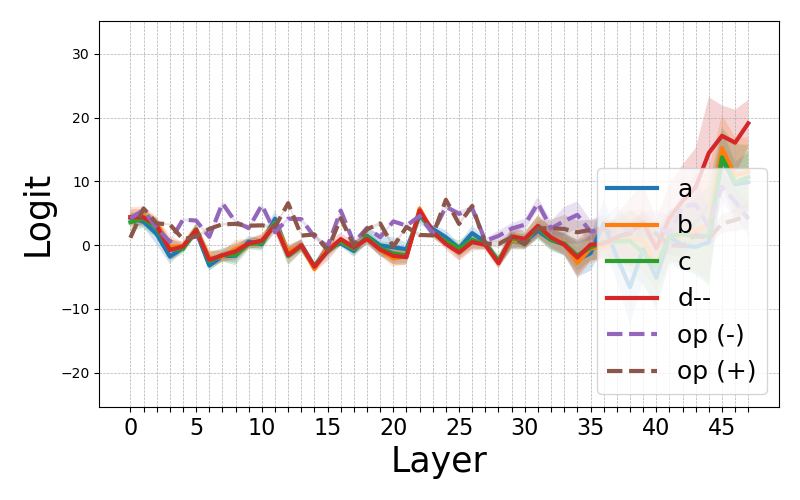}
        \subcaption{MLP}
    \end{minipage}
    \begin{minipage}{0.32\textwidth}
        \includegraphics[width=\linewidth]{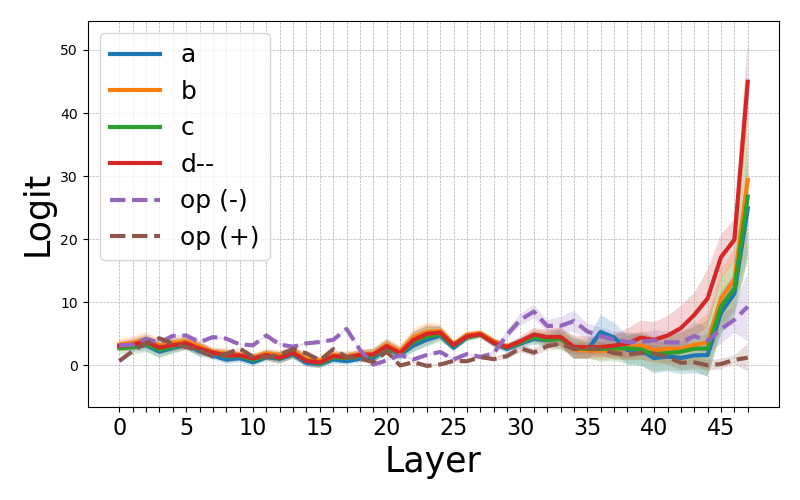}
        \subcaption{Resid Final}
    \end{minipage}

    \caption{Visualizations of internal computations at last token position in \textbf{Qwen 2.5 14B} for \textbf{two-operation} math word problems. \textbf{First row:} for $a + b + c$. \textbf{Second row:} for $a + b - c$. \textbf{Third row:} for $a - b + c$. \textbf{Fourth row:} for $a - b - c$.}
    \label{fig:qwen14b-2op-logit}
\end{figure*}

See Figure~\ref{fig:qwen7b-2op-logit} and Figure~\ref{fig:qwen14b-2op-logit} for logit attribution results for  Qwen 2.5 7B and Qwen 2.5 14B, respectively.









\subsection{Cross-Prompt Patching Additional Results}\label{app:cross_patching_results}
\begin{figure*}[ht]
    \centering
    
    \begin{subfigure}[b]{0.24\linewidth}
        \centering
        \includegraphics[width=\linewidth]{images/Llama-8b-instruct_cross_symbolic_paired_substraction_addition_layer_logprob_line_graph.png}
        \caption{Paired $-$ to $+$}
        \label{fig:add_plot1}
    \end{subfigure}
    \hfill
    \begin{subfigure}[b]{0.24\linewidth}
        \centering
        \includegraphics[width=\linewidth]{images/Llama-8b-instruct_cross_symbolic_substraction_addition_layer_logprob_line_graph.png}
        \caption{$-$ to $+$}
        \label{fig:add_plot2}
    \end{subfigure}
    \hfill
    \begin{subfigure}[b]{0.24\linewidth}
        \centering
        \includegraphics[width=\linewidth]{images/Llama-8b-instruct_cross_symbolic_multiplication_addition_layer_logprob_line_graph.png}
        \caption{$\times$ to $+$}
        \label{fig:add_plot3}
    \end{subfigure}
    \hfill
    \begin{subfigure}[b]{0.24\linewidth}
        \centering
        \includegraphics[width=\linewidth]{images/Llama-8b-instruct_cross_symbolic_division_addition_layer_logprob_line_graph.png}
        \caption{$\div$ to $+$}
        \label{fig:add_plot4}
    \end{subfigure}

    \begin{subfigure}[b]{0.24\linewidth}
        \centering
        \includegraphics[width=\linewidth]{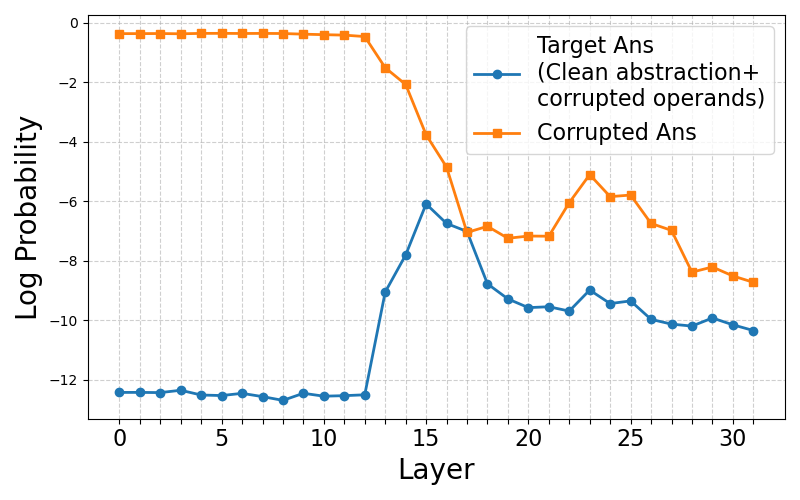}
        \caption{Paired $+$ to $-$}
        \label{fig:add_plot1}
    \end{subfigure}
    \hfill
    \begin{subfigure}[b]{0.24\linewidth}
        \centering
        \includegraphics[width=\linewidth]{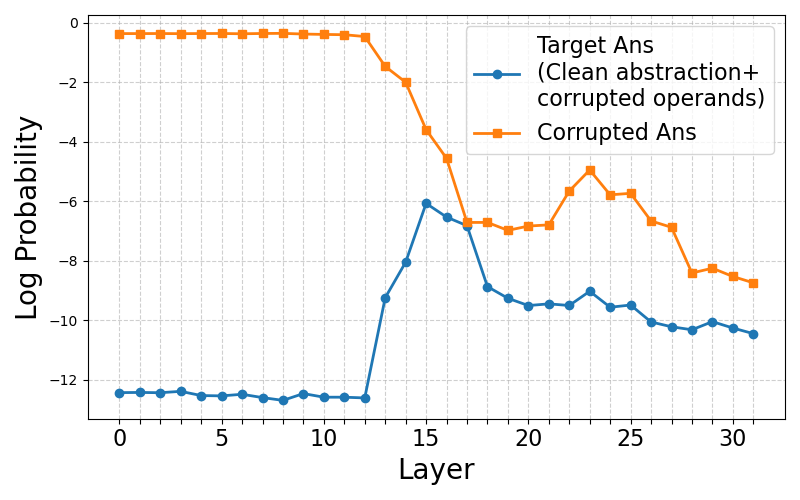}
        \caption{$+$ to $-$}
        \label{fig:add_plot2}
    \end{subfigure}
    \hfill
    \begin{subfigure}[b]{0.24\linewidth}
        \centering
        \includegraphics[width=\linewidth]{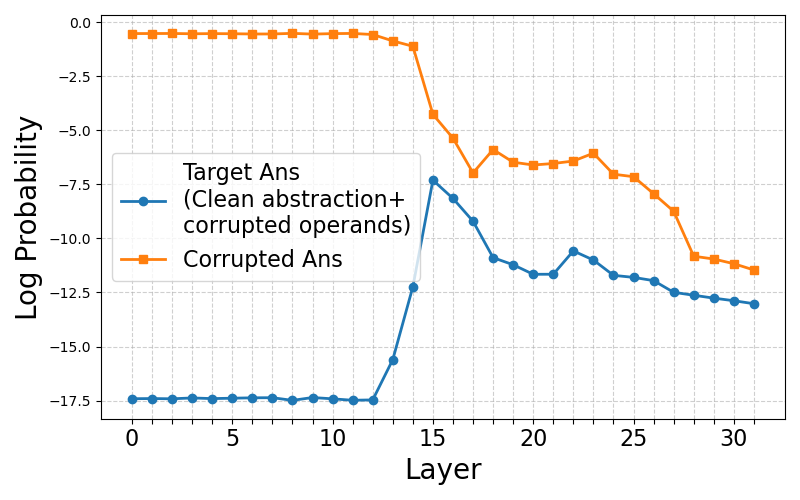}
        \caption{$\times$ to $-$}
        \label{fig:add_plot3}
    \end{subfigure}
    \hfill
    \begin{subfigure}[b]{0.24\linewidth}
        \centering
        \includegraphics[width=\linewidth]{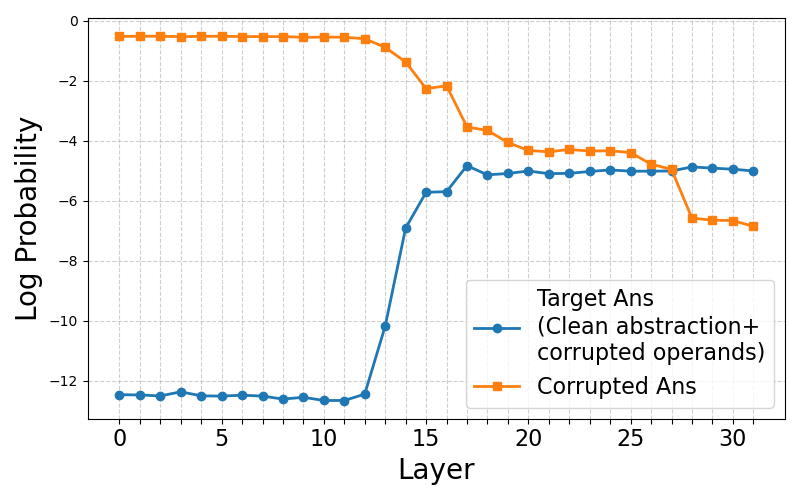}
        \caption{$\div$ to $-$}
        \label{fig:add_plot4}
    \end{subfigure}

    \begin{subfigure}[b]{0.24\linewidth}
        \centering
        \includegraphics[width=\linewidth]{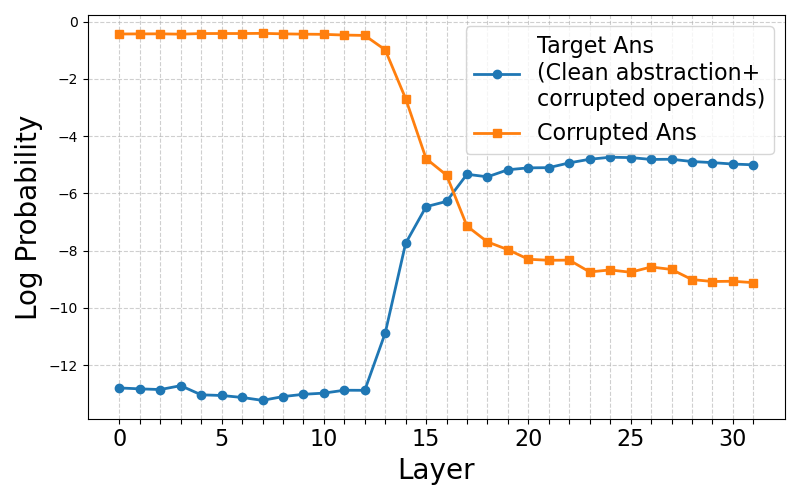}
        \caption{Paired $\div$ to $\times$}
        \label{fig:mult_plot1}
    \end{subfigure}
    \hfill
    \begin{subfigure}[b]{0.24\linewidth}
        \centering
        \includegraphics[width=\linewidth]{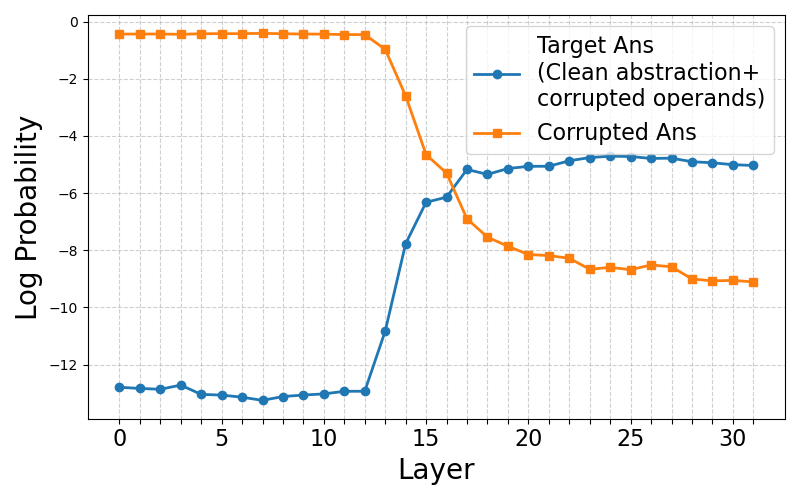}
        \caption{$\div$ to $\times$}
        \label{fig:mult_plot2}
    \end{subfigure}
    \hfill
    \begin{subfigure}[b]{0.24\linewidth}
        \centering
        \includegraphics[width=\linewidth]{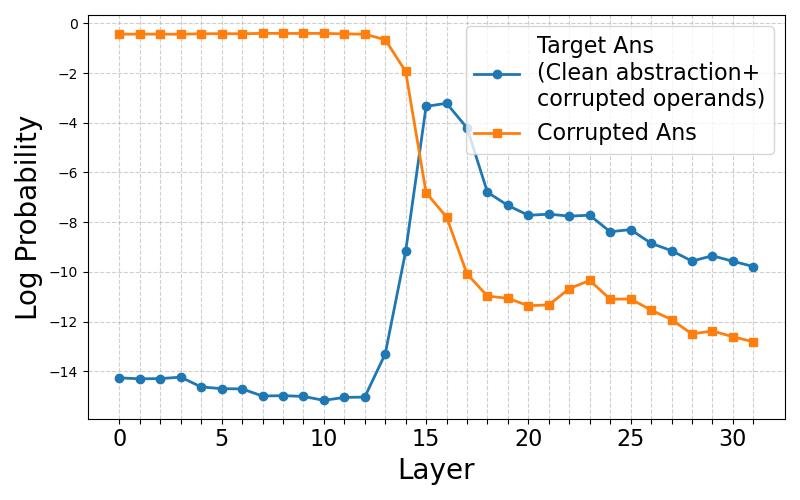}
        \caption{$+$ to $\times$}
        \label{fig:mult_plot3}
    \end{subfigure}
    \hfill
    \begin{subfigure}[b]{0.24\linewidth}
        \centering
        \includegraphics[width=\linewidth]{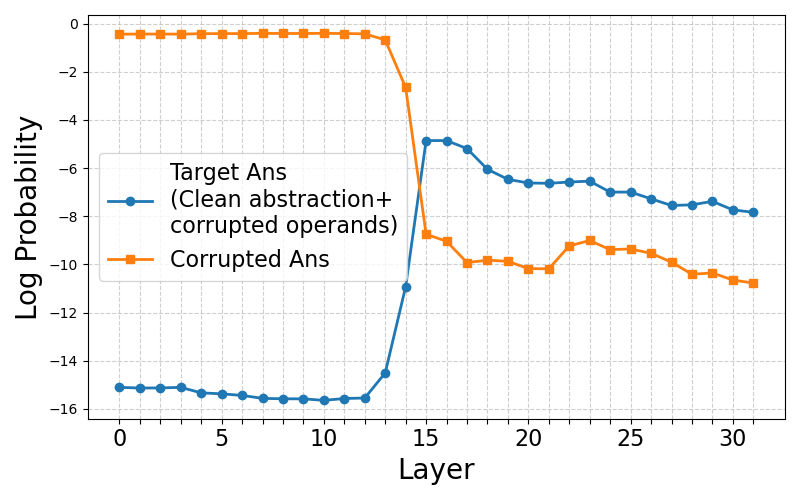}
        \caption{$-$ to $\times$}
        \label{fig:mult_plot4}
    \end{subfigure}

    \begin{subfigure}[b]{0.24\linewidth}
        \centering
        \includegraphics[width=\linewidth]{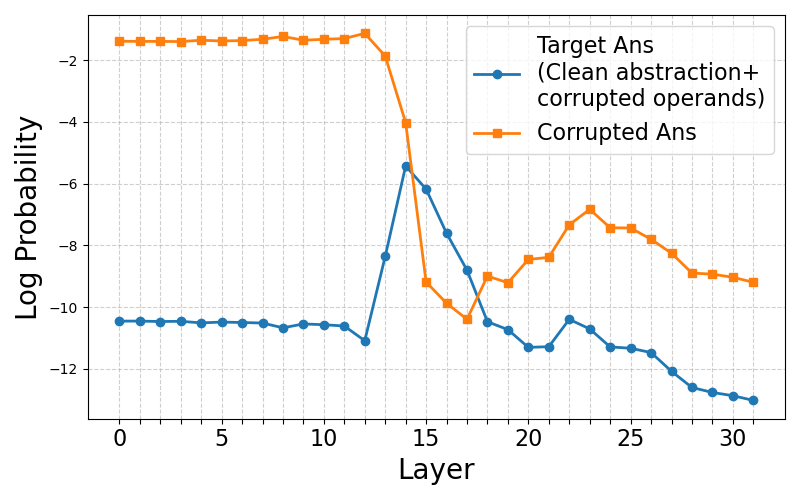}
        \caption{Paired $\times$ to $\div$}
        \label{fig:mult_plot1}
    \end{subfigure}
    \hfill
    \begin{subfigure}[b]{0.24\linewidth}
        \centering
        \includegraphics[width=\linewidth]{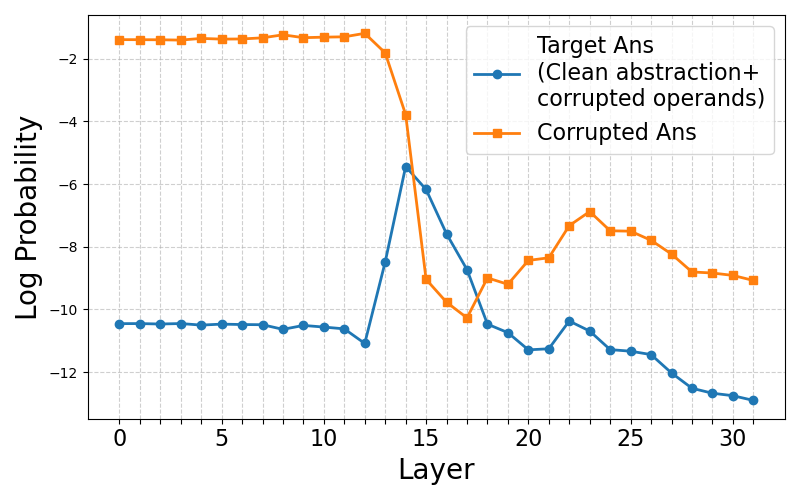}
        \caption{$\times$ to $\div$}
        \label{fig:mult_plot2}
    \end{subfigure}
    \hfill
    \begin{subfigure}[b]{0.24\linewidth}
        \centering
        \includegraphics[width=\linewidth]{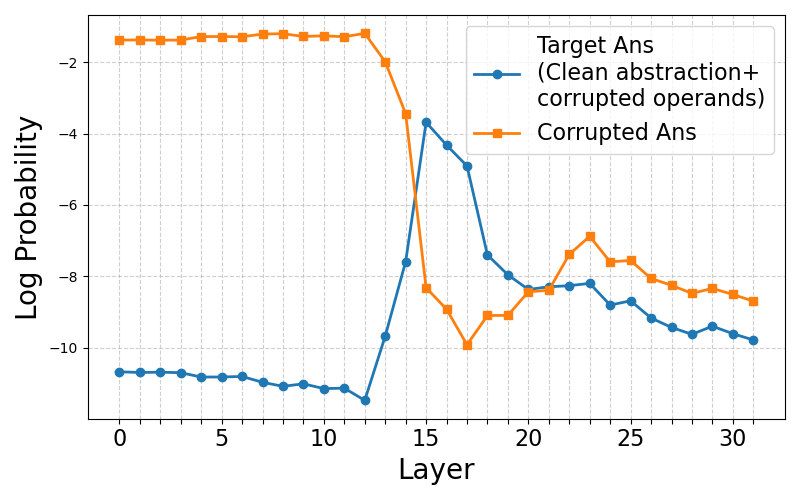}
        \caption{$+$ to $\div$}
        \label{fig:mult_plot3}
    \end{subfigure}
    \hfill
    \begin{subfigure}[b]{0.24\linewidth}
        \centering
        \includegraphics[width=\linewidth]{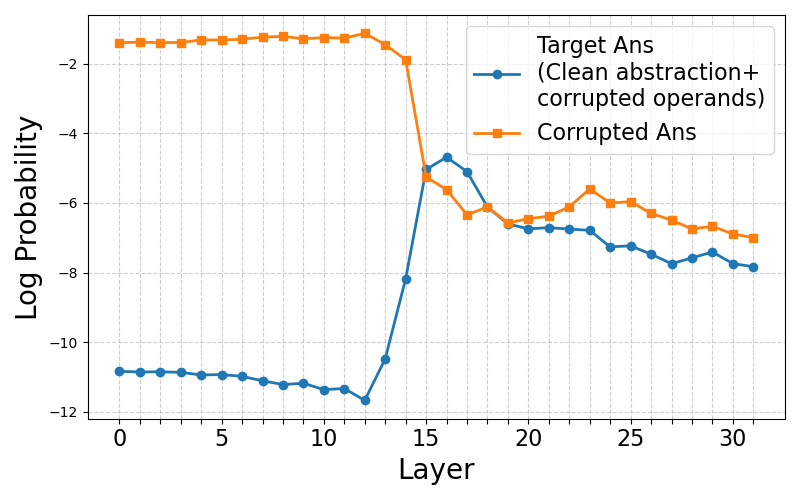}
        \caption{$-$ to $\div$}
        \label{fig:mult_plot4}
    \end{subfigure}
    
    \caption{\textbf{Llama-3 8B} cross-prompt patching for \textbf{symbolic abstraction} results: \textbf{First row}: patching symbolic logic to concrete \textbf{addition}; \textbf{Second row}: patching symbolic logic to concrete \textbf{subtraction}; \textbf{Third row}: patching symbolic logic to concrete \textbf{multiplication}; \textbf{Fourth row}: patching symbolic logic to concrete \textbf{division};}
    \label{fig:Llama3-cross-patching-full}
\end{figure*}

\begin{figure*}[ht]
    \centering
    
    \begin{subfigure}[b]{0.24\linewidth}
        \centering
        \includegraphics[width=\linewidth]{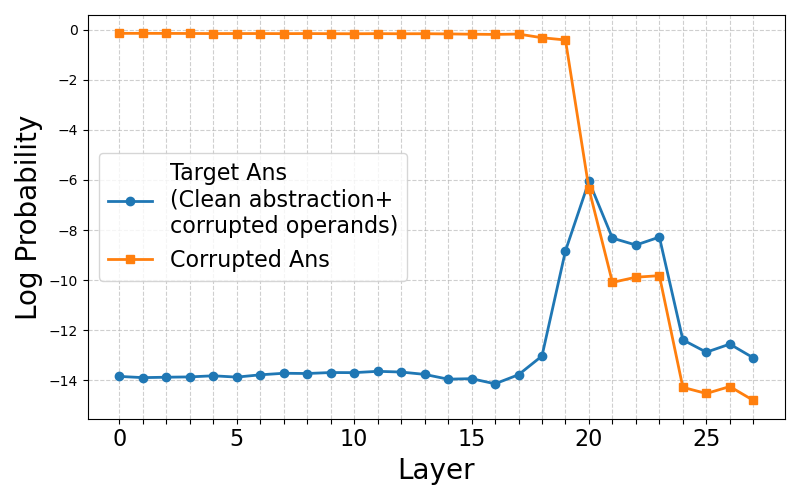}
        \caption{Paired $-$ to $+$}
        \label{fig:add_plot1}
    \end{subfigure}
    \hfill
    \begin{subfigure}[b]{0.24\linewidth}
        \centering
        \includegraphics[width=\linewidth]{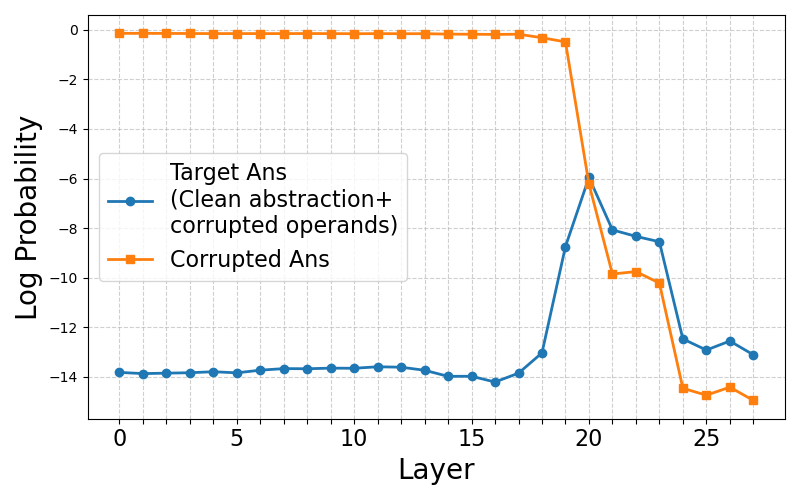}
        \caption{$-$ to $+$}
        \label{fig:add_plot2}
    \end{subfigure}
    \hfill
    \begin{subfigure}[b]{0.24\linewidth}
        \centering
        \includegraphics[width=\linewidth]{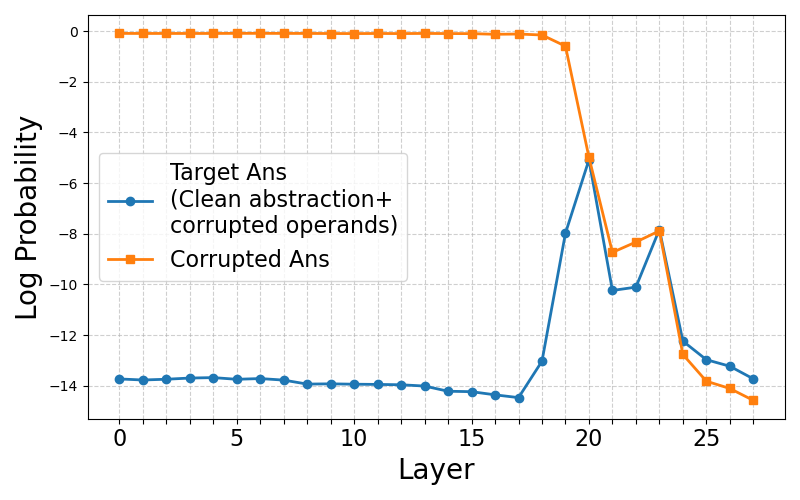}
        \caption{$\times$ to $+$}
        \label{fig:add_plot3}
    \end{subfigure}
    \hfill
    \begin{subfigure}[b]{0.24\linewidth}
        \centering
        \includegraphics[width=\linewidth]{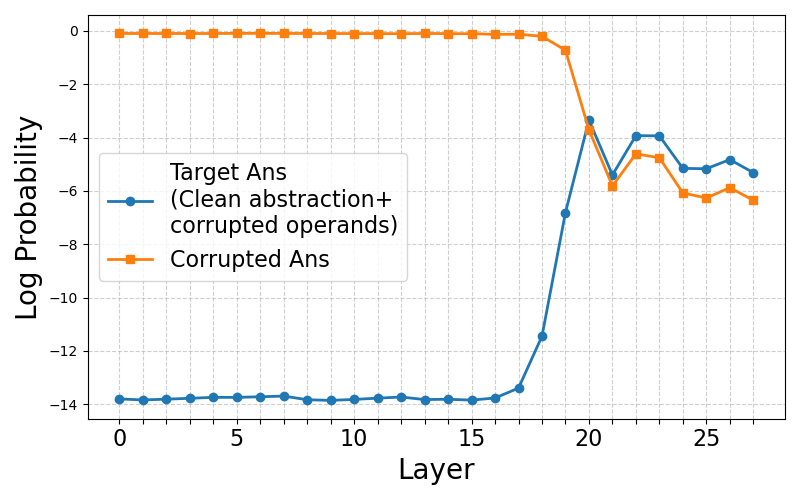}
        \caption{$\div$  to $+$}
        \label{fig:add_plot4}
    \end{subfigure}

    \begin{subfigure}[b]{0.24\linewidth}
        \centering
        \includegraphics[width=\linewidth]{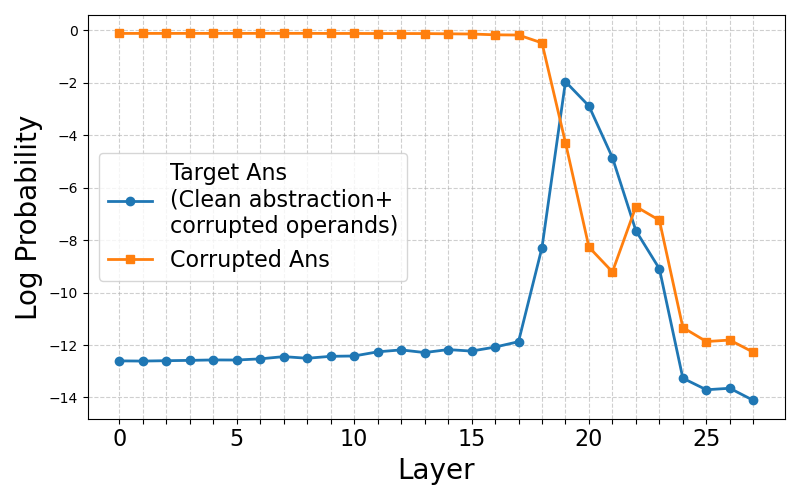}
        \caption{Paired $+$ to $-$}
        \label{fig:add_plot1}
    \end{subfigure}
    \hfill
    \begin{subfigure}[b]{0.24\linewidth}
        \centering
        \includegraphics[width=\linewidth]{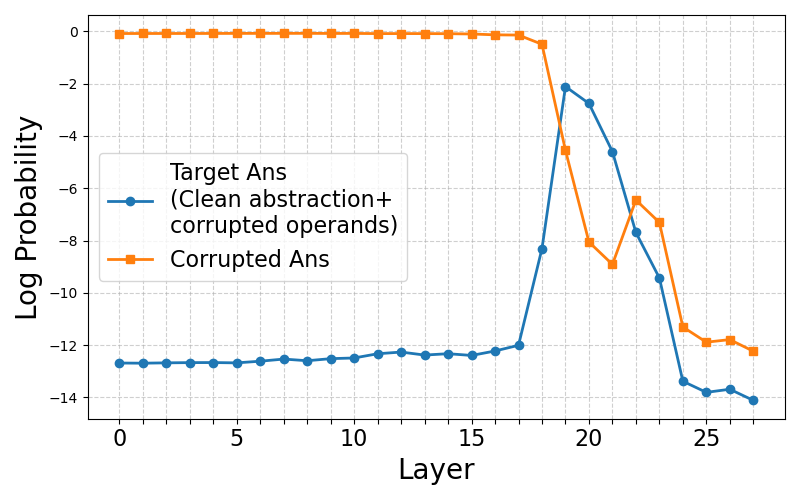}
        \caption{$+$ to $-$}
        \label{fig:add_plot2}
    \end{subfigure}
    \hfill
    \begin{subfigure}[b]{0.24\linewidth}
        \centering
        \includegraphics[width=\linewidth]{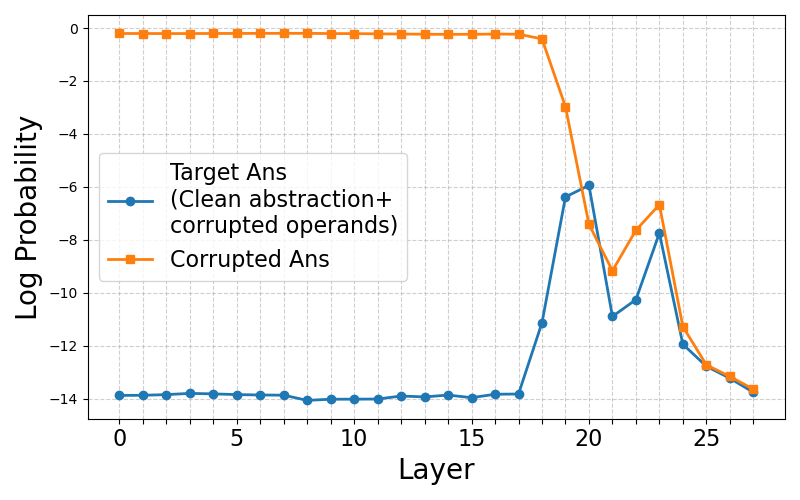}
        \caption{$\times$ to $-$}
        \label{fig:add_plot3}
    \end{subfigure}
    \hfill
    \begin{subfigure}[b]{0.24\linewidth}
        \centering
        \includegraphics[width=\linewidth]{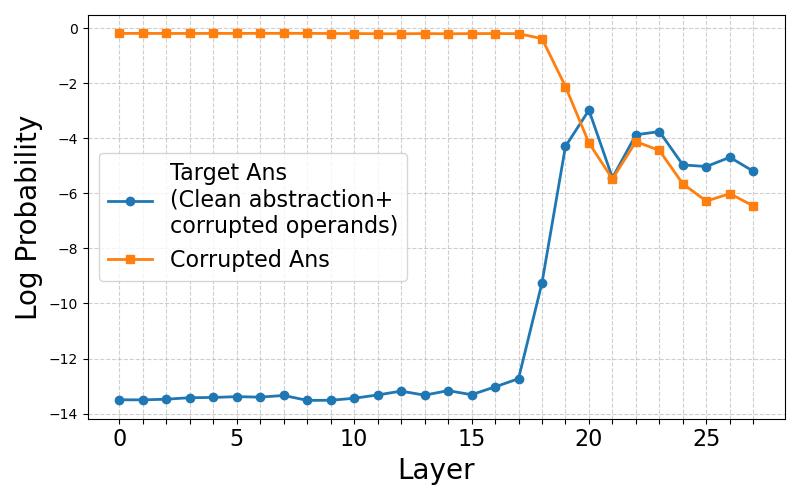}
        \caption{$\div$  to $-$}
        \label{fig:add_plot4}
    \end{subfigure}

    \begin{subfigure}[b]{0.24\linewidth}
        \centering
        \includegraphics[width=\linewidth]{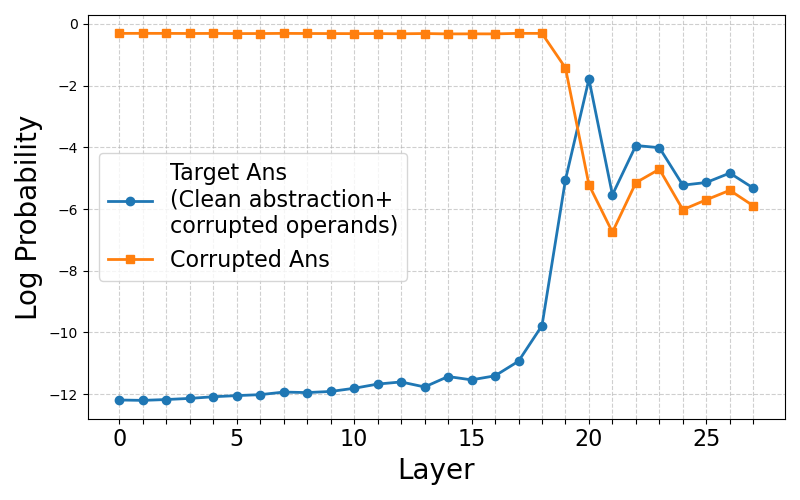}
        \caption{Paired $\div$  to $\times$}
        \label{fig:mult_plot1}
    \end{subfigure}
    \hfill
    \begin{subfigure}[b]{0.24\linewidth}
        \centering
        \includegraphics[width=\linewidth]{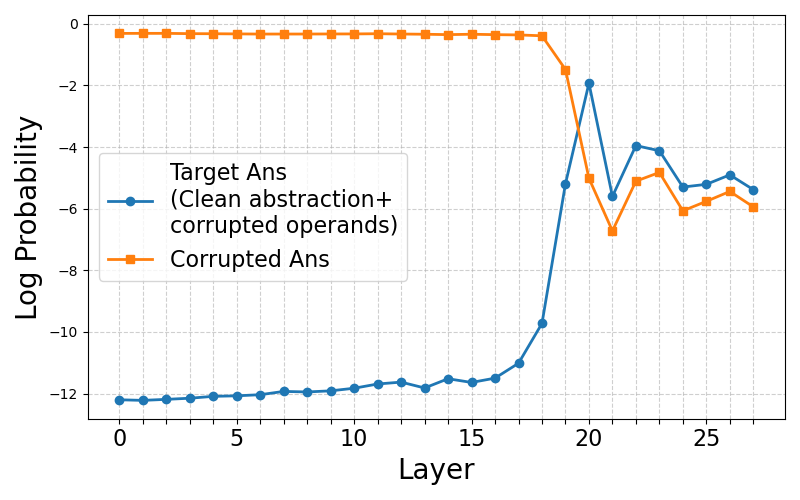}
        \caption{$\div$  to $\times$}
        \label{fig:mult_plot2}
    \end{subfigure}
    \hfill
    \begin{subfigure}[b]{0.24\linewidth}
        \centering
        \includegraphics[width=\linewidth]{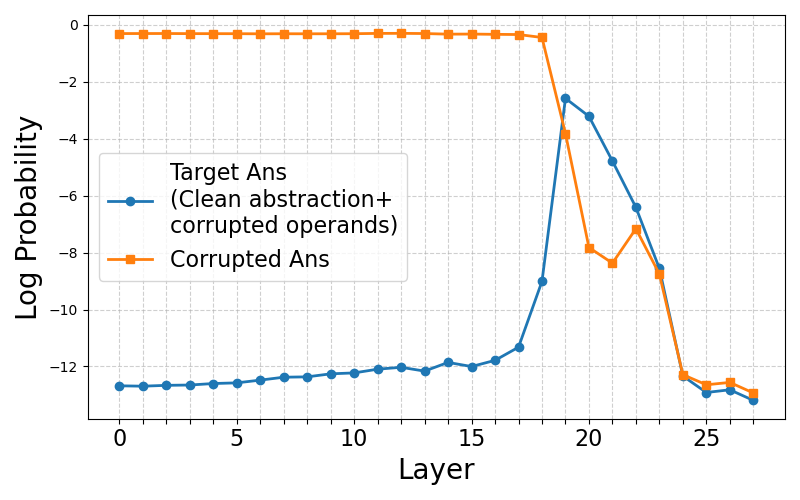}
        \caption{$+$ to $\times$}
        \label{fig:mult_plot3}
    \end{subfigure}
    \hfill
    \begin{subfigure}[b]{0.24\linewidth}
        \centering
        \includegraphics[width=\linewidth]{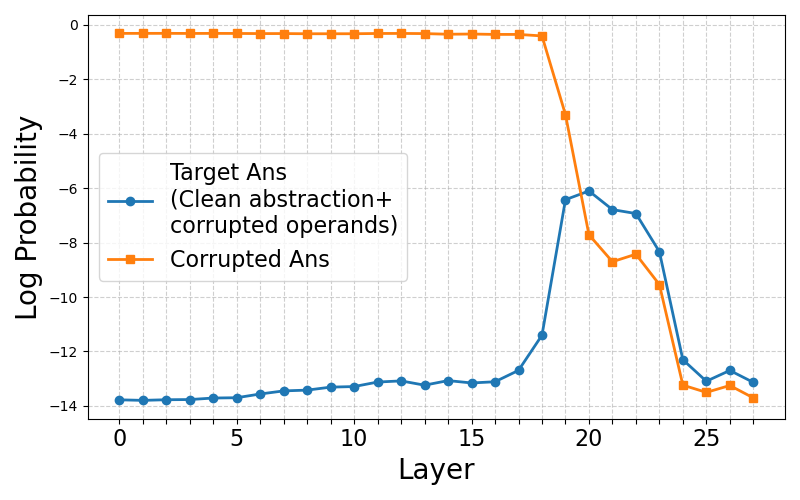}
        \caption{$-$ to $\times$}
        \label{fig:mult_plot4}
    \end{subfigure}

    \begin{subfigure}[b]{0.24\linewidth}
        \centering
        \includegraphics[width=\linewidth]{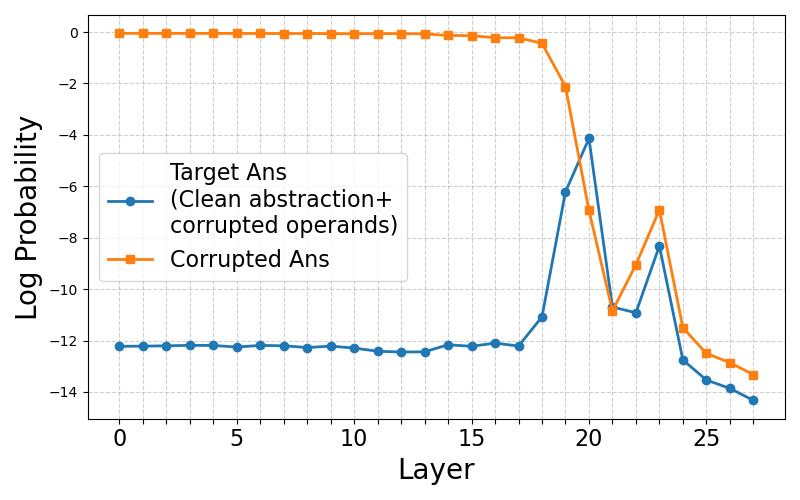}
        \caption{Paired $\times$ to $\div$  }
        \label{fig:mult_plot1}
    \end{subfigure}
    \hfill
    \begin{subfigure}[b]{0.24\linewidth}
        \centering
        \includegraphics[width=\linewidth]{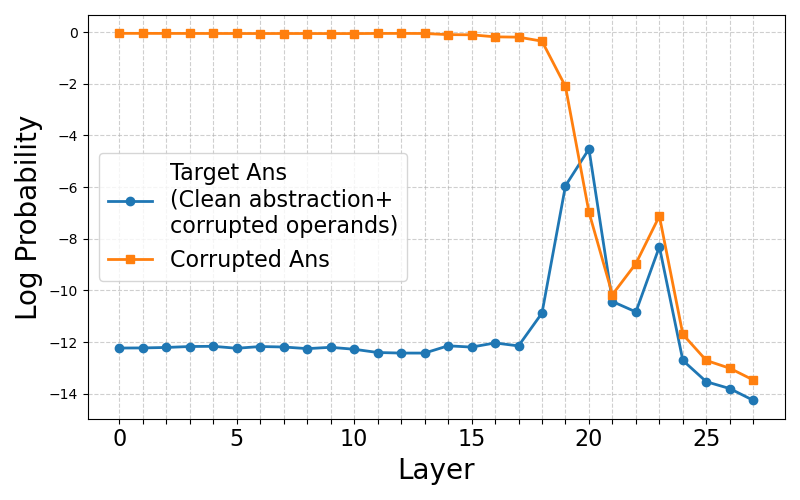}
        \caption{$\times$ to ÷}
        \label{fig:mult_plot2}
    \end{subfigure}
    \hfill
    \begin{subfigure}[b]{0.24\linewidth}
        \centering
        \includegraphics[width=\linewidth]{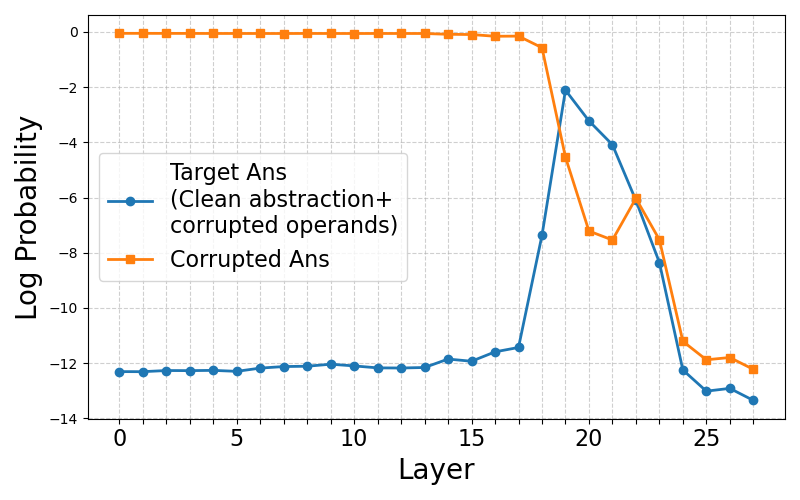}
        \caption{$+$ to ÷}
        \label{fig:mult_plot3}
    \end{subfigure}
    \hfill
    \begin{subfigure}[b]{0.24\linewidth}
        \centering
        \includegraphics[width=\linewidth]{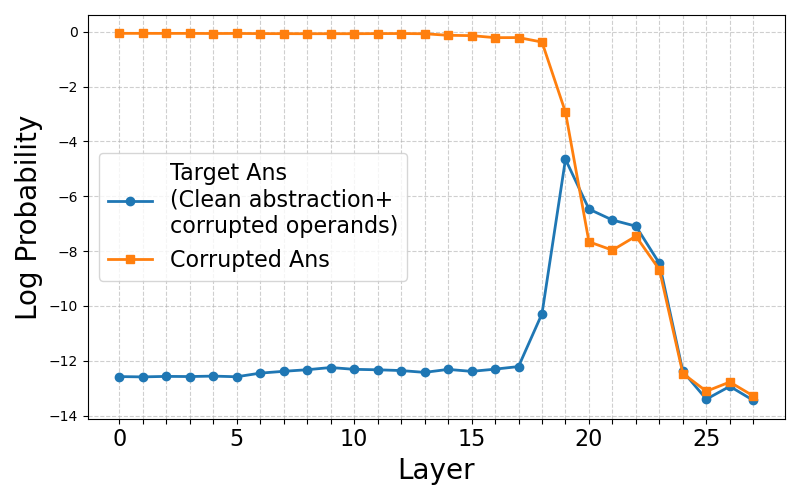}
        \caption{$-$ to ÷}
        \label{fig:mult_plot4}
    \end{subfigure}
    
    \caption{\textbf{Qwen-7b} cross-prompt patching for \textbf{symbolic abstraction} results: \textbf{First row}: patching symbolic logic to concrete \textbf{addition}; \textbf{Second row}: patching symbolic logic to concrete \textbf{subtraction}; \textbf{Third row}: patching symbolic logic to concrete \textbf{multiplication}; \textbf{Fourth row}: patching symbolic logic to concrete \textbf{division};}
    \label{fig:Qwen7b-cross-patching-full}
\end{figure*}

\begin{figure*}[ht]
    \centering
    
    \begin{subfigure}[b]{0.24\linewidth}
        \centering
        \includegraphics[width=\linewidth]{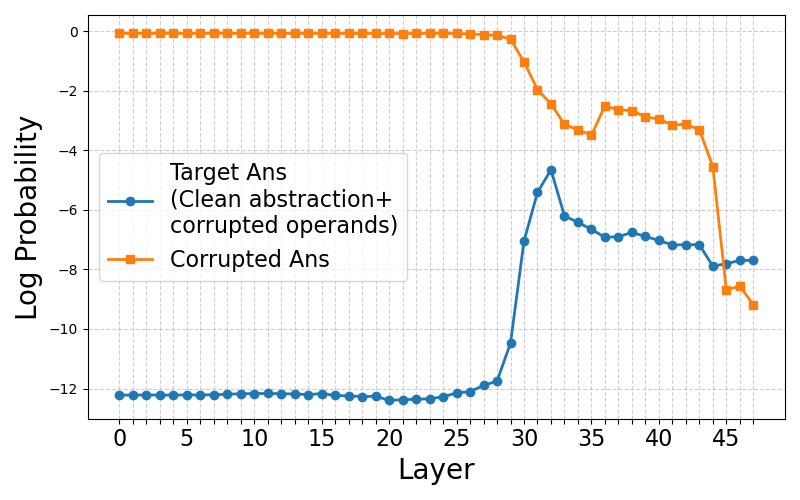}
        \caption{Paired $-$ to $+$}
        \label{fig:add_plot1}
    \end{subfigure}
    \hfill
    \begin{subfigure}[b]{0.24\linewidth}
        \centering
        \includegraphics[width=\linewidth]{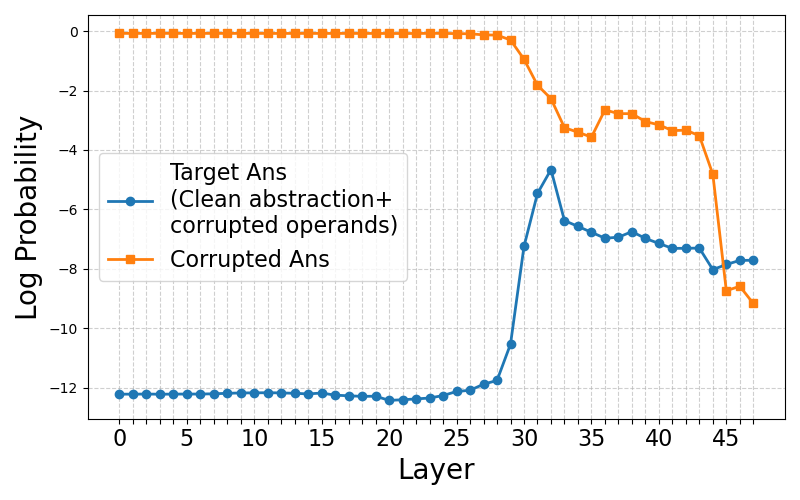}
        \caption{$-$ to $+$}
        \label{fig:add_plot2}
    \end{subfigure}
    \hfill
    \begin{subfigure}[b]{0.24\linewidth}
        \centering
        \includegraphics[width=\linewidth]{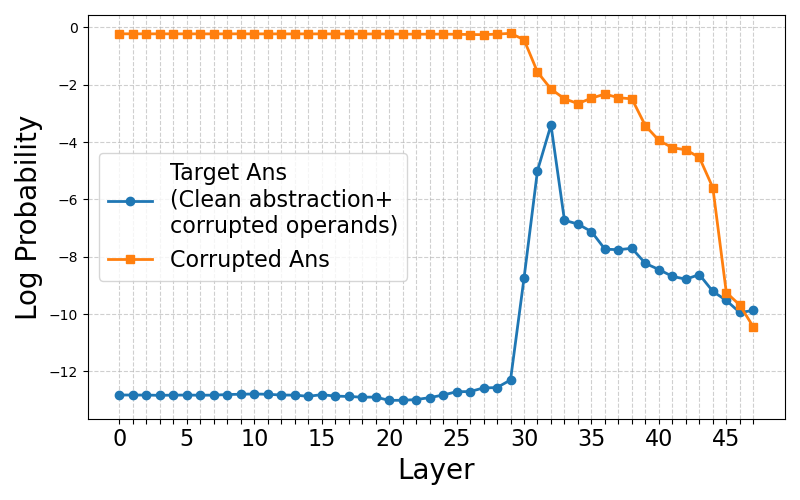}
        \caption{$\times$ to $+$}
        \label{fig:add_plot3}
    \end{subfigure}
    \hfill
    \begin{subfigure}[b]{0.24\linewidth}
        \centering
        \includegraphics[width=\linewidth]{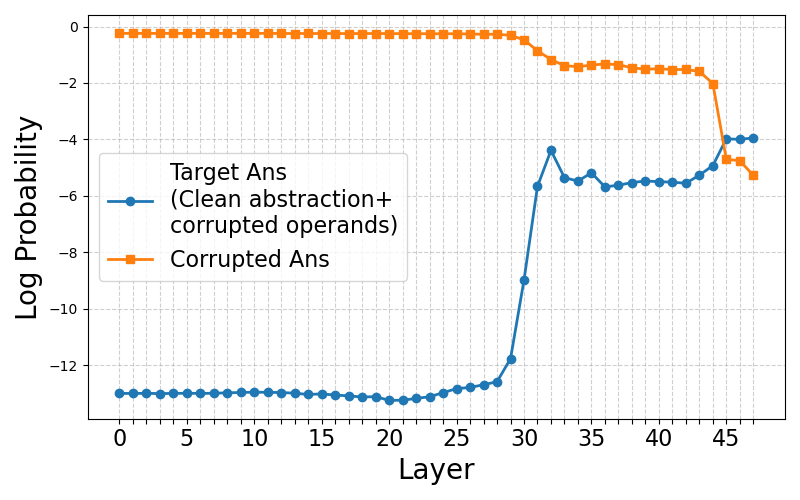}
        \caption{$\div$  to $+$}
        \label{fig:add_plot4}
    \end{subfigure}

    \begin{subfigure}[b]{0.24\linewidth}
        \centering
        \includegraphics[width=\linewidth]{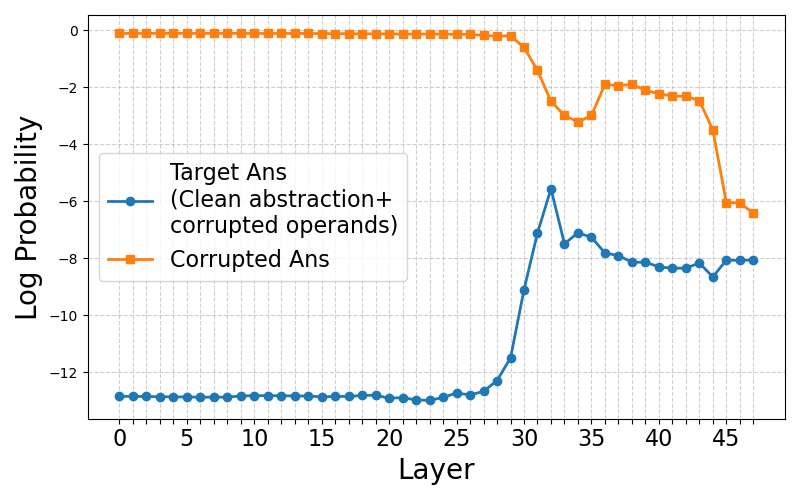}
        \caption{Paired $+$ to $-$}
        \label{fig:add_plot1}
    \end{subfigure}
    \hfill
    \begin{subfigure}[b]{0.24\linewidth}
        \centering
        \includegraphics[width=\linewidth]{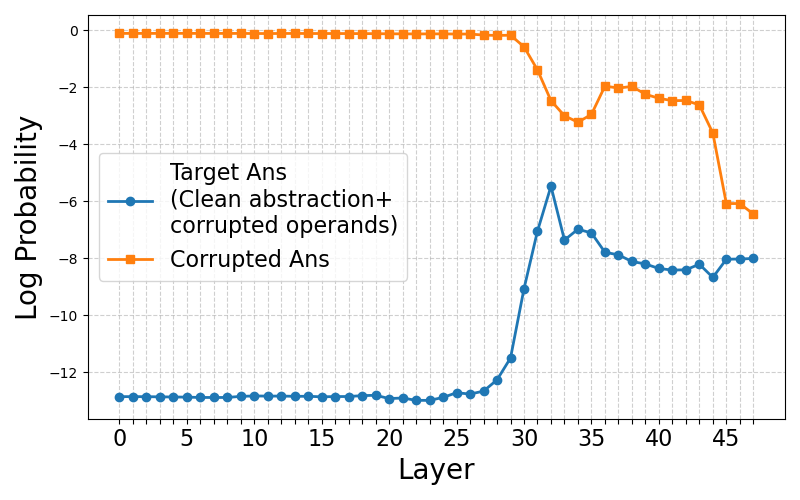}
        \caption{$+$ to $-$}
        \label{fig:add_plot2}
    \end{subfigure}
    \hfill
    \begin{subfigure}[b]{0.24\linewidth}
        \centering
        \includegraphics[width=\linewidth]{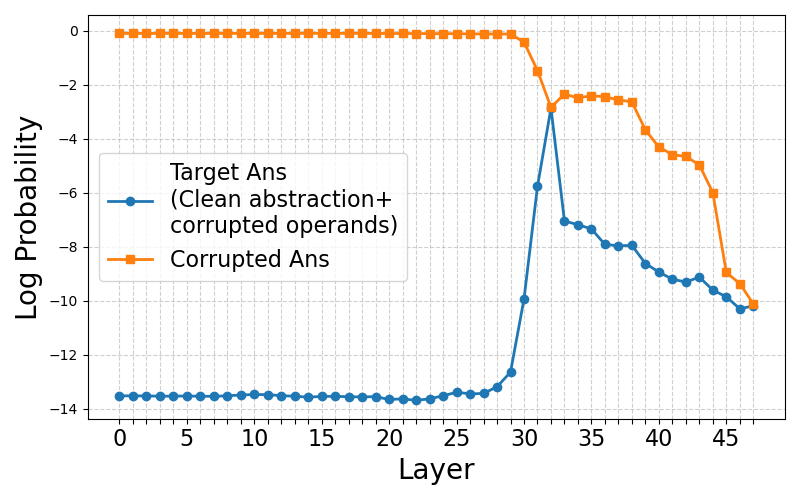}
        \caption{$\times$ to $-$}
        \label{fig:add_plot3}
    \end{subfigure}
    \hfill
    \begin{subfigure}[b]{0.24\linewidth}
        \centering
        \includegraphics[width=\linewidth]{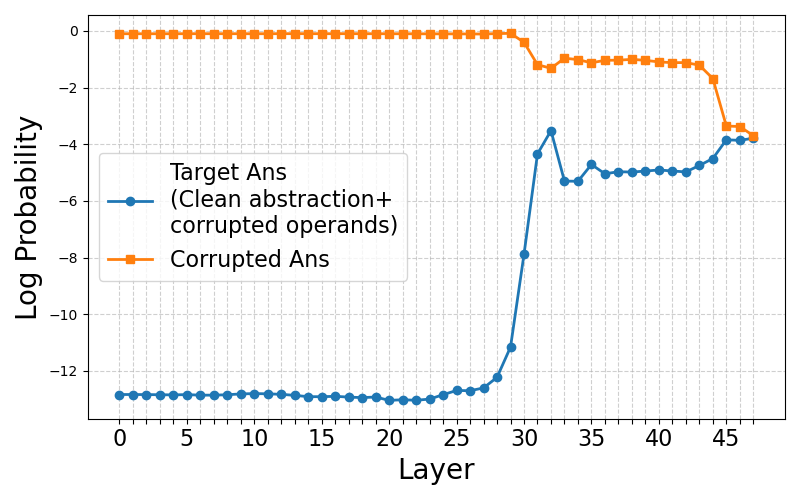}
        \caption{$\div$  to $-$}
        \label{fig:add_plot4}
    \end{subfigure}

    \begin{subfigure}[b]{0.24\linewidth}
        \centering
        \includegraphics[width=\linewidth]{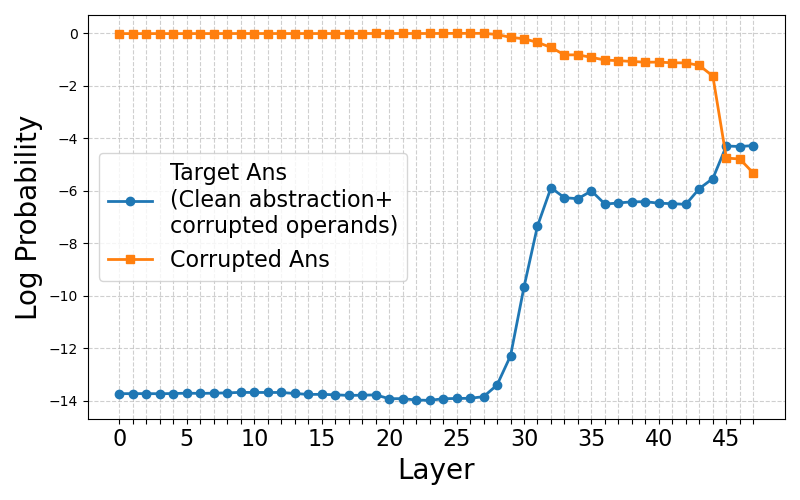}
        \caption{Paired $\div$  to $\times$}
        \label{fig:mult_plot1}
    \end{subfigure}
    \hfill
    \begin{subfigure}[b]{0.24\linewidth}
        \centering
        \includegraphics[width=\linewidth]{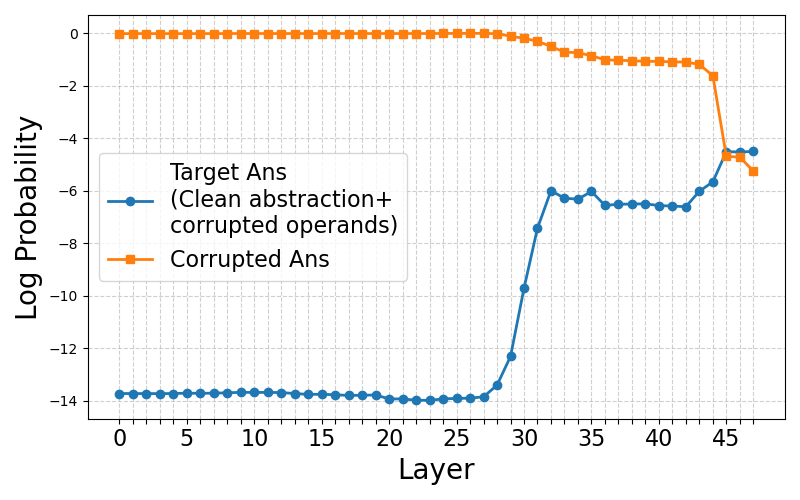}
        \caption{$\div$  to $\times$}
        \label{fig:mult_plot2}
    \end{subfigure}
    \hfill
    \begin{subfigure}[b]{0.24\linewidth}
        \centering
        \includegraphics[width=\linewidth]{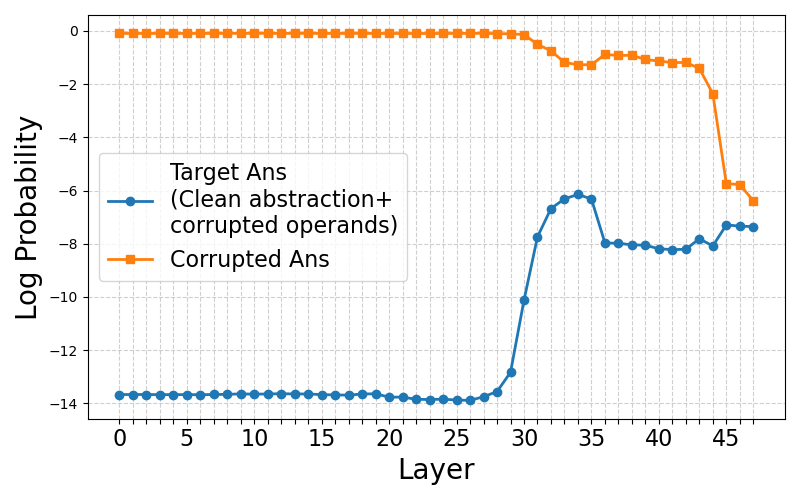}
        \caption{$+$ to $\times$}
        \label{fig:mult_plot3}
    \end{subfigure}
    \hfill
    \begin{subfigure}[b]{0.24\linewidth}
        \centering
        \includegraphics[width=\linewidth]{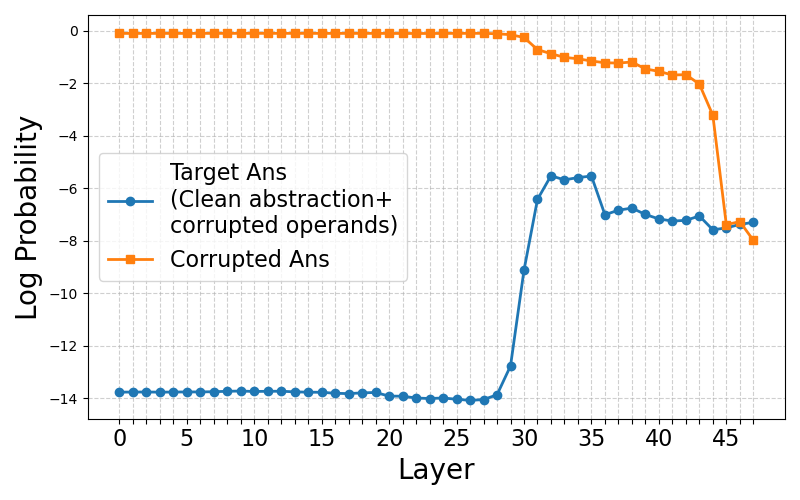}
        \caption{$-$ to $\times$}
        \label{fig:mult_plot4}
    \end{subfigure}

    \begin{subfigure}[b]{0.24\linewidth}
        \centering
        \includegraphics[width=\linewidth]{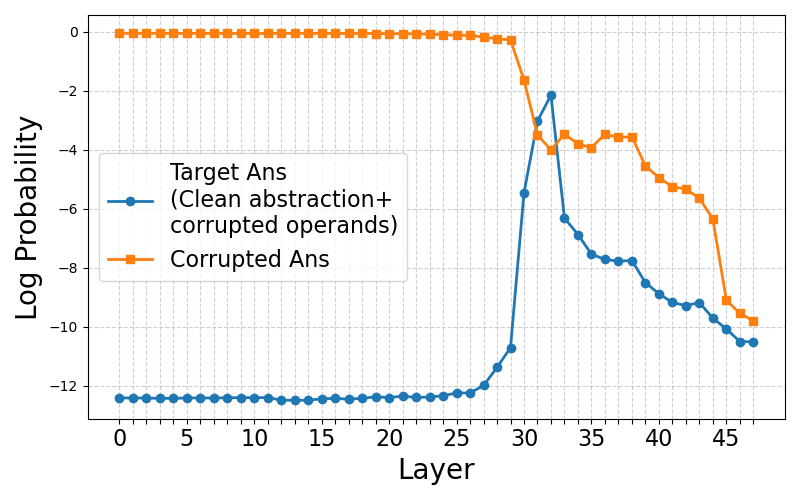}
        \caption{Paired $\times$ to $\div$  }
        \label{fig:mult_plot1}
    \end{subfigure}
    \hfill
    \begin{subfigure}[b]{0.24\linewidth}
        \centering
        \includegraphics[width=\linewidth]{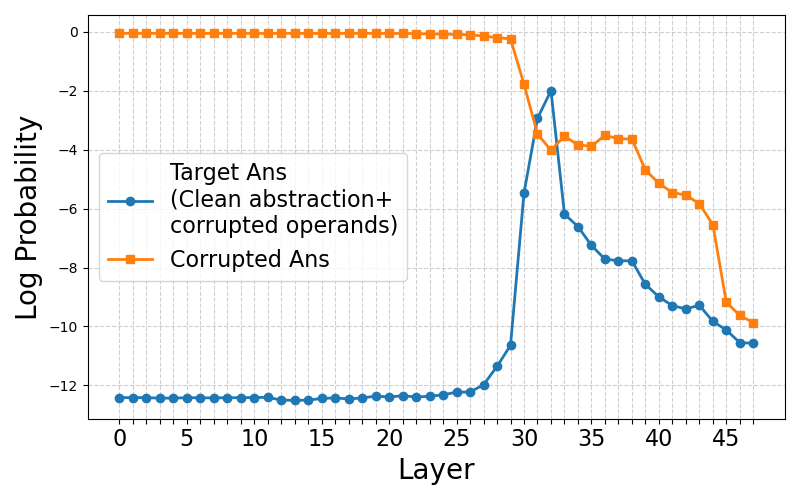}
        \caption{$\times$ to ÷}
        \label{fig:mult_plot2}
    \end{subfigure}
    \hfill
    \begin{subfigure}[b]{0.24\linewidth}
        \centering
        \includegraphics[width=\linewidth]{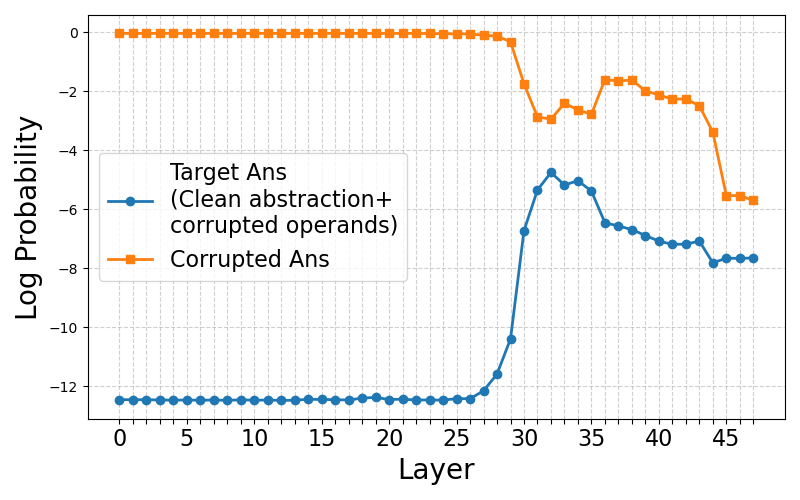}
        \caption{$+$ to ÷}
        \label{fig:mult_plot3}
    \end{subfigure}
    \hfill
    \begin{subfigure}[b]{0.24\linewidth}
        \centering
        \includegraphics[width=\linewidth]{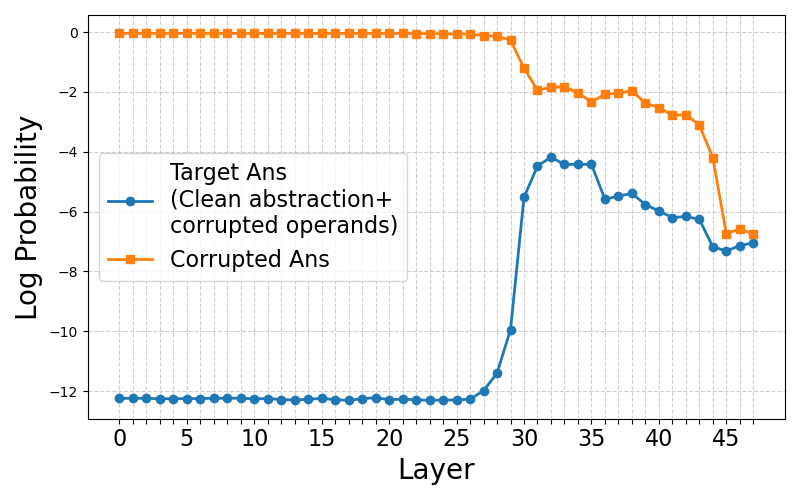}
        \caption{$-$ to ÷}
        \label{fig:mult_plot4}
    \end{subfigure}
    
    \caption{\textbf{Qwen-14b} cross-prompt patching for \textbf{symbolic abstraction} results: \textbf{First row}: patching symbolic logic to concrete \textbf{addition}; \textbf{Second row}: patching symbolic logic to concrete \textbf{subtraction}; \textbf{Third row}: patching symbolic logic to concrete \textbf{multiplication}; \textbf{Fourth row}: patching symbolic logic to concrete \textbf{division};}
    \label{fig:Qwen14b-cross-patching-full}
\end{figure*}

\paragraph{Other Models}
See Figure~\ref{fig:Llama3-cross-patching-full}, Figure~\ref{fig:Qwen7b-cross-patching-full}, and Figure~\ref{fig:Qwen14b-cross-patching-full} for symbolic abstraction cross-prompt patching results (for single operators: $+, -, \times, \div$) on Llama3 8B, Qwen 2.5 7B, and Qwen 2.5 14B, respectively. The results are consistent across models: the likelihood of the target answer peaks at the abstraction stage, while the likelihood of the corrupted answer drops significantly starting from the same stage.

\begin{figure*}[ht]
    \centering
    \begin{subfigure}[b]{0.24\linewidth}
        \centering
        \includegraphics[width=\linewidth]{images/llama-8b-instruct_cross_concrete_addition_substraction_layer_logprob_line_graph.png}
        \caption{Cl: $a+b$; Cor: $a-b$}
        \label{fig:mult_plot1}
    \end{subfigure}
    \hfill
    \begin{subfigure}[b]{0.24\linewidth}
        \centering
        \includegraphics[width=\linewidth]{images/llama-8b-instruct_cross_concrete_substraction_addition_layer_logprob_line_graph.png}
        \caption{Cl: $a-b$; Cor: $a+b$}
        \label{fig:mult_plot2}
    \end{subfigure}
    \hfill
    \begin{subfigure}[b]{0.24\linewidth}
        \centering
        \includegraphics[width=\linewidth]{images/llama-8b-instruct_cross_concrete_multiplication_division_layer_logprob_line_graph.png}
        \caption{Cl: $a\times b$; Cor: $a\div b$}
        \label{fig:mult_plot3}
    \end{subfigure}
    \hfill
    \begin{subfigure}[b]{0.24\linewidth}
        \centering
        \includegraphics[width=\linewidth]{images/llama-8b-instruct_cross_concrete_division_multiplication_layer_logprob_line_graph.png}
        \caption{Cl: $a\div b$; Cor: $a\times b$}
        \label{fig:mult_plot4}
    \end{subfigure}
    
    \begin{subfigure}[b]{0.24\linewidth}
        \centering
        \includegraphics[width=\linewidth]{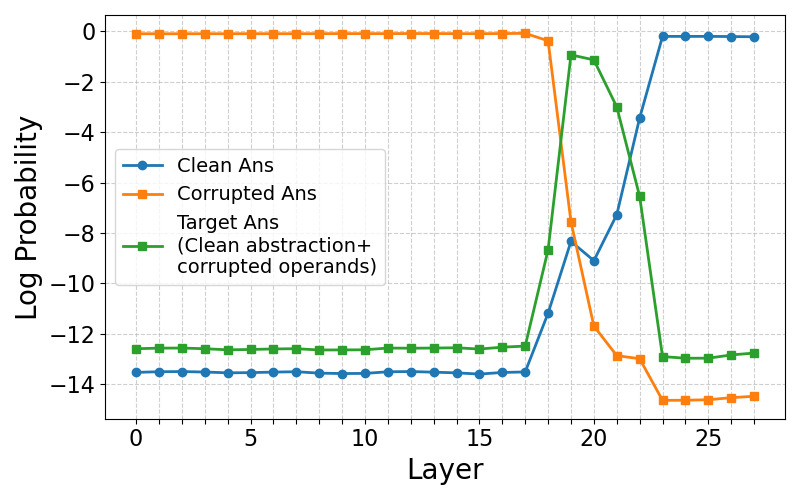}
        \caption{Cl: $a+b$; Cor: $a-b$}
        \label{fig:mult_plot1}
    \end{subfigure}
    \hfill
    \begin{subfigure}[b]{0.24\linewidth}
        \centering
        \includegraphics[width=\linewidth]{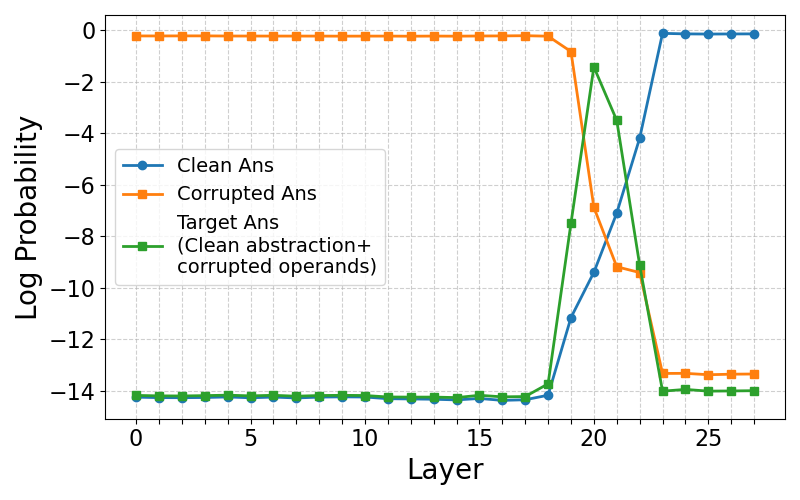}
        \caption{Cl: $a-b$; Cor: $a+b$}
        \label{fig:mult_plot2}
    \end{subfigure}
    \hfill
    \begin{subfigure}[b]{0.24\linewidth}
        \centering
        \includegraphics[width=\linewidth]{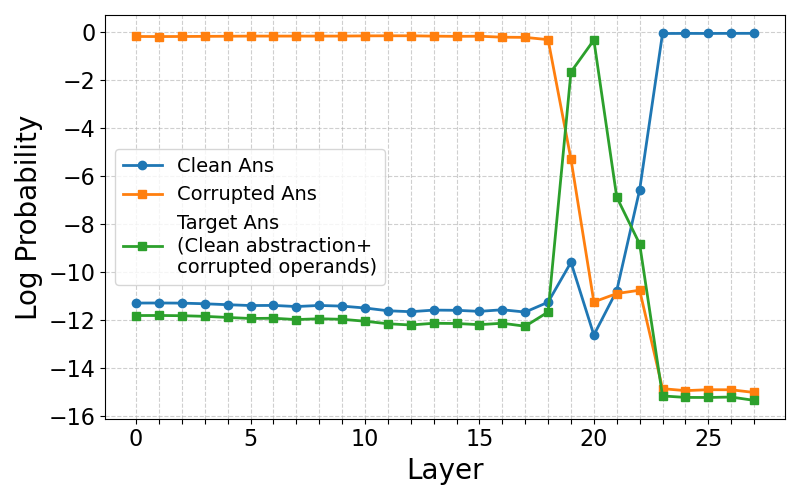}
        \caption{Cl: $a\times b$; Cor: $a\div b$}
        \label{fig:mult_plot3}
    \end{subfigure}
    \hfill
    \begin{subfigure}[b]{0.24\linewidth}
        \centering
        \includegraphics[width=\linewidth]{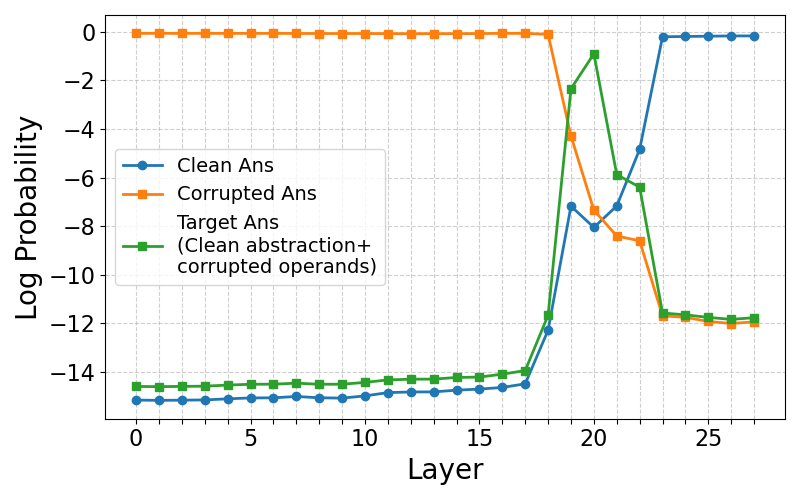}
        \caption{Cl: $a\div b$; Cor: $a\times b$}
        \label{fig:mult_plot4}
    \end{subfigure}

    \begin{subfigure}[b]{0.24\linewidth}
        \centering
        \includegraphics[width=\linewidth]{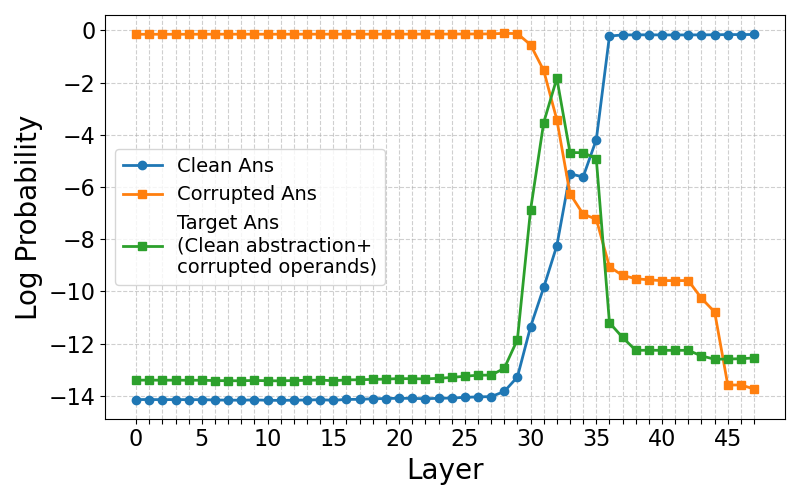}
        \caption{Cl: $a+b$; Cor: $a-b$}
        \label{fig:mult_plot1}
    \end{subfigure}
    \hfill
    \begin{subfigure}[b]{0.24\linewidth}
        \centering
        \includegraphics[width=\linewidth]{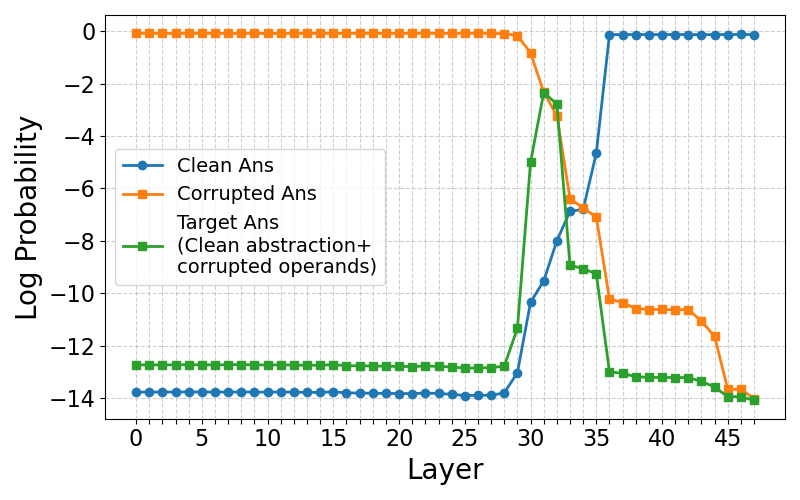}
        \caption{Cl: $a-b$; Cor: $a+b$}
        \label{fig:mult_plot2}
    \end{subfigure}
    \hfill
    \begin{subfigure}[b]{0.24\linewidth}
        \centering
        \includegraphics[width=\linewidth]{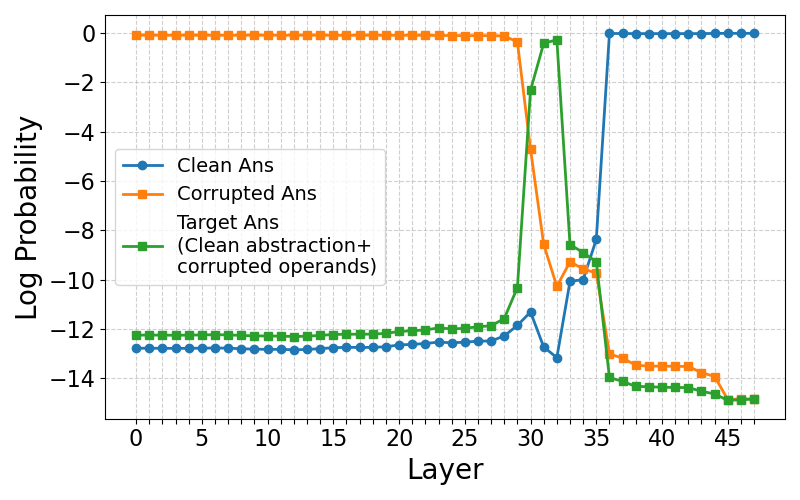}
        \caption{Cl: $a\times b$; Cor: $a\div b$}
        \label{fig:mult_plot3}
    \end{subfigure}
    \hfill
    \begin{subfigure}[b]{0.24\linewidth}
        \centering
        \includegraphics[width=\linewidth]{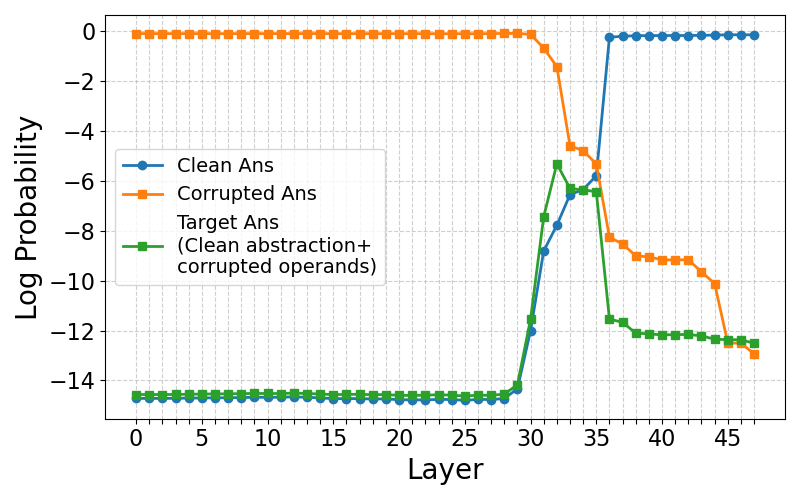}
        \caption{Cl: $a\div b$; Cor: $a\times b$}
        \label{fig:mult_plot4}
    \end{subfigure}

    \caption{Cross-prompt patching results for \textbf{numerical abstraction}. \textbf{First row:} results for \textbf{Llama-3 8B} with corresponding clean and corrupted run. \textbf{Second row:} results for \textbf{Qwen2.5 7B} with corresponding clean and corrupted run. \textbf{Third row:} results for \textbf{Qwen2.5 14B} with corresponding clean and corrupted run.}
    \label{fig:concrete-cross-patching-full}
\end{figure*}

See Figure~\ref{fig:concrete-cross-patching-full} for numerical abstraction cross-prompt patching results on Llama3 8B, Qwen 2.5 7B, and Qwen 2.5 14B. We observe consistent trends across all models: the probability of the target answer begins to rise at the onset of the abstraction stage and peaks by its end. Meanwhile, the clean answer probability increases steadily throughout the abstraction stage, reaching a log-probability of 0 at the start of the computation stage.



\paragraph{Two-Operator Dataset}
For the two-operator dataset, we report results only for Qwen 2.5 7B and Qwen 2.5 14B, as Llama-3 8B performs poorly on this setting, achieving only 16.5\% accuracy.

\begin{figure*}[ht]
    \centering
    
    \begin{subfigure}[b]{0.24\linewidth}
        \centering
        \includegraphics[width=\linewidth]{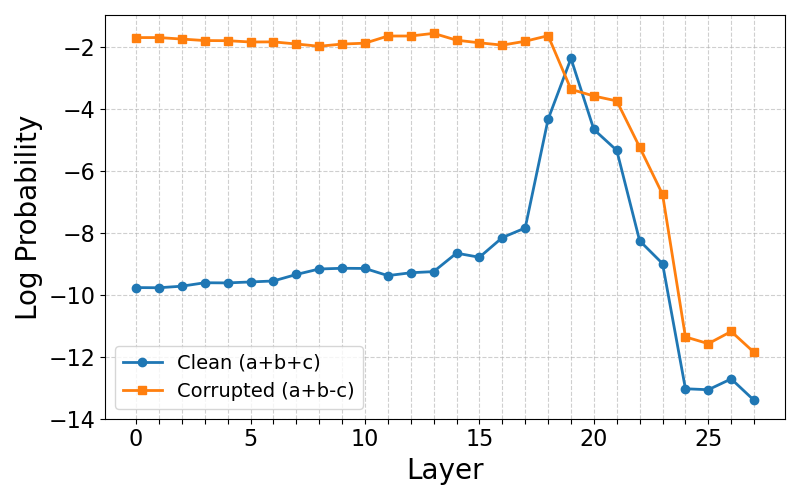}
        \caption{$x+y+z$ to $a+b-c$}
        \label{fig:add_plot1}
    \end{subfigure}
    \hfill
    \begin{subfigure}[b]{0.24\linewidth}
        \centering
        \includegraphics[width=\linewidth]{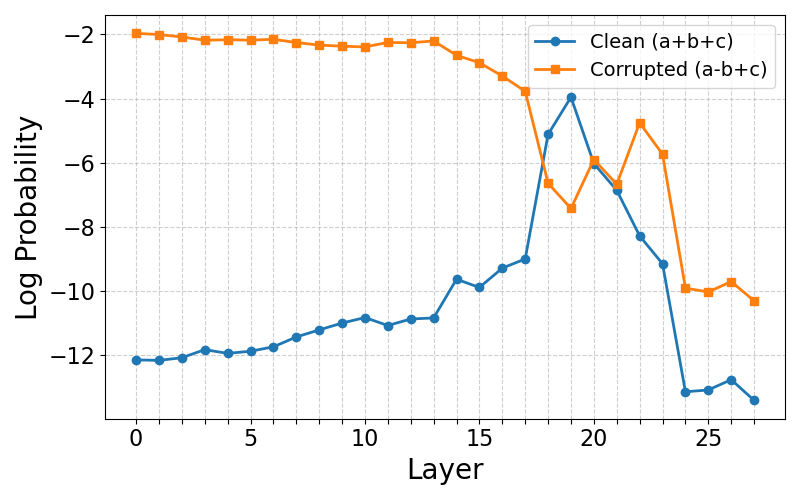}
        \caption{$x+y+z$ to $a-b+c$}
        \label{fig:add_plot2}
    \end{subfigure}
    \hfill
    \begin{subfigure}[b]{0.24\linewidth}
        \centering
        \includegraphics[width=\linewidth]{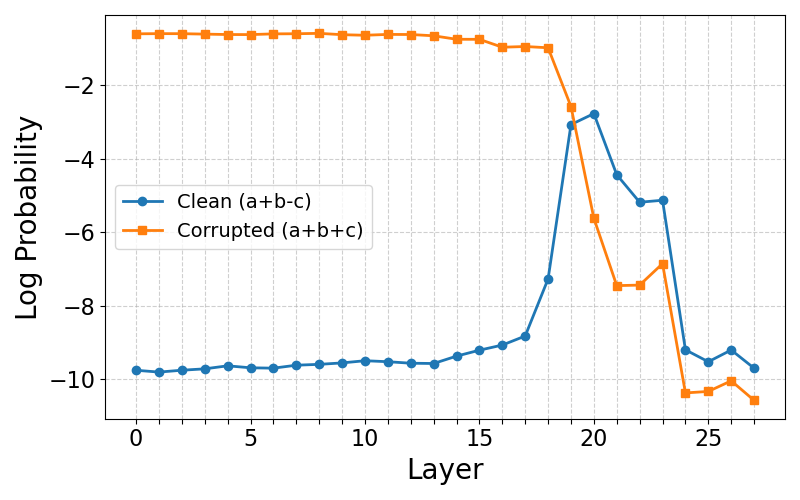}
        \caption{$x+y-z$ to $a+b+c$}
        \label{fig:add_plot3}
    \end{subfigure}
    \hfill
    \begin{subfigure}[b]{0.24\linewidth}
        \centering
        \includegraphics[width=\linewidth]{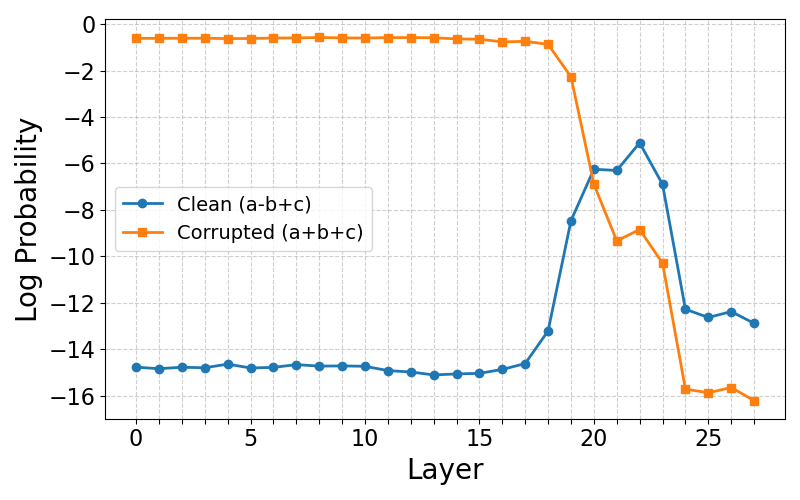}
        \caption{$x-y+z$ to $a+b+c$}
        \label{fig:add_plot4}
    \end{subfigure}

    \begin{subfigure}[b]{0.24\linewidth}
        \centering
        \includegraphics[width=\linewidth]{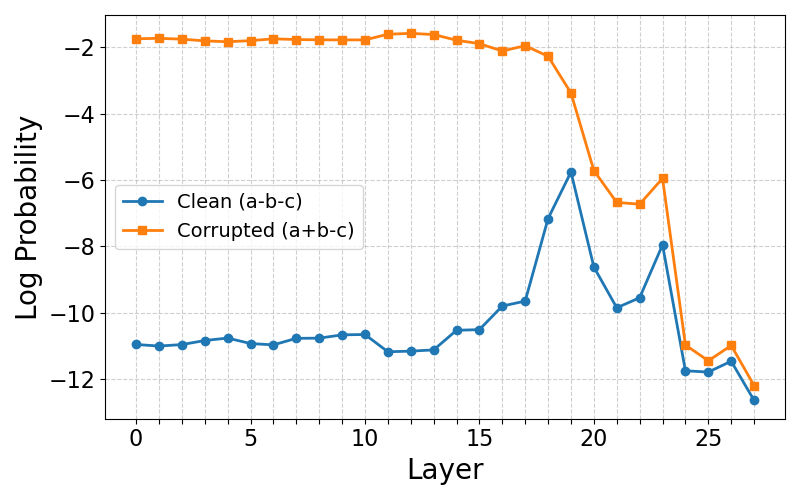}
        \caption{$x-y-z$ to $a+b-c$}
        \label{fig:add_plot1}
    \end{subfigure}
    \hfill
    \begin{subfigure}[b]{0.24\linewidth}
        \centering
        \includegraphics[width=\linewidth]{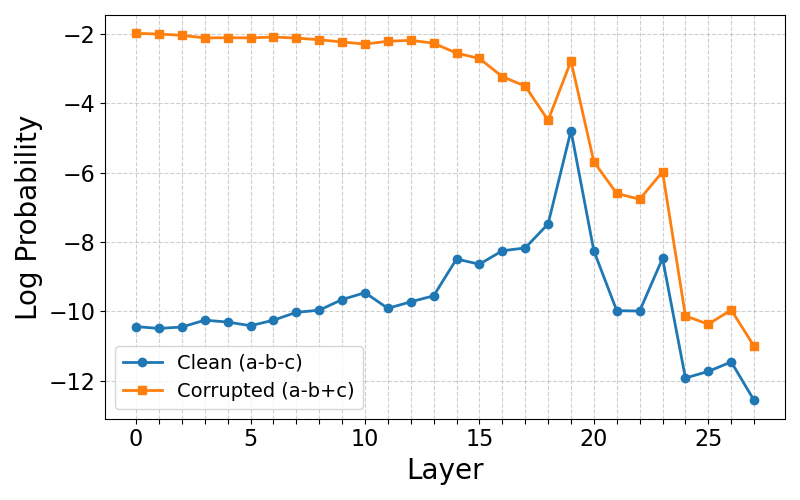}
        \caption{$x-y-z$ to $a-b+c$}
        \label{fig:add_plot2}
    \end{subfigure}
    \hfill
    \begin{subfigure}[b]{0.24\linewidth}
        \centering
        \includegraphics[width=\linewidth]{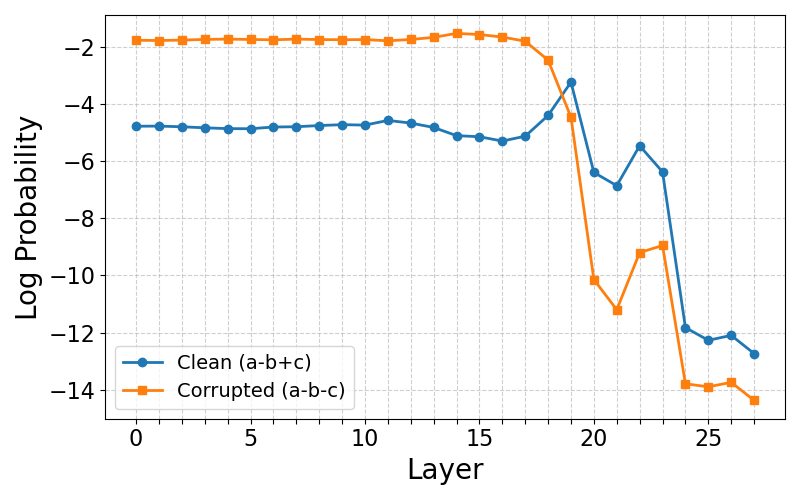}
        \caption{$x-y+z$ to $a-b-c$}
        \label{fig:add_plot3}
    \end{subfigure}
    \hfill
    \begin{subfigure}[b]{0.24\linewidth}
        \centering
        \includegraphics[width=\linewidth]{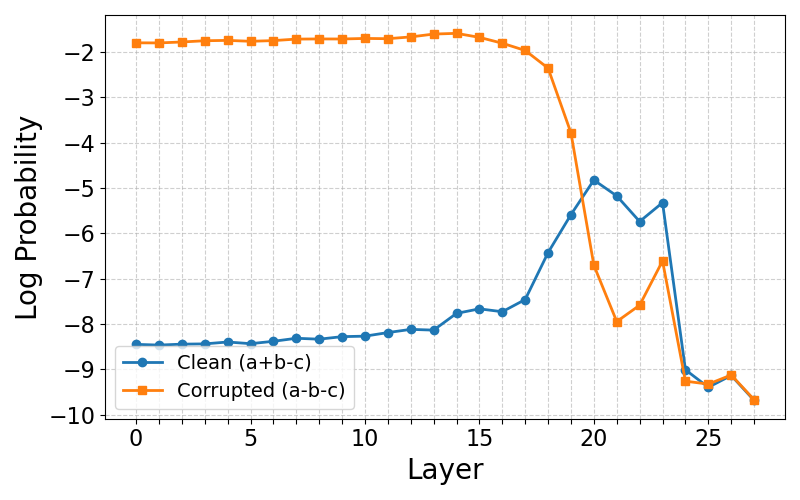}
        \caption{$x+y-z$ to $a-b-c$}
        \label{fig:add_plot4}
    \end{subfigure}

    \begin{subfigure}[b]{0.24\linewidth}
        \centering
        \includegraphics[width=\linewidth]{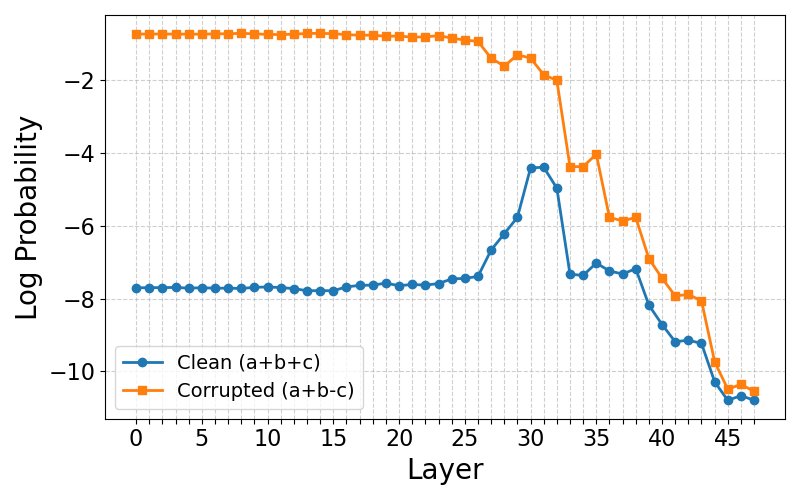}
        \caption{$x+y+z$ to $a+b-c$}
        \label{fig:add_plot1}
    \end{subfigure}
    \hfill
    \begin{subfigure}[b]{0.24\linewidth}
        \centering
        \includegraphics[width=\linewidth]{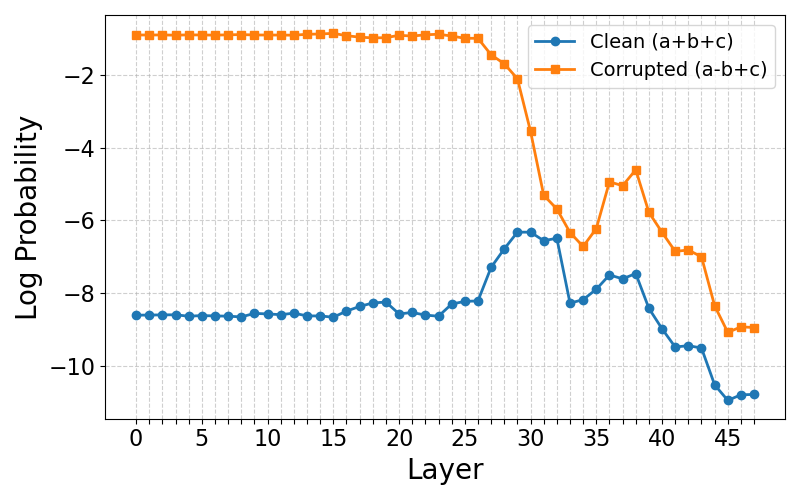}
        \caption{$x+y+z$ to $a-b+c$}
        \label{fig:add_plot2}
    \end{subfigure}
    \hfill
    \begin{subfigure}[b]{0.24\linewidth}
        \centering
        \includegraphics[width=\linewidth]{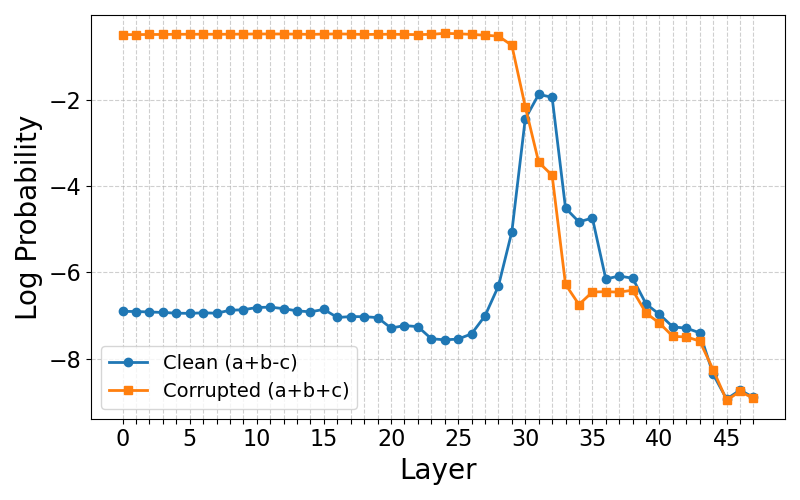}
        \caption{$x+y-z$ to $a+b+c$}
        \label{fig:add_plot3}
    \end{subfigure}
    \hfill
    \begin{subfigure}[b]{0.24\linewidth}
        \centering
        \includegraphics[width=\linewidth]{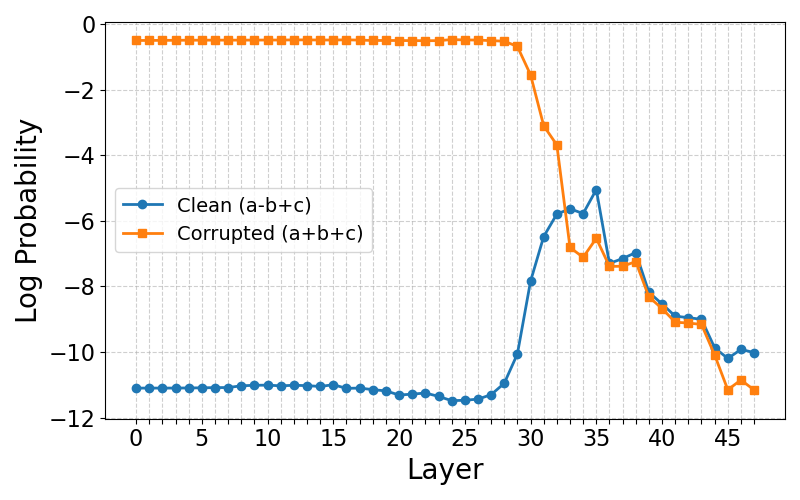}
        \caption{$x-y+z$ to $a+b+c$}
        \label{fig:add_plot4}
    \end{subfigure}

    \begin{subfigure}[b]{0.24\linewidth}
        \centering
        \includegraphics[width=\linewidth]{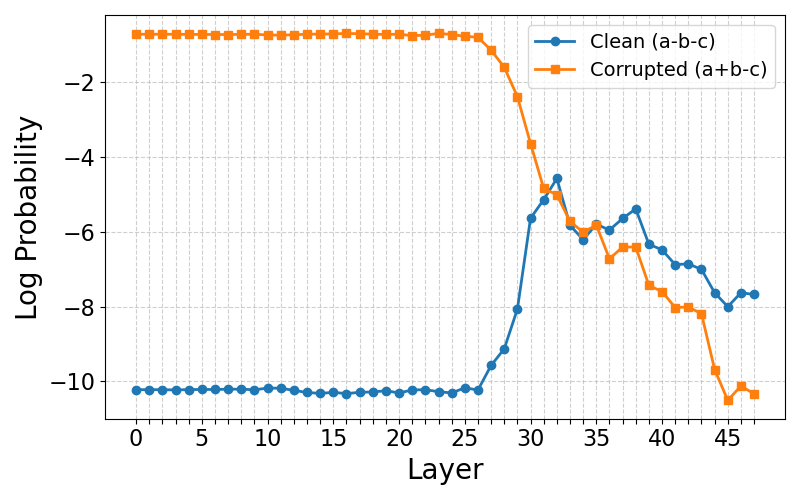}
        \caption{$x-y-z$ to $a+b-c$}
        \label{fig:add_plot1}
    \end{subfigure}
    \hfill
    \begin{subfigure}[b]{0.24\linewidth}
        \centering
        \includegraphics[width=\linewidth]{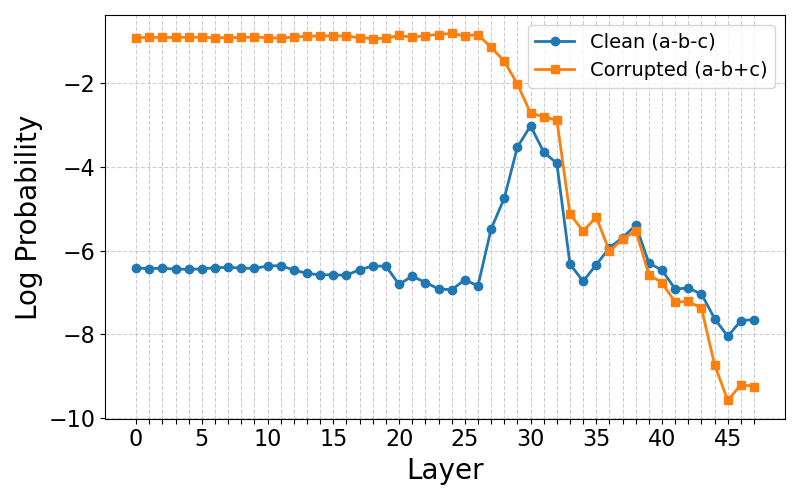}
        \caption{$x-y-z$ to $a-b+c$}
        \label{fig:add_plot2}
    \end{subfigure}
    \hfill
    \begin{subfigure}[b]{0.24\linewidth}
        \centering
        \includegraphics[width=\linewidth]{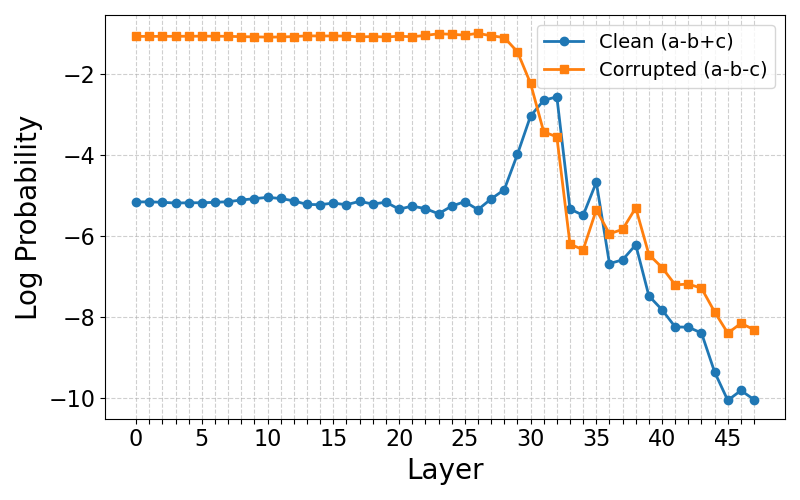}
        \caption{$x-y+z$ to $a-b-c$}
        \label{fig:add_plot3}
    \end{subfigure}
    \hfill
    \begin{subfigure}[b]{0.24\linewidth}
        \centering
        \includegraphics[width=\linewidth]{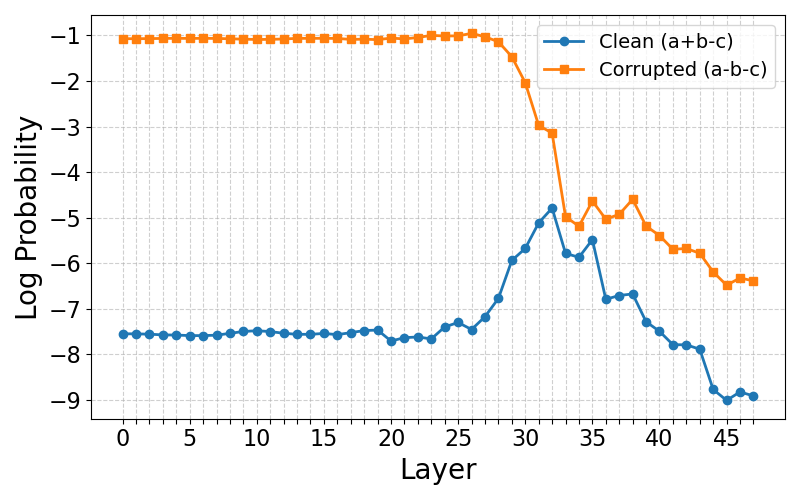}
        \caption{$x+y-z$ to $a-b-c$}
        \label{fig:add_plot4}
    \end{subfigure}
    
    \caption{\textbf{Two-operation} cross-prompt patching for \textbf{symbolic abstraction} results: \textbf{First row \& Second row}: patching symbolic logic to concrete problems for Qwen 2.5 7B. \textbf{Third row \& Fourth row}: patching symbolic logic to concrete problems for Qwen 2.5 14B.}
    \label{fig:symbolic-cross-patching-full}
\end{figure*}

See Figure~\ref{fig:symbolic-cross-patching-full} for symbolic abstraction cross-prompt patching results on Qwen 2.5 7B and Qwen 2.5 14B.



\end{document}